\documentclass{article}

\PassOptionsToPackage{numbers, compress}{natbib}
\usepackage[eandd, final]{neurips_2026}
\usepackage{amsmath} 
\usepackage[utf8]{inputenc} 
\usepackage[T1]{fontenc}    
\usepackage{hyperref}
\usepackage{xurl}
\usepackage{url}            
\usepackage{booktabs}       
\usepackage{amsfonts}       
\usepackage{nicefrac}       
\usepackage{microtype}      
\usepackage{xcolor}         
\usepackage{comment}
\usepackage{booktabs}  
\usepackage{multirow}  
\usepackage{graphicx}  
\usepackage{float}  

\title{Quantitative Assessment of Crystal Structure Prediction}

  \author{%
    Sergio Rinc\'on\thanks{Equal contribution. Correspondence to: \texttt{\{si.rincon,gf.gonzalez\}@uniandes.edu.co}} \quad
    Gabriel Gonz\'alez\footnotemark[1] \quad
    Nicol\'as Andrade \quad
    Rafael Vel\'asquez \quad
    Sebasti\'an Ojeda \\
    \textbf{Juanita Puentes} \quad
    \textbf{Nicol\'as Aparicio} \quad
    \textbf{Paula C\'ardenas} \quad
    \textbf{Pablo Arbel\'aez} \\[0.5em]
    Universidad de Los Andes, Colombia
  }
\newcommand{\modelname}{OPAL-CSP}
\newcommand{\benchname}{OPAL-Bench}
\begin{document}

\maketitle
\setcounter{footnote}{0}

\begin{abstract}
Crystal Structure Prediction (CSP) aims to predict a material's 3D structure from its chemical composition, with applications from energy storage to biomedical devices. Despite its importance, existing evaluation frameworks suffer from key limitations: they rely on threshold-sensitive or incomparable metrics, focus predominantly on geometric performance while overlooking physical plausibility, and fail to account for polymorphism, where a chemical composition can crystallize into multiple stable structures. In parallel, recent Symmetry-Informed CSP (SICSP) methods often report results alongside standard CSP baselines while receiving crystallographic templates at inference, which constrain the search space and make SICSP a fundamentally different task. To address these limitations, we introduce an Orientation-invariant, Polymorph-Aware, Large Benchmark (\benchname \footnote{Code and evaluation pipeline at: \url{https://github.com/BCV-Uniandes/OPAL}}) that evaluates CSP and SICSP as separate tasks. We evaluate 7 CSP and 2 SICSP methods under OPAL-Bench across 4 datasets using threshold-free geometric and physical distance metrics. Our results show that no method dominates across all metrics, and that geometric performance is not strongly correlated with physical plausibility. We also release \modelname{}, an SE(3)-equivariant flow-matching baseline for the CSP track that performs competitively on geometric metrics and ranks among the best on physical ones. By unifying geometric and physical metrics under a polymorph-aware protocol, \benchname{} enables a more comprehensive assessment of generative quality in CSP.
\end{abstract}

\section{Introduction}

Crystalline materials play a central role in modern technology and scientific discovery, with applications spanning energy storage \cite{suh2012hydrogen}, microelectronics \cite{mannodi2018scoping}, catalysis \cite{shi2021catalysis}, and biomedical devices \cite{zhou2014piezoelectric}. The periodic organization of atoms in a crystal governs a broad spectrum of mechanical, optical, magnetic, and electrical properties, making the three-dimensional arrangement of crystalline structures the primary target of computational study for materials discovery \cite{xiao2020origin, curtarolo2013high}. Accelerating this process increasingly depends on crystal structure prediction (CSP), the task of predicting how atoms arrange within a crystal structure directly from its composition \cite{oganov2019structure}. Deep generative models have become the dominant paradigm for CSP, displacing traditional search-based approaches that explore vast configurational spaces through costly physical simulations \cite{oganov2006crystal}. These methods leverage SE(3)- or periodic-E(3)-equivariant graph neural networks and have rapidly advanced the state of the art through a growing number of diffusion-based \cite{xie2021crystal, jiao2023crystal, zeni2025generative} and flow matching architectures \cite{luo2025crystalflow, miller2024flowmm, sriram2024flowllm}.

\begin{figure}
\centering
\includegraphics[width=1\linewidth]{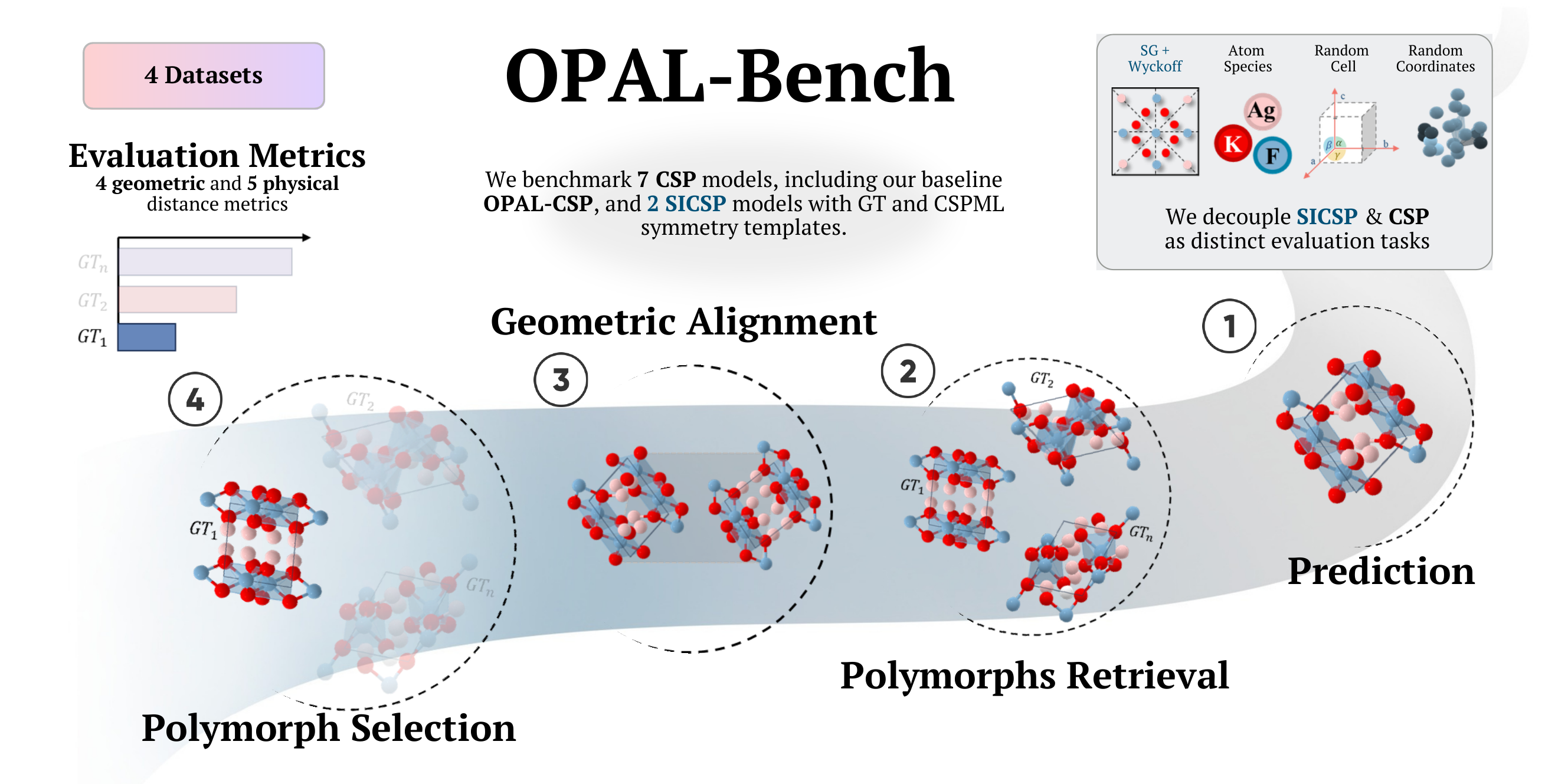}
\caption{\textbf{\benchname{} evaluation pipeline.} \benchname{} evaluates seven CSP and two SICSP methods across four datasets, with four geometric and five physical metrics, treating CSP and SICSP as fundamentally distinct tasks. For each composition, a model predicts one structure, and we retrieve all known polymorphs, the distinct stable structures of the same composition, as candidate references. We then apply species-aware SE(3) alignment for geometric metrics, while physical metrics are SE(3)-invariant and require no alignment, and score each metric against its closest polymorph.}
\label{fig:graphical_abstract}
\end{figure}

As generative methods proliferate, the question of how to evaluate them rigorously has become as consequential as the modeling advances themselves. Current evaluation protocols, however, are insufficient for a complete assessment of generative CSP methods and overlook important dimensions of model behavior. In particular, the field suffers from four main limitations that together obscure the true relative performance of CSP methods.

First, standard metrics depend on arbitrary tolerance thresholds and report errors over structurally distinct subsets of predictions across methods, making reported performance both sensitive to evaluation choices and not directly comparable between methods \cite{xie2021crystal, jiao2023crystal, miller2024flowmm}. Second, most evaluation protocols \cite{xie2021crystal, jiao2023crystal, zeni2025generative, luo2025crystalflow, miller2024flowmm, sriram2024flowllm} compare each prediction against a single fixed reference structure, even though a single chemical composition can exhibit polymorphism, adopting multiple distinct yet energetically stable arrangements that differ in geometry and symmetry \cite{cruz2014conformational, martirossyan2026all}. Third, existing CSP evaluation frameworks from the machine learning community \cite{xie2021crystal, jiao2023crystal, miller2024flowmm} measure only structural similarity and overlook physical plausibility, even though physically meaningful structures are the ultimate goal of CSP. In contrast, the crystallographic community routinely incorporates physical metrics when evaluating traditional methods \cite{wei2024cspbench, wei2024towards}. Fourth, methods that receive crystallographic symmetry templates as input at inference \cite{jiao2024space, chang2025space}, a setting we refer to as Symmetry-Informed CSP (SICSP), routinely report results alongside methods that see only the chemical formula, despite the substantial reduction in search space that symmetry information provides.

We address these limitations with an \textbf{O}rientation-invariant, \textbf{P}olymorph-\textbf{A}ware, \textbf{L}arge \textbf{Bench}mark (\textbf{\benchname{}}) for CSP and SICSP. Figure \ref{fig:graphical_abstract} summarizes the framework. Rather than comparing predictions against a single fixed reference, \benchname{} evaluates each prediction against all known polymorphs of the same composition and selects the closest as the reference. Before computing any geometric distance, it applies an explicit SE(3) alignment that enforces atom-identity constraints, ensuring that the distances we compute reflect true structural divergence rather than arbitrary orientation differences. \benchname{} further extends evaluation with physical plausibility metrics, namely formation energy, relaxed formation energy, smooth overlap of atomic positions (SOAP), X-ray diffraction pattern, and density, providing a more complete picture of generative quality. Finally, \benchname{} evaluates CSP and SICSP methods as separate tracks, enabling fair comparisons within each setting.

Alongside the benchmark, we release \modelname{}, an SE(3)-equivariant graph transformer that we train with flow matching to serve as a reference baseline for the CSP track. We use \benchname{} to evaluate seven CSP and two SICSP methods, finding that no current method dominates across all metrics and datasets, and that a structured gap separates geometric and physical performance. Our main contributions are as follows.

\begin{enumerate}
\item We introduce \benchname{}, a threshold-free, polymorph-aware benchmark that applies explicit SE(3) alignment with atom-identity constraints, unifies geometric and physical distance metrics, and separates CSP and SICSP into distinct evaluation tracks.

\item Through a systematic re-evaluation of seven CSP and two SICSP methods, we 
empirically demonstrate that geometric performance does not reliably imply physical 
plausibility, a dimension of generative quality that existing machine learning evaluation frameworks overlook.

\item We release \modelname{}, an SE(3)-equivariant flow-matching reference implementation for the CSP track. Tuned with \benchname{} in view, it performs competitively on both geometric and physical metrics, showing that the two can coexist in a single model and offering a strong baseline for future work.

\end{enumerate}

\benchname{}, along with the \modelname{} training code and model checkpoints, is available at \url{https://github.com/BCV-Uniandes/OPAL}.

\section{Related Work}
\label{sec:related_work}

\subsection{CSP Evaluation}
A CSP model takes a chemical formula as input and predicts one plausible crystal structure, comprising a periodic lattice and fractional atomic coordinates \cite{oganov2019structure, woodley2008crystal}. To assess prediction quality, most prior work in the machine learning community \cite{xie2021crystal, jiao2023crystal, luo2025crystalflow, miller2024flowmm, sriram2024flowllm} reports two output-based metrics, Match Rate and RMSE. Match Rate measures the fraction of structures the model produces that match a ground truth within predefined tolerances, and RMSE quantifies atomic positional errors normalized by the cell volume, computing them only for predictions that successfully match a reference.

These two metrics suffer from fundamental limitations. Match Rate is highly sensitive to tolerance choice, and the relative ranking of methods changes across tolerance settings, so conclusions about which model performs best depend on the threshold. RMSE faces an analogous problem. As each method computes it over its own subset of predictions that match a reference, the reported values reflect different populations, making cross-method comparisons uninformative. We empirically validate both effects in Figures  \ref{fig:distributions_MR_RMSE} and \ref{fig:comparisons_tolerances} of Appendix \ref{app:legacy_critique}. A further bias, which Martirossyan et al.~\cite{martirossyan2026all} also identify, arises from the protocol itself. Evaluation proceeds per crystal ID rather than per unique composition, so distinct polymorphs of the same formula serve as independent targets, penalizing methods that recover a different known polymorph of the same composition. To mitigate it, they propose METRe and cRMSE, which measure how well a set of generated structures covers the known references, but both still depend on predefined matching tolerances and remain purely geometric.

In contrast, the crystallographic community has evaluated traditional CSP algorithms with geometric and physical metrics. Wei et al.~\cite{wei2024towards} and CSPBench~\cite{wei2024cspbench} represent the first large-scale attempts to assess CSP algorithms using 3D point set distances, treating crystals as point clouds and computing Chamfer, Hausdorff, Sinkhorn, and RMSE across sets of 16 and 180 test structures, respectively. CSPBench additionally covers physical metrics, including XRD spectrum, fingerprint, and formation energy distances, and compares predictions against available polymorphs using energy distance as the selection criterion.

Despite these contributions, both benchmarks share important geometric limitations. First, none of the geometric metrics enforce atomic identity constraints, meaning distances can reflect cross-species comparisons that distort the true structural divergence. Second, only RMSE applies any form of rotational alignment before distance computation, and even this alignment ignores atom types during superposition, while Chamfer, Hausdorff, and Sinkhorn receive no rotational alignment at all. Although CSPBench takes a step toward physical evaluation, both benchmarks evaluate only traditional methods, omitting the deep generative models that now define the state of the art and the four standard datasets the community has adopted for comparison.

OPAL-Bench addresses these gaps directly by performing explicit SE(3) alignment that enforces one-to-one correspondence between atoms of the same type before computing any distance metric, evaluating each prediction against all known polymorphs of the same composition without matching tolerances, applying this protocol to the four standard generative CSP datasets (Perov-5~\cite{castelli2012new}, Carbon-24~\cite{pickard2020airss}, MP-20~\cite{materials_project}, and MPTS-52~\cite{materials_project}), and extending evaluation with physical metrics.

\subsection{CSP Methods}

\textbf{Standard CSP.} Deep generative approaches to CSP fall into two main 
paradigms, and in both the atom types are given as input and remain fixed throughout generation. Diffusion-based methods inject noise into lattice parameters and atomic 
coordinates, then recover the structure using score functions they learn during training. CDVAE~\cite{xie2021crystal} follows a diffusion variational autoencoder 
(VAE) formulation that first predicts the lattice and then refines 
coordinates through annealed Langevin dynamics~\cite{song2019generative}. 
MatterGen~\cite{zeni2025generative} employs an SE(3)-equivariant GemNet-dT~\cite{gasteiger2021gemnet} 
score network in a diffusion denoising process over lattices and positions, while DiffCSP~\cite{jiao2023crystal} uses a periodic-E(3) denoiser over 
fractional coordinates to jointly generate lattices and atom positions.

Flow matching methods instead learn velocity fields that transport a simple prior 
to the data distribution. CrystalFlow~\cite{luo2025crystalflow} uses Conditional 
Flow Matching to jointly generate lattice parameters and fractional coordinates. FlowMM~\cite{miller2024flowmm} generalizes Riemannian Flow 
Matching~\cite{chen2023flow} to accommodate the translation, rotation, permutation, 
and periodic boundary conditions that characterize crystalline structures, and 
FlowLLM~\cite{sriram2024flowllm} combines a large language model with Riemannian 
Flow Matching to predict metastable structures from text-based representations. As a reference baseline for the CSP track, we release \modelname{}, an SE(3)-equivariant graph transformer that we train with flow matching to provide a consistent point of comparison across the four standard datasets.

\textbf{Symmetry-Informed CSP.} SICSP methods constrain the search space with crystallographic symmetry information, namely the space group and, optionally, the Wyckoff positions, which specify the symmetry-equivalent sites that atoms occupy. DiffCSP++~\cite{jiao2024space} takes the space group as input and enforces symmetry constraints on the lattice and fractional coordinates during diffusion. SGEquiDiff~\cite{chang2025space} combines space group equivariant diffusion of atomic coordinates with a discrete lattice sampler. Both methods can operate with ground truth symmetry templates or with predicted ones. CSPML~\cite{kusaba2022cspml}, for example, predicts them by retrieving structurally similar crystals from a training database via composition-based similarity.

\section{Orientation-invariant, Polymorph-Aware, Large Benchmark}
\label{sec:benchmark_methodology}

OPAL-Bench evaluates any CSP model through the following protocol. Given a test set, the model generates one prediction per unique chemical composition, and OPAL-Bench scores that prediction by computing metric-specific minimum distances to the known polymorphs of the same composition. For geometric metrics, it first applies the species-aware SE(3) alignment of Section 3.4 before computing distances; for physical metrics, it applies them directly to the predicted and reference structures. Sections 3.1-3.5 describe each component in detail.

\subsection{Tasks Formulation}

\paragraph{Standard CSP} We formally define a crystal structure as the periodic arrangement of $N$ atoms 
within a unit cell, with a vector of atom species 
$A = [a_1, a_2, \dots, a_N] \in \mathcal{A}^N$, a lattice matrix 
$L = [l_1, l_2, l_3] \in \mathbb{R}^{3 \times 3}$ 
with $|\det L| > 0$ defining the periodic unit cell, and fractional 
coordinates $F = [f_1, f_2, \dots, f_N] \in [0,1)^{3 \times N}$, which relate to 
Cartesian coordinates $X = [x_1, x_2, \dots, x_N] \in \mathbb{R}^{3 \times N}$ via $X = LF$. 
The crystal structure is invariant under global rotations and translations 
of $X$, as well as permutations of atoms sharing the same species $a_i$.

CSP consists of learning the conditional 
distribution $p(F, L \mid A)$ over the space of valid crystal structures, 
where the search space includes all pairs 
$(F, L) \in [0,1)^{3 \times N} \times \{L \in \mathbb{R}^{3\times 3} : 
|\det L| > 0\}$ consistent with the input composition $A$. A CSP model 
samples from this distribution to produce a plausible crystal structure 
$\hat{\mathcal{M}} = (A, \hat{F}, \hat{L})$. A key property of 
this distribution is polymorphism, whereby a single composition $A$ can correspond 
to multiple distinct yet energetically stable structures. We denote the set 
of known polymorphs as $\Pi(A) = \{\mathcal{M}^{(1)}, \mathcal{M}^{(2)}, 
\dots, \mathcal{M}^{(K)}\}$, where each $\mathcal{M}^{(k)} = 
(A, F^{(k)}, L^{(k)})$ shares the same composition but differs in geometry.

\paragraph{Symmetry-Informed CSP} SICSP extends this formulation by incorporating crystallographic symmetry as additional input. Formally, a SICSP model receives the chemical composition $A$, a space group label $S \in \{1, \dots, 230\}$, and optionally a set of Wyckoff positions $\mathcal{W}$, which specify the symmetry-equivalent atomic sites within the unit cell. Together, $S$ and $\mathcal{W}$ constrain both the admissible lattice geometries and the positions atoms may occupy. With the full template, a SICSP model learns $p(F, L \mid A, S, \mathcal{W})$, whose support is restricted to the pairs $(F, L)$ consistent with $S$ and $\mathcal{W}$, a subset of the CSP search space. When the input does not include $\mathcal{W}$, some methods first predict it from $A$ and $S$ with a learned model and then sample from the same distribution. Because the input assumptions of CSP and SICSP differ substantially, the two tasks are not fairly comparable, and we evaluate them under separate tracks in \benchname{}.

\subsection{Metrics} \label{main_metrics}
Our benchmark evaluates CSP models along two complementary axes, geometric 
and physical. All metrics quantify the discrepancy between a predicted 
structure $\hat{\mathcal{M}}$ and a reference $\mathcal{M}^{(k)} \in \Pi(A)$, 
each capturing a notion of distance appropriate to the property it measures. 
We provide further details on each metric in Appendix \ref{app:metrics_description}.

Geometric metrics treat crystals as labeled point clouds, where atoms act 
as points in $\mathbb{R}^3$ and chemical elements serve as labels, and measure 
structural similarity in terms of point cloud distances. We consider four such 
metrics: Chamfer distance, Hausdorff distance, Sinkhorn distance, and RMSE. By default, these metrics do not enforce atom identity during comparison 
and are not invariant to rotations, translations, and same-species permutations. 
We therefore apply an explicit alignment procedure and a per-species aggregation 
scheme before computing them, both of which we describe in Section \ref{sec:alignment}. 

Physical metrics quantify the distance between $\hat{\mathcal{M}}$ and $\mathcal{M}^{(k)}$ in terms of experimentally measurable or computationally derived properties, namely formation energy and relaxed formation energy, which we compute with Machine Learning Interatomic Potentials (MLIPs) following the setup of LeMat-GenBench~\cite{betala2025lemat}, smooth overlap of atomic positions (SOAP) \cite{SOAPmusil2021physics}, X-ray diffraction patterns, and density. As these metrics depend only on the physical properties of the structure rather than its coordinate representation, they are invariant to rotations, translations, and permutations by definition, and we apply them directly without any preprocessing. Physical metrics also complement the geometric ones beyond the recovery of known structures. A prediction that differs geometrically from every known polymorph but relaxes to a structure of comparable formation energy receives a large geometric distance and a small relaxed formation energy distance, a distinction that becomes more relevant as CSP methods move toward the discovery of new structures, as in de novo generation. Appendix \ref{physical_appendix} provides full MLIP details. To facilitate comparison with prior work, we additionally report Match Rate and RMSE in Appendix \ref{app:legacy_full_results}.

\subsection{Polymorph Handling}
\label{sec:polymorph}

We handle polymorphism by evaluating each prediction against all known 
polymorphs of the same composition. Formally, let $d(\cdot, \cdot)$ be 
any structural distance metric defined between two crystal structures of 
the same composition. Given a test set where multiple structures may share 
the same chemical formula, we generate a single prediction $\hat{\mathcal{M}}$ 
per unique composition $A$, rather than one per crystal ID. We then evaluate 
$\hat{\mathcal{M}}$ against the set of known polymorphs 
$\Pi(A) = \{\mathcal{M}^{(1)}, \mathcal{M}^{(2)}, \dots, \mathcal{M}^{(K)}\}$ 
and select the closest reference as $\mathcal{M}^* = \mathop{\mathrm{argmin}}_{\mathcal{M}^{(k)} \in \Pi(A)} d(\hat{\mathcal{M}}, \mathcal{M}^{(k)})$, reporting $d(\hat{\mathcal{M}}, \mathcal{M}^*)$ as the final score for 
that prediction. This procedure computes a metric-specific distance to the set of known valid polymorphs, reflecting that different metrics capture different notions of structural or physical proximity. Consequently, the closest polymorph may differ across metrics for the same prediction, an effect we analyze in depth in Appendices~\ref{app:correlation_analysis} and~\ref{app:polymorph_count}. 

\subsection{Species-Aware Alignment and Aggregation}
\label{sec:alignment}
Computing meaningful geometric distances between two crystal structures 
requires resolving three sources of ambiguity, namely global rotation, global 
translation, and atom permutation. We address these jointly through an 
iterative alignment procedure that finds the rigid transform 
$(R, t) \in SE(3)$ and atom permutation $\sigma$ that minimize the RMSE 
between a predicted structure $\hat{\mathcal{M}}$ and a reference 
$\mathcal{M}^*$.

Formally, let $\hat{X} \in \mathbb{R}^{3 \times N}$ and 
$X^* \in \mathbb{R}^{3 \times N}$ denote the Cartesian coordinates of 
the predicted and reference structures, respectively. We seek the rigid 
transform $(R, t)$ and permutation $\sigma$ that solve $\min_{R \in SO(3),\, t \in \mathbb{R}^3,\, \sigma \in \mathrm{Sym}(N)} \frac{1}{N} \sum_{i=1}^{N} \| R\hat{x}_i + t - x^*_{\sigma(i)} \|^2$, subject to $a_i = a^*_{\sigma(i)}$ for all $i$, with $\mathrm{Sym}(N)$ the set of permutations of $N$ atoms. We solve this by 
alternating between two steps. For the current transform $(R, t)$, 
we compute the optimal atom assignment $\sigma^*$ using the Hungarian 
algorithm with $+\infty$ cost for cross-species matches. For $\sigma^*$, 
we update $(R, t)$ by minimizing the RMSE via gradient-based optimization. We provide further implementation details in Appendix \ref{app:species_alignment}.

Once we compute the alignment, we evaluate each geometric metric in a 
species-aware manner. Let $\mathcal{A} = \{1, \dots, g\}$ denote the set 
of distinct atom species in the crystal. For each species $a \in \mathcal{A}$, 
we extract the sub-clouds $\hat{X}_a$ and $X^*_a$ containing only the 
aligned predicted and reference atoms of type $a$, with $N_a$ atoms each. 
We then compute each geometric metric $d$ independently on each pair 
$(\hat{X}_a, X^*_a)$ and aggregate the per-species scores as $D(\hat{X}, X^*) = \sum_{a \in \mathcal{A}} (N_a / N) \cdot d(\hat{X}_a, X^*_a)$, where $N = \sum_{a \in \mathcal{A}} N_a$ is the total number of atoms. 
This aggregation preserves label consistency, avoids cross-species matches, 
and reduces to the standard unlabeled metric when all atoms share the 
same species.

\subsection{Evaluation Considerations} \label{sec:eval_details}

We report the median score across test compositions for each metric, as it is less sensitive to outliers across diverse chemical systems. SICSP predictions require additional handling because some structures have degenerate lattices or atom counts inconsistent with the input composition. For atom-count mismatches, we construct compatible supercells by repeating the predicted and reference structures according to the least common multiple of their atom counts; we consistently exclude cases producing excessively large supercells across SICSP methods. Unlike Match Rate and conventional RMSE, which hide such failures by recording non-matches or omitting RMSE values, \benchname{} makes them explicit through per-method validity rates, which we report in Appendix~\ref{app:validity_rates}. Since these validity rates differ across methods, we compute every metric on the intersection of compositions for which all methods produce a valid prediction, ensuring a common evaluation set across methods. The differences in valid prediction sets and supercell construction between standard CSP and SICSP further motivate evaluating the two tracks separately rather than comparing them directly.
\section{\modelname{}}
\label{sec:our_model}
\modelname{} generates crystal structures via flow matching over the joint 
lattice-coordinate space, and two considerations motivate its design. First, flow matching 
has produced the strongest geometric results on CSP~\cite{luo2025crystalflow, miller2024flowmm}, 
indicating its continuous-time formulation suits joint lattice-coordinate prediction. 
Second, SE(3)-equivariant graph transformers respect by construction the symmetries CSP 
requires~\cite{fuchs2020se}, so we build on EquiformerV2~\cite{equiformer_v2}, which extends this family to higher-degree representations. 
\modelname{} uses separate velocity-field heads for $\hat{L}$ and $\hat{F}$ and follows the formulation of CrystalFlow~\cite{luo2025crystalflow}, with linear interpolation on the lattice and a wrapped path on fractional 
coordinates that respects periodic boundary conditions.
Full architecture, training, and hyperparameter details are in Appendix~\ref{app:opal_implementation}.

\section{Experiments}
\label{sec:benchmark_results}

\subsection{Experimental Setup}

\paragraph{Datasets} We evaluate all models on four CSP datasets: Perov-5~\cite{castelli2012new}, Carbon-24~\cite{pickard2020airss}, MP-20~\cite{materials_project}, and MPTS-52~\cite{materials_project}. Perov-5 focuses on perovskite materials with 5 atoms per cell, containing 18,928 structures, which span only 5 space groups, making it the most structurally homogeneous dataset. Carbon-24 includes 10,153 carbon-only structures with 6 to 24 atoms per cell, and exhibits the most pronounced polymorphism of the four datasets, as these structures share only 10 distinct formulas, yielding many alternative structures per composition. It spans 56 space groups, although most of its structures adopt low-symmetry triclinic or monoclinic arrangements. MP-20 and MPTS-52 contain selected inorganic crystals from the Materials Project, comprising 45,229 structures with up to 20 atoms and 40,476 structures with up to 52 atoms, respectively. They are also the most structurally diverse, with long-tailed distributions over 179 and 208 space groups. Following prior CSP work, we use the standard splits: 60/20/20 for Perov-5, Carbon-24, and MP-20, and 67.6/12.4/20 for MPTS-52. Appendices~\ref{app:polymorphism_stats} and~\ref{app:space_group_stats} provide a more detailed statistical analysis of the datasets, including polymorphism rates and space-group distributions.

\paragraph{Baselines} We evaluate seven CSP models under \benchname{}: six previously published methods, CDVAE~\cite{xie2021crystal}, DiffCSP~\cite{jiao2023crystal}, MatterGen~\cite{zeni2025generative}, CrystalFlow~\cite{luo2025crystalflow}, FlowMM~\cite{miller2024flowmm}, and FlowLLM~\cite{sriram2024flowllm}, together with our reference baseline \modelname{}. Although several recent models perform \textit{de novo} crystal generation \cite{cao2025space, yang2023scalable, gruver2024fine}, our benchmark targets CSP, where the model receives the composition as input. For the SICSP track, we evaluate DiffCSP++ and SGEquiDiff using both ground truth templates (space group and Wyckoff positions) and templates that CSPML predicts. We retrain each model from scratch following its original training procedure and sample structures as the corresponding paper describes; Appendices \ref{app:baseline_implementation} and \ref{app:computational_resources} provide implementation details and computational resources for all models.

\subsection{\benchname{} Results} \label{results_main}

Tables~\ref{tab:csp_results} and~\ref{tab:sicsp_results} report the median score
of each metric across test compositions for the CSP and SICSP tasks, respectively.
We provide additional results in Appendix~\ref{app:complete_results}, including
95\% bootstrap confidence intervals in
Appendix~\ref{app:confidence_intervals} Figures~\ref{fig:csp_ci_supertable}
and~\ref{fig:sicsp_ci_supertable}, and qualitative predictions in
Appendix~\ref{app:qualitative_results} Figures~\ref{fig:qualitative_1}--\ref{fig:qualitative_8}.
We consider a difference between methods statistically significant when their 95\% confidence intervals do not overlap. We evaluate each task on its own group-specific intersection of valid predictions (the compositions on which every method in the track has a valid value; Appendix~\ref{app:validity_rates}).

\begin{table}[t]
\centering
\caption{\textbf{CSP Results.} For each geometric and physical metric, we report the minimum distance from each prediction to the known polymorphs of the same composition and take the median across test compositions on the four standard datasets. Lower is better for all metrics. \textbf{Bold} marks the best method per dataset and metric; \underline{underline} marks the second best. Carbon-24 results should be read with caution, as its large number of polymorphs per composition makes comparisons less conclusive.}
\label{tab:csp_results}
{\renewcommand{\arraystretch}{1.3}\setlength{\tabcolsep}{4pt}%
\resizebox{\linewidth}{!}{%
\begin{tabular}{llrrrrrrrrr}
\toprule
\multicolumn{2}{c}{} & \multicolumn{4}{c}{\textbf{Geometric Distances (\AA)}} & \multicolumn{5}{c}{\textbf{Physical Distances}} \\
\cmidrule(lr){3-6}\cmidrule(lr){7-11}
 & \textbf{Model} & \textbf{Chamf.} $\downarrow$ & \textbf{Haus.} $\downarrow$ & \textbf{Sink.} $\downarrow$ & \textbf{RMSE} $\downarrow$ & \textbf{$\rho$ (g/cm$^3$)} $\downarrow$ & \textbf{SOAP} $\downarrow$ & \textbf{XRD} $\downarrow$ & \textbf{FE (eV/at.)} $\downarrow$ & \textbf{R.\ FE (eV/at.)} $\downarrow$ \\
\midrule
\multirow{7}{*}{\rotatebox[origin=c]{90}{\textsc{Perov-5}}} & FlowLLM~\cite{sriram2024flowllm} & \underline{1.218} & \underline{0.623} & 0.970 & \underline{0.610} & 0.434 & 8.05 & 2.164 & 0.175 & $1.430 \cdot 10^{-1}$ \\
 & CDVAE~\cite{xie2021crystal} & 2.295 & 1.272 & 1.655 & 1.172 & 0.413 & 10.81 & 3.180 & 3.008 & $1.784 \cdot 10^{-1}$ \\
 & CrystalFlow~\cite{luo2025crystalflow} & 1.328 & 0.698 & 1.058 & 0.670 & 1.141 & 12.11 & 3.116 & 0.422 & $1.542 \cdot 10^{-1}$ \\
 & MatterGen~\cite{zeni2025generative} & 1.231 & 0.631 & 0.955 & 0.619 & \underline{0.251} & \underline{6.48} & \underline{1.678} & \underline{0.101} & \underline{$1.330 \cdot 10^{-1}$} \\
 & DiffCSP~\cite{jiao2023crystal} & 1.229 & 0.625 & \underline{0.921} & 0.616 & \textbf{0.236} & 6.53 & \textbf{1.672} & 0.116 & {\boldmath $1.256 \cdot 10^{-1}$} \\
 & FlowMM~\cite{miller2024flowmm} & 1.226 & 0.631 & 0.956 & 0.616 & 0.345 & 7.06 & 1.870 & 0.109 & $1.429 \cdot 10^{-1}$ \\
\cmidrule(lr){2-11}
 & \modelname{} & \textbf{1.211} & \textbf{0.617} & \textbf{0.909} & \textbf{0.607} & 0.285 & \textbf{6.25} & 1.715 & \textbf{0.072} & $1.348 \cdot 10^{-1}$ \\
\midrule
\multirow{7}{*}{\rotatebox[origin=c]{90}{\textsc{Carbon-24}}} & FlowLLM & \underline{1.553} & \underline{1.317} & \underline{12.120} & 0.931 & \underline{0.002} & 17.38 & 3.230 & 3.791 & $1.331 \cdot 10^{-1}$ \\
 & CDVAE & 1.572 & 1.324 & 13.445 & 0.947 & 0.004 & 10.03 & \underline{1.608} & 0.243 & $3.262 \cdot 10^{-3}$ \\
 & CrystalFlow & 1.650 & 1.430 & 12.992 & 0.984 & 0.128 & 42.88 & 2.346 & 0.448 & $5.228 \cdot 10^{-3}$ \\
 & MatterGen & 1.788 & 1.463 & 13.565 & 1.029 & 0.003 & 13.45 & 2.035 & 0.182 & $3.198 \cdot 10^{-3}$ \\
 & DiffCSP & 1.587 & 1.450 & 13.255 & 1.004 & 0.009 & \underline{8.24} & \textbf{1.191} & \underline{0.006} & \underline{$2.250 \cdot 10^{-4}$} \\
 & FlowMM & \textbf{1.488} & 1.401 & \textbf{10.401} & \textbf{0.833} & 0.008 & 8.76 & 1.759 & 0.069 & $4.947 \cdot 10^{-3}$ \\
\cmidrule(lr){2-11}
 & \modelname{} & 1.561 & \textbf{1.281} & 12.563 & \underline{0.889} & \textbf{0.001} & \textbf{5.67} & 1.644 & \textbf{0.002} & {\boldmath $5.700 \cdot 10^{-5}$} \\
\midrule
\multirow{7}{*}{\rotatebox[origin=c]{90}{\textsc{MP-20}}} & FlowLLM & 4.039 & 2.980 & 8.154 & 2.329 & 1.142 & 11.16 & 3.090 & 2.481 & $1.515 \cdot 10^{-1}$ \\
 & CDVAE & 3.987 & 2.958 & 8.129 & 2.289 & \underline{0.155} & 3.80 & 1.605 & 0.216 & $5.070 \cdot 10^{-2}$ \\
 & CrystalFlow & 3.911 & 2.888 & 7.819 & 2.236 & 1.187 & 8.22 & 2.402 & 0.298 & {\boldmath $8.100 \cdot 10^{-5}$} \\
 & MatterGen & 4.120 & 3.021 & 8.321 & 2.366 & 0.221 & 3.58 & 1.621 & 0.066 & $4.113 \cdot 10^{-2}$ \\
 & DiffCSP & 3.919 & 2.884 & 7.869 & 2.231 & \textbf{0.128} & \textbf{2.18} & \textbf{0.941} & \textbf{0.019} & $8.094 \cdot 10^{-3}$ \\
 & FlowMM & \underline{3.866} & \underline{2.853} & \underline{7.751} & \underline{2.194} & 0.203 & \underline{2.42} & \underline{1.088} & 0.032 & \underline{$1.250 \cdot 10^{-4}$} \\
\cmidrule(lr){2-11}
 & \modelname{} & \textbf{3.700} & \textbf{2.668} & \textbf{6.952} & \textbf{2.084} & 0.214 & 2.57 & 1.138 & \underline{0.028} & $1.490 \cdot 10^{-4}$ \\
\midrule
\multirow{7}{*}{\rotatebox[origin=c]{90}{\textsc{MPTS-52}}} & FlowLLM & 4.352 & 3.851 & 21.946 & 2.779 & 2.616 & 29.67 & 4.157 & 3264.376 & $4.117 \cdot 10^{0}$ \\
 & CDVAE & 4.437 & 3.832 & 21.665 & 2.741 & 0.433 & 5.72 & 2.084 & 0.785 & $1.978 \cdot 10^{-1}$ \\
 & CrystalFlow & \textbf{4.275} & \textbf{3.653} & \textbf{20.649} & \textbf{2.600} & 0.284 & 4.17 & \underline{1.420} & 0.274 & \underline{$9.169 \cdot 10^{-2}$} \\
 & MatterGen & 4.523 & 3.927 & 22.291 & 2.803 & 0.264 & 4.11 & 1.494 & \underline{0.150} & $1.035 \cdot 10^{-1}$ \\
 & DiffCSP & 4.323 & 3.709 & 20.825 & 2.641 & \underline{0.202} & \underline{3.96} & 1.464 & 0.234 & $1.047 \cdot 10^{-1}$ \\
 & FlowMM & \underline{4.293} & \underline{3.683} & \underline{20.692} & \underline{2.630} & 0.290 & 4.86 & 1.649 & 0.619 & $1.342 \cdot 10^{-1}$ \\
\cmidrule(lr){2-11}
 & \modelname{} & 4.325 & 3.698 & 20.908 & 2.642 & \textbf{0.183} & \textbf{3.47} & \textbf{1.338} & \textbf{0.113} & {\boldmath $7.748 \cdot 10^{-2}$} \\
\bottomrule
\end{tabular}%
}}
\end{table}

\begin{table}[t]
\centering
\caption{\textbf{SICSP results.} For each geometric and physical metric, we report the minimum distance from each prediction to the known polymorphs of the same composition and take the median across test compositions, restricted to the intersection of valid predictions across all four method-template combinations. Lower is better for all metrics. \textbf{Bold} marks the best method per dataset and metric; \underline{underline} marks the second best. Carbon-24 results should be read with caution, as its large number of polymorphs per composition makes comparisons less conclusive.}
\label{tab:sicsp_results}
{\renewcommand{\arraystretch}{1.3}\setlength{\tabcolsep}{4pt}%
\resizebox{\linewidth}{!}{%
\begin{tabular}{llrrrrrrrrr}
\toprule
\multicolumn{2}{c}{} & \multicolumn{4}{c}{\textbf{Geometric Distances (\AA)}} & \multicolumn{5}{c}{\textbf{Physical Distances}} \\
\cmidrule(lr){3-6}\cmidrule(lr){7-11}
 & \textbf{Model} & \textbf{Chamf.} $\downarrow$ & \textbf{Haus.} $\downarrow$ & \textbf{Sink.} $\downarrow$ & \textbf{RMSE} $\downarrow$ & \textbf{$\rho$ (g/cm$^3$)} $\downarrow$ & \textbf{SOAP} $\downarrow$ & \textbf{XRD} $\downarrow$ & \textbf{FE (eV/at.)} $\downarrow$ & \textbf{R.\ FE (eV/at.)} $\downarrow$ \\
\midrule
\multirow{4}{*}{\rotatebox[origin=c]{90}{\scalebox{0.75}{\textsc{Perov-5}}}} & SGEquiDiff~\cite{chang2025space} (CSPML) & 1.171 & 0.594 & 0.895 & 0.588 & 0.315 & 7.21 & 1.872 & 0.152 & $2.782 \cdot 10^{-1}$ \\
 & DiffCSP++~\cite{jiao2024space} (CSPML) & 1.200 & 0.609 & 0.901 & 0.602 & 0.337 & 7.39 & 1.936 & 0.164 & $1.080 \cdot 10^{-1}$ \\
 & SGEquiDiff (GT) & \textbf{0.076} & \textbf{0.040} & \textbf{0.058} & \textbf{0.038} & \underline{0.175} & \textbf{1.96} & \textbf{0.809} & \textbf{0.013} & \underline{$5.959 \cdot 10^{-2}$} \\
 & DiffCSP++ (GT) & \underline{0.323} & \underline{0.165} & \underline{0.245} & \underline{0.162} & \textbf{0.173} & \underline{2.09} & \underline{0.856} & \underline{0.016} & {\boldmath $2.033 \cdot 10^{-2}$} \\
\midrule
\multirow{4}{*}{\rotatebox[origin=c]{90}{\scalebox{0.75}{\textsc{Carbon-24}}}} & SGEquiDiff (CSPML) & 1.752 & 1.563 & \underline{13.211} & 1.048 & 0.489 & 83.14 & 5.410 & 1.488 & $1.609 \cdot 10^{-1}$ \\
 & DiffCSP++ (CSPML) & \textbf{1.521} & \underline{1.352} & \textbf{12.890} & \textbf{0.899} & \textbf{0.006} & \textbf{8.68} & \textbf{1.522} & \textbf{0.003} & {\boldmath $8.680 \cdot 10^{-4}$} \\
 & SGEquiDiff (GT) & 1.797 & 1.595 & 15.819 & 1.101 & 1.225 & 243.99 & 4.535 & 3.917 & $1.602 \cdot 10^{-1}$ \\
 & DiffCSP++ (GT) & \underline{1.669} & \textbf{1.351} & 13.716 & \underline{1.025} & \underline{0.007} & \underline{9.56} & \underline{1.769} & \underline{0.078} & \underline{$1.432 \cdot 10^{-3}$} \\
\midrule
\multirow{4}{*}{\rotatebox[origin=c]{90}{\scalebox{0.75}{\textsc{MP-20}}}} & SGEquiDiff (CSPML) & 4.158 & 3.509 & 15.768 & 2.519 & 0.223 & 2.30 & 1.075 & 0.029 & $7.890 \cdot 10^{-4}$ \\
 & DiffCSP++ (CSPML) & 4.104 & 3.439 & 15.655 & 2.475 & \underline{0.212} & 1.96 & 0.954 & \underline{0.022} & $4.860 \cdot 10^{-4}$ \\
 & SGEquiDiff (GT) & \underline{3.988} & \underline{3.354} & \underline{15.175} & \underline{2.392} & 0.216 & \underline{1.88} & \underline{0.944} & 0.023 & \underline{$3.680 \cdot 10^{-4}$} \\
 & DiffCSP++ (GT) & \textbf{3.949} & \textbf{3.313} & \textbf{15.085} & \textbf{2.369} & \textbf{0.206} & \textbf{1.86} & \textbf{0.931} & \textbf{0.021} & {\boldmath $3.120 \cdot 10^{-4}$} \\
\midrule
\multirow{4}{*}{\rotatebox[origin=c]{90}{\scalebox{0.75}{\textsc{MPTS-52}}}} & SGEquiDiff (CSPML) & 4.389 & 3.878 & 28.575 & 2.722 & \underline{0.302} & 4.64 & 1.586 & 0.280 & $8.320 \cdot 10^{-2}$ \\
 & DiffCSP++ (CSPML) & 4.342 & 3.840 & 27.860 & 2.696 & 0.430 & 5.11 & 1.606 & 0.222 & $6.440 \cdot 10^{-2}$ \\
 & SGEquiDiff (GT) & \underline{4.161} & \underline{3.593} & \underline{26.652} & \underline{2.548} & \textbf{0.236} & \textbf{3.68} & \textbf{1.332} & \underline{0.202} & \underline{$5.551 \cdot 10^{-2}$} \\
 & DiffCSP++ (GT) & \textbf{4.114} & \textbf{3.533} & \textbf{25.903} & \textbf{2.505} & 0.323 & \underline{3.99} & \underline{1.386} & \textbf{0.192} & {\boldmath $5.258 \cdot 10^{-2}$} \\
\bottomrule
\end{tabular}%
}}
\end{table}
\paragraph{CSP Results} 
Table~\ref{tab:csp_results} and Appendix Figure~\ref{fig:csp_ci_supertable} show that no single method achieves uniformly best results across all metrics and datasets, confirming that CSP remains an open problem where different methods excel at different aspects of structural quality. \modelname{}, FlowMM, and DiffCSP emerge as the most consistently competitive methods overall, though their relative rankings shift considerably across metrics and datasets. We analyze the agreement between the rankings that different metrics produce in Appendix~\ref{app:ranking_agreement}.

Regarding geometric metrics, flow-based methods tend to lead. \modelname{} and FlowMM perform strongly on Perov-5 and MP-20, whereas CrystalFlow exhibits pronounced dataset dependence, ranking near the bottom on smaller datasets but leading on MPTS-52. Diffusion-based methods generally lag behind on geometric metrics on Perov-5, MP-20, and MPTS-52. On Carbon-24, no method achieves a statistically significant advantage, owing to the high structural variance of carbon polymorphs.

On physical metrics, the flow advantage diminishes. \modelname{} and DiffCSP show the strongest performance, with DiffCSP particularly competitive on density, a proxy for unit cell quality independent of atomic positions. This gap between the geometric and physical performance of flow- and diffusion-based methods suggests that the two families of metrics can rank methods differently. FlowLLM illustrates this most clearly. It achieves competitive geometric scores yet produces a median formation energy error of 3264 eV/atom on MPTS-52 (Table~\ref{tab:csp_results}), orders of magnitude above the eV/atom-scale errors of every other method, as its predictions contain atoms at near-coincident positions (Appendix~\ref{app:qualitative_results}). This dissociation is not specific to FlowLLM. Across all methods, we find that geometric and physical distances are only weakly correlated on Perov-5, MP-20, and MPTS-52 (Appendices~\ref{app:correlation_analysis} and~\ref{app:physical_sensitivity} detail the full analysis, including single-polymorph subsets and alternative SOAP and XRD settings). Thus, even under a polymorph-aware protocol, geometric performance does not reliably imply physical plausibility.

\modelname{} achieves the best geometric scores on Perov-5 and MP-20, the best formation energy on three of the four datasets, and the best score on every physical metric on MPTS-52, showing that strong geometric and physical performance can coexist in a single model. Consistent with the broader pattern that no method dominates across all metrics, DiffCSP outperforms \modelname{} on several physical metrics on Perov-5 and MP-20, and CrystalFlow and FlowMM outperform it on geometric metrics on MPTS-52. \modelname{} uses per-dataset hyperparameters, while the baselines run at their released configurations (Appendix \ref{app:baseline_implementation}).

Dataset characteristics further shape relative standings. Perov-5's structural homogeneity yields clear separations between methods, MPTS-52 produces the most consistent rankings within each metric family despite its structural complexity, MP-20 provides the broadest comparison as the largest dataset, and Carbon-24 remains the most difficult to evaluate conclusively, given its high number of polymorphs per composition.

\paragraph{SICSP Results} 
Table~\ref{tab:sicsp_results} and Appendix Figure \ref{fig:sicsp_ci_supertable} present the performance results, and Appendix \ref{app:validity_rates} reports validity rates and evaluation-set intersection sizes. As expected, ground-truth templates yield better performance in most datasets, with Carbon-24 as a notable exception, where CSPML templates outperform ground truth for both methods. This likely arises from the high structural diversity of carbon polymorphs, which may constrain ground truth templates to a space group misaligned with their learned representations. Across datasets, DiffCSP++ tends to outperform SGEquiDiff regardless of template source, with two recurring exceptions. On Perov-5, SGEquiDiff shows a statistically significant advantage in geometric metrics and formation energy, though differences in the remaining physical metrics fall within confidence intervals. On MPTS-52, SGEquiDiff achieves a significant advantage only on density and SOAP, while most other metrics remain competitive. The recurrence of these patterns across both template sources suggests they reflect architectural differences rather than sensitivity to template quality.

These results depend on what each template source assumes. Ground truth templates presuppose knowledge of the target symmetry, which is one of the key unknowns that CSP seeks to determine, making this setting an idealized upper bound. CSPML templates avoid this assumption but introduce an additional source of error. The composition that CSPML retrieves may exhibit multiple valid symmetries due to polymorphism, and the correct space group may depend on several factors that the template does not encode. By constraining generation to a single fixed symmetry, both ground truth and CSPML templates implicitly assume that one space group is sufficient to describe the target. This assumption breaks down for compositions with high polymorphic diversity, where multiple distinct symmetries are equally valid and the correct one may not match the one that the template provides.

\subsection{Ablation Study}

We ablate the two main components of the \benchname{} protocol: geometric
alignment and polymorph handling. We compare four variants: the full protocol,
alignment only, polymorph handling only, and neither component. Because alignment
affects only geometric metrics, we restrict this analysis to Chamfer, Hausdorff,
Sinkhorn, and RMSE; Appendix~\ref{app:opal_bench_ablations} reports the full
results. Geometric alignment consistently reduces distances across datasets and
models (roughly 25 to 60\% on Carbon-24, MP-20, and MPTS-52 and fourfold on Perov-5) by removing unresolved rotations, translations, and same-species
permutations, making geometric errors reflect structural differences rather than
coordinate-frame artifacts. Polymorph handling has a smaller effect on Perov-5,
MP-20, and MPTS-52 (below 10\%), where each composition has relatively few known polymorphs,
but a much larger effect on Carbon-24 (30 to 40\%), where many polymorphs share the same
formula and the reference choice strongly affects the score. As each metric takes the minimum over the known polymorphs, compositions with more reference structures tend to obtain lower distances (we further analyze this effect in Appendix~\ref{app:polymorph_count}). Together, these
results confirm that both components are complementary and independently
necessary for a fair and informative evaluation of CSP models.

\subsection{Limitations and Broader Impact}

\benchname{} operates within the following constraints. First, it evaluates predictions
against the polymorphs available in the reference datasets, introducing
a bias tied to the polymorph diversity those datasets capture; we expect this
bias to diminish naturally as crystal databases grow. The minimum over known polymorphs also favors compositions with more reference structures, an effect that is strongest on Carbon-24 and that we quantify in Appendix~\ref{app:polymorph_count}. We discuss evaluation beyond known polymorphs in Appendix~\ref{app:beyond_known}.

Second, the
benchmark currently covers a representative set of physical metrics, and future
work may extend it with additional properties and relaxation pipelines as the
field advances. Formation energy metrics are necessarily tied to the MLIP
and relaxation pipeline that we use for evaluation. We follow the LeMat-GenBench \cite{betala2025lemat} setup and use its ensemble of three MLIPs, so that an external configuration fixes the scoring function in advance rather than us choosing it. Although averaging may not remove limitations shared by the three potentials, we score every method under the same ensemble, and rankings are largely preserved under each individual potential (Appendix~\ref{app:mlip_sensitivity}).

Third, \benchname{} evaluates a single prediction per composition. Prior CSP work often draws several samples per composition and reports the best match~\cite{xie2021crystal, jiao2023crystal}, but under \benchname{} this selection would require choosing which metric defines the best sample, favoring that metric over the others. Multi-sample evaluation also raises a distinct question, the generative diversity of a method, for instance how many of the known polymorphs of a composition its samples cover, which a best-of-$k$ protocol does not capture. Evaluating diversity under a multi-sample protocol therefore remains a natural direction for future investigation.

Fourth, \benchname{} is agnostic to how a model is trained, so in principle it can also evaluate pretrained models and compare them with the methods reported here. Doing so, however, requires checking for data leakage between their training data and the test splits, since overlap would inflate their scores. For example, MP-20 and MPTS-52 derive from the Materials Project, which also underlies common pretraining corpora.

This work advances the rigorous evaluation of CSP, with direct positive
implications for materials discovery in domains such as energy storage,
catalysis, and microelectronics. However, more capable generative models could accelerate the design of materials with hazardous properties, such as toxic or energetic compounds, though we assess this risk as limited given that any computationally predicted structure must still undergo rigorous experimental synthesis and characterization before physical realization.

\section{Conclusion}
\label{sec:conclusion}
Crystal structure prediction is central to materials discovery, and the precision of its evaluation bounds progress in the field. We introduce \benchname{}, an orientation-invariant polymorph-aware benchmark for crystal structure prediction that addresses four fundamental limitations of existing evaluation frameworks: the sensitivity of Match Rate to arbitrary tolerance thresholds and the comparison of RMSE over method-specific subsets of predictions, the systematic bias of single-reference evaluation, the absence of physical plausibility metrics in machine learning evaluation protocols, and the direct comparison between symmetry-informed and standard CSP methods. By unifying geometric and physical metrics under a principled polymorph-aware protocol with explicit SE(3) alignment, \benchname{} enables comprehensive and fair comparisons across generative CSP models. Our experiments reveal that geometric performance does not reliably imply physical plausibility, demonstrating that structural metrics alone are insufficient to assess generative quality and that physical metrics constitute a necessary component of rigorous CSP evaluation. We also release \modelname{}, a flow-matching baseline that performs competitively on geometric metrics and ranks among the best on physical ones, showing that both can coexist in a single model. \benchname{} provides the CSP community with a standard evaluation framework, enabling reliable measurement of progress in crystal structure prediction for materials discovery.

\begin{ack}
This work was supported by Azure sponsorship credits granted by Microsoft's AI for Good Research Lab. The authors declare no competing interests.
\end{ack}

{\small
\bibliographystyle{unsrtnat}
\bibliography{references}
}

\newpage
\appendix

\newcommand{\ablationFourMetricWidth}{0.58\linewidth}
\newcommand{\ablationSevenMetricWidth}{0.82\linewidth}
\newcommand{\fullCSPGeometricWidth}{0.82\linewidth}
\newcommand{\fullCSPPhysicalWidth}{0.95\linewidth}

\section{Critique and Justification of Metrics}

\subsection{Empirical critique of Match Rate and RMSE} \label{app:legacy_critique}
The CSP literature standardizes evaluation around two metrics: Match Rate, which measures the fraction of predictions matching a reference within fixed tolerances, and RMSE, which quantifies positional error among matched predictions. Both metrics depend on a tolerance triplet $(s_\mathrm{tol}, \theta_\mathrm{tol}, l_\mathrm{tol})$ that controls the structural similarity threshold for matching. We show that this dependence introduces two failure modes that compromise cross-method comparisons: (i) reported values shift substantially with the tolerance choice, and (ii) the standard tolerance setting yields rankings that contradict the rankings at neighboring configurations. Subsection~\ref{app:tolerance_sensitivity} demonstrates both failures empirically. Subsection~\ref{app:legacy_full_results} reports Match Rate and RMSE for all benchmarked methods at the standard tolerance setting for compatibility with prior work, and we explicitly carry forward the limitations from \ref{app:tolerance_sensitivity}.

\subsubsection{Sensitivity to tolerance thresholds} \label{app:tolerance_sensitivity}

We sample 571 tolerance configurations uniformly from $s_\mathrm{tol} \in [0.1, 0.95]$, $\theta_\mathrm{tol} \in [5^\circ, 85^\circ]$, and $l_\mathrm{tol} \in [0.1, 0.95]$, and recompute Match Rate and RMSE on MPTS-52 under every configuration. Recomputing both metrics requires re-running the structural matching algorithm for every prediction under every triplet, which restricts this analysis to four representative methods; since the instability that we document below reflects properties of the evaluation protocol rather than of any particular method, the pattern extends to the remaining methods in the benchmark.

\begin{figure}[H]
    \centering
    \includegraphics[width=0.7\linewidth]{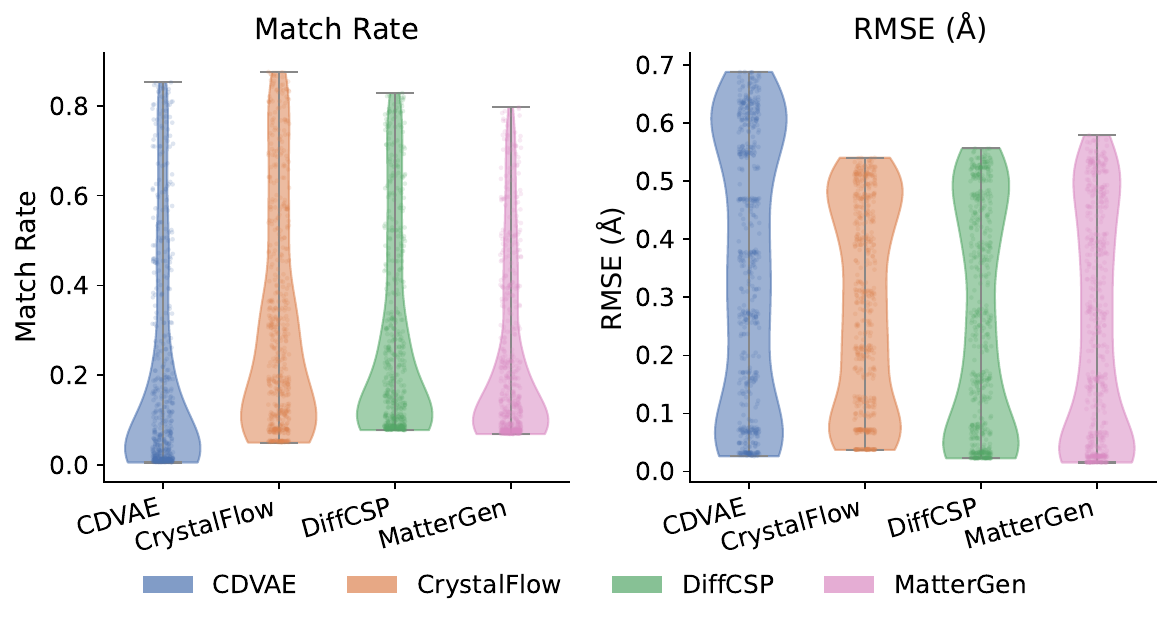}
    \caption{\textbf{Distribution of Match Rate (left) and RMSE (right) for each method across 571 tolerance configurations on MPTS-52.} We sample 571 tolerance configurations uniformly from $s_\mathrm{tol} \in [0.1, 0.95]$, $\theta_\mathrm{tol} \in [5^\circ, 85^\circ]$, and $l_\mathrm{tol} \in [0.1, 0.95]$, and recompute both metrics under each configuration. Each violin shows the full distribution across configurations. The wide spread within each method indicates that reported metric values depend strongly on the choice of tolerance parameters.}
    \label{fig:distributions_MR_RMSE}
\end{figure}

Figure~\ref{fig:distributions_MR_RMSE} shows the distribution of Match Rate and RMSE across the 571 configurations. Both metrics span nearly their entire admissible range within every method, with Match Rate covering roughly $[0, 0.85]$ and RMSE covering roughly $[0, 0.65]$~\AA. Reporting either metric at any single tolerance setting therefore yields one realization of a random variable whose support covers nearly the full range of plausible values.

\begin{figure}[H]
    \centering
    \includegraphics[width=1.0\linewidth]{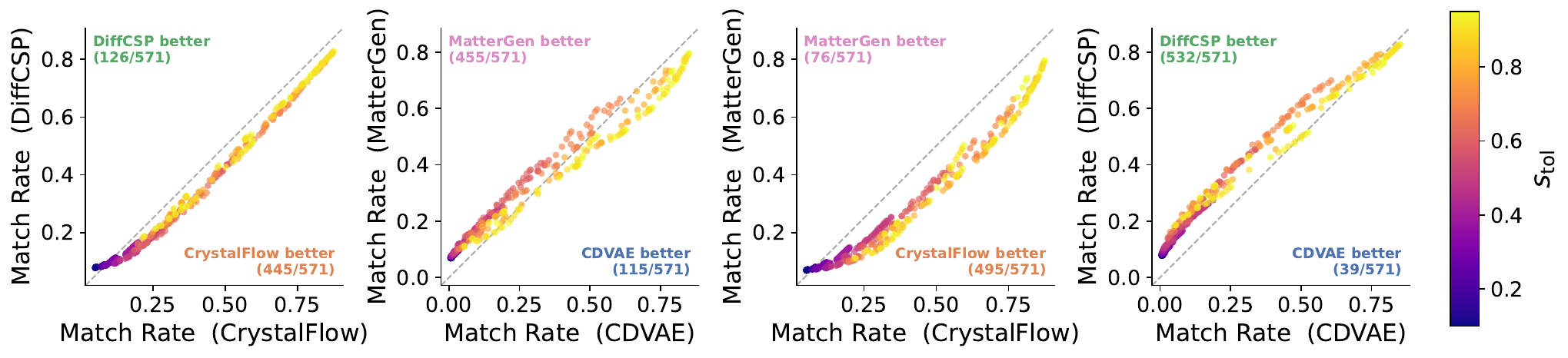}
    \caption{\textbf{Pairwise comparison of Match Rate across 571 tolerance configurations on MPTS-52.} We sample 571 tolerance configurations uniformly from $s_\mathrm{tol} \in [0.1, 0.95]$, $\theta_\mathrm{tol} \in [5^\circ, 85^\circ]$, and $l_\mathrm{tol} \in [0.1, 0.95]$. Each point corresponds to one configuration, colored by $s_\mathrm{tol}$. Points above the dashed diagonal indicate that the $y$-axis method achieves higher Match Rate than the $x$-axis method under that configuration; points below favor the $x$-axis method. Annotations report the number of configurations favoring each method out of 571. No method dominates among the pairs shown, and the standard tolerance setting ($s_\mathrm{tol}=0.5$, $\theta_\mathrm{tol}=10^\circ$, $l_\mathrm{tol}=0.3$) sits in regions where small perturbations flip rankings. Table~\ref{tab:pairwise_wins} reports the only two pairs in which one method dominates across all configurations.}
    \label{fig:comparisons_tolerances}
\end{figure}

Table~\ref{tab:pairwise_wins} and Figure~\ref{fig:comparisons_tolerances} examine whether relative rankings remain stable across configurations even when individual values do not. Two observations follow. First, most pairs exhibit substantial ranking instability, with reversal rates reaching up to 22.1\%, indicating that the relative ranking between two methods is in general a parameter choice rather than an empirical finding; only CDVAE--CrystalFlow and DiffCSP--MatterGen show full dominance among the methods that we analyze, and constitute the only conclusions robust to the tolerance choice. Second, $s_\mathrm{tol}$ accounts for most of the ranking instability, with $\theta_\mathrm{tol}$ and $l_\mathrm{tol}$ contributing secondarily.

\begin{table}[H]                                   
  \centering                                
  \caption{\textbf{Pairwise Match Rate comparison across 571 tolerance configurations on MPTS-52.} Each row compares two methods and reports how many tolerance configurations favor each one, where a method is favored when it achieves the higher Match Rate. The reversal rate is the fraction of configurations in which the less frequently favored method wins, indicating how sensitive the pairwise ranking is to the tolerance choice.}                             
  \label{tab:pairwise_wins}                          
  \begin{tabular}{lcccc}                             
  \toprule                                           
  \textbf{Method A} & \textbf{Method B} & \textbf{A  
  wins} & \textbf{B wins} & \textbf{Reversal rate} \\
  \midrule
  CrystalFlow & DiffCSP     & 445 & 126 & 22.1\% \\  
  CDVAE       & MatterGen   & 115 & 455 & 20.1\% \\
  CrystalFlow & MatterGen   & 495 &  76 & 13.3\% \\  
  CDVAE       & DiffCSP     &  39 & 532 &  6.8\% \\
  CDVAE       & CrystalFlow &   0 & 571 &  0.0\% \\  
  DiffCSP     & MatterGen   & 571 &   0 &  0.0\% \\  
  \bottomrule                                        
  \end{tabular}                                      
  \end{table}

\subsubsection{Full Match Rate and RMSE results} \label{app:legacy_full_results}

Tables~\ref{tab:supp_mr_rmse} and~\ref{tab:supp_mr_rmse_cat_b} report Match Rate and RMSE at the standard tolerance setting ($s_\mathrm{tol}=0.5$, $\theta_\mathrm{tol}=10^\circ$, $l_\mathrm{tol}=0.3$) for all CSP and SICSP methods, respectively, on the four standard datasets. We include these results for compatibility with prior work, since most published CSP papers report values under this convention. We retrained every baseline from its released code and configuration. DiffCSP on MP-20 reaches a Match Rate of 0.490 and an RMSE of 0.065, against the published 0.515 and 0.063, and a small number of other baselines, such as CDVAE, show similar isolated gaps. We could not identify a cause beyond stochastic variability, since we followed each method's released repository and procedure without modification.

Two limitations affect the interpretation of these tables. First, both metrics depend on the chosen tolerance triplet, and the relative ordering of methods is not invariant to that choice; the values we report here correspond to the convention that prior work adopts. Second, each method computes RMSE only over its own matched subset, so each method averages the reported values over different populations of structures: a method matching 30\% of test compositions reports RMSE over a different and typically easier subset than a method matching 80\%, since the matched subset of a weaker method leans toward compositions where matching succeeded.

The tolerance-free metrics we introduce in Section~\ref{main_metrics} address both limitations directly: they compute distances over the full set of valid predictions without depending on a matching threshold,\footnote{SOAP uses a fixed neighbor cutoff radius of $r_\mathrm{cut}=4.0$~\AA{} and XRD uses a fixed $2\theta$ range of $5^{\circ}$–$90^{\circ}$. These are descriptor-level design choices, not matching tolerances, and do not determine whether the protocol counts a prediction as correct.} and they evaluate every method over the same set of compositions. Appendix~\ref{app:physical_sensitivity} shows that our conclusions are robust to the SOAP and XRD descriptor settings.

\begin{table}[H]
  \centering
    \caption{\textbf{Conventional Match Rate and RMSE results for CSP methods.} We report the standard threshold-based metrics used by prior CSP methods on Perov-5, Carbon-24, MP-20, and MPTS-52, using $(s_{\mathrm{tol}}=0.5, \theta_{\mathrm{tol}}=10, l_{\mathrm{tol}}=0.3)$. Match Rate (MR; higher is better) measures the fraction of predictions that match a reference structure under these tolerances, while RMSE (lower is better) is computed only over matched predictions. \textbf{Bold} marks the best value and \underline{underline} marks the second-best value for each dataset and metric.}
  \label{tab:supp_mr_rmse}
  \small
  \setlength{\tabcolsep}{4pt}
  \begin{tabular}{lcccccccc}
    \toprule
    \multirow{2}{*}{\textbf{Model}} & \multicolumn{2}{c}{\textsc{Perov-5}} & \multicolumn{2}{c}{\textsc{Carbon-24}} & \multicolumn{2}{c}{\textsc{MP-20}} & \multicolumn{2}{c}{\textsc{MPTS-52}} \\
    \cmidrule(lr){2-3}\cmidrule(lr){4-5}\cmidrule(lr){6-7}\cmidrule(lr){8-9}
     & MR $\uparrow$ & RMSE $\downarrow$ & MR $\uparrow$ & RMSE $\downarrow$ & MR $\uparrow$ & RMSE $\downarrow$ & MR $\uparrow$ & RMSE $\downarrow$ \\
    \midrule
    CDVAE & 0.447 & 0.330 & \textbf{0.194} & 0.355 & 0.333 & 0.123 & 0.042 & 0.280 \\
    DiffCSP & 0.516 & 0.081 & 0.155 & \underline{0.299} & 0.490 & 0.065 & 0.127 & 0.149 \\
    MatterGen & 0.507 & \underline{0.073} & 0.167 & 0.337 & 0.293 & 0.063 & 0.100 & \underline{0.124} \\
    FlowLLM & \underline{0.520} & 0.097 & \underline{0.194} & 0.419 & 0.223 & 0.080 & 0.006 & 0.395 \\
    CrystalFlow & \textbf{0.523} & 0.098 & 0.176 & 0.316 & \textbf{0.618} & \underline{0.051} & \textbf{0.223} & 0.138 \\
    FlowMM & 0.511 & 0.080 & 0.157 & 0.301 & \underline{0.598} & 0.063 & 0.125 & 0.212 \\
    \midrule
    \modelname & 0.511 & \textbf{0.066} & 0.147 & \textbf{0.277} & 0.555 & \textbf{0.049} & \underline{0.171} & \textbf{0.111} \\
    \bottomrule
  \end{tabular}
\end{table}

\begin{table}[H]
  \centering
    \caption{\textbf{Conventional Match Rate and RMSE results for SICSP methods.} We report the standard threshold-based metrics used by prior CSP and SICSP methods on Perov-5, Carbon-24, MP-20, and MPTS-52, using $(s_{\mathrm{tol}}=0.5, \theta_{\mathrm{tol}}=10, l_{\mathrm{tol}}=0.3)$. Results are shown for SGEquiDiff and DiffCSP++ under both CSPML-predicted symmetry templates and ground truth (GT) symmetry templates. Match Rate (MR; higher is better) measures the fraction of predictions that match a reference structure under these tolerances, while RMSE (lower is better) is computed only over matched predictions. \textbf{Bold} marks the best value and \underline{underline} marks the second-best value for each dataset and metric.}
  \label{tab:supp_mr_rmse_cat_b}
  \small
  \setlength{\tabcolsep}{4pt}
  \begin{tabular}{lcccccccc}
    \toprule
    \multirow{2}{*}{\textbf{Model}} & \multicolumn{2}{c}{\textsc{Perov-5}} & \multicolumn{2}{c}{\textsc{Carbon-24}} & \multicolumn{2}{c}{\textsc{MP-20}} & \multicolumn{2}{c}{\textsc{MPTS-52}} \\
    \cmidrule(lr){2-3}\cmidrule(lr){4-5}\cmidrule(lr){6-7}\cmidrule(lr){8-9}
     & MR $\uparrow$ & RMSE $\downarrow$ & MR $\uparrow$ & RMSE $\downarrow$ & MR $\uparrow$ & RMSE $\downarrow$ & MR $\uparrow$ & RMSE $\downarrow$ \\
    \midrule
    SGEquiDiff (CSPML) & 0.530 & 0.098 & 0.103 & 0.231 & 0.635 & 0.041 & 0.277 & \textbf{0.078} \\
    DiffCSP++ (CSPML)  & 0.511 & 0.093 & \textbf{0.290} & 0.171 & \textbf{0.705} & \textbf{0.030} & 0.361 & \underline{0.093} \\
    SGEquiDiff (GT)    & \underline{0.981} & \textbf{0.025} & \underline{0.288} & \underline{0.147} & \underline{0.756} & 0.039 & \underline{0.417} & 0.093 \\
    DiffCSP++ (GT)     & \textbf{0.984} & \underline{0.051} & 0.288 & \textbf{0.112} & 0.787 & \underline{0.034} & \textbf{0.432} & 0.096 \\
    \bottomrule
  \end{tabular}
\end{table}

\subsection{Detailed description of proposed metrics}
\label{app:metrics_description}

\subsubsection{Geometric Distance Metrics}

Geometric metrics quantify structural similarity by treating each crystal as a point cloud in $\mathbb{R}^3$ and comparing the Cartesian positions of predicted and reference atoms. Unlike physical metrics, they depend on the coordinate representation and are not invariant to rotations, translations, or same-species permutations, so we compute them after the species-aware SE(3) alignment of Appendix~\ref{app:species_alignment} and aggregate them per species (Section~\ref{sec:alignment}). Throughout, $P = \{p_i\}_{i=1}^{m}$ and $Q = \{q_j\}_{j=1}^{n}$ denote the predicted and reference point sets of a given species.

\paragraph{Chamfer.} To quantify the average nearest-neighbor discrepancy between two point sets, we adopt the Chamfer distance,
\begin{equation}
    d_{\mathrm{Chamfer}}(P, Q)
    = \frac{1}{m}\sum_{p \in P} \min_{q \in Q} \|p - q\|_2
    + \frac{1}{n}\sum_{q \in Q} \min_{p \in P} \|q - p\|_2.
    \label{eq:chamfer}
\end{equation}
The bidirectional formulation penalises both atoms the prediction fails to place and atoms the prediction adds beyond the reference, making it sensitive to the global nearest-neighbor structure of each point cloud.

\paragraph{Hausdorff.} To capture the worst-case nearest-neighbor discrepancy between two point sets, we adopt the Hausdorff distance,
\begin{equation}
    d_{\mathrm{Hausdorff}}(P, Q)
    = \max\!\left(
        \max_{p \in P}\min_{q \in Q}\|p - q\|_2,\;
        \max_{q \in Q}\min_{p \in P}\|q - p\|_2
    \right).
    \label{eq:hausdorff}
\end{equation}
Unlike Chamfer, this metric is sensitive to outliers and therefore exposes large local deviations in atomic positions that average-based scores can obscure.

\paragraph{Sinkhorn.} As an entropic-regularised approximation to the Earth Mover's Distance~\citep{wei2024towards}, Sinkhorn distance is obtained from the pairwise cost matrix $C_{ij} = \|p_i - q_j\|_2$ and uniform marginals $\mathbf{a} = \mathbf{1}_m$, $\mathbf{b} = \mathbf{1}_n$ by solving
\begin{equation}
    d_{\mathrm{Sinkhorn}}(P, Q) = \langle T^*,\, C \rangle,
    \label{eq:sinkhorn}
\end{equation}
where $T^*$ minimises $\langle T, C \rangle - \varepsilon\, H(T)$ over transport plans with marginals $\mathbf{a}$ and $\mathbf{b}$, with $H(T) = -\sum_{ij} T_{ij} \log T_{ij}$ and $\varepsilon = 10^{-3}$. We run at most 100 iterations with a convergence threshold of $10^{-3}$. By balancing sensitivity to global mass transport with robustness to local outliers, this metric complements Chamfer and Hausdorff.

\paragraph{RMSE.} Given two point sets of equal size $N$, RMSE quantifies the root mean squared deviation after optimal alignment,
\begin{equation}
    d_{\mathrm{RMSE}}(P, Q)
    = \sqrt{\frac{1}{N}\sum_{i=1}^{N}\|p_i - q_i\|_2^2},
    \label{eq:rmsd}
\end{equation}
where points pair according to the optimal assignment $\sigma^*$ from the alignment procedure. In contrast to the legacy RMSE reported in prior CSP work, we evaluate it across all test compositions rather than restricting to matched predictions only.

\subsubsection{Physical Distance Metrics} \label{physical_appendix}

Physical metrics quantify structural quality in terms of experimentally observable or computationally derived properties. Each metric depends only on the physical content of the structure rather than its coordinate representation, making it invariant to rigid-body transformations and same-species permutations by construction. We apply all physical metrics directly to predicted and reference structures without any preprocessing.

\paragraph{Density.} The density of a crystal structure is $\rho = m / V$, where $m$ denotes the total atomic mass in the unit cell and $V = |\det L|$ is the cell volume. We report the absolute difference $|\hat{\rho} - \rho^*|$ in g/cm$^3$. Density serves as a proxy for unit cell quality independent of atomic positions, capturing lattice errors that geometric point-cloud metrics may miss when atomic coordinates are locally accurate.

\paragraph{SOAP.} The Smooth Overlap of Atomic Positions (SOAP) descriptor~\citep{SOAPmusil2021physics} encodes the local chemical environment of each atom via a rotationally invariant expansion of the atomic density in a basis of spherical harmonics and radial functions. We compute SOAP fingerprints using DScribe~\citep{dscribe2} with cutoff radius $r_{\mathrm{cut}} = 4.0$~\AA, $n_{\mathrm{max}} = 12$ radial basis functions, $l_{\mathrm{max}} = 9$ angular momentum channels, Gaussian broadening $\sigma = 0.5$~\AA, periodic boundary conditions, and outer product averaging across all atoms in the unit cell. The distance between predicted and reference fingerprints is the $\ell_2$ norm $\|\hat{f} - f^*\|_2$ between the two averaged descriptor vectors.

\paragraph{X-Ray Diffraction.} We simulate powder XRD patterns using the \texttt{Scatter} module of \texttt{Dans\_Diffraction}\footnote{\url{https://github.com/DanPorter/Dans_Diffraction}} over the range $2\theta \in [5^\circ, 90^\circ]$ and normalise each pattern by its total integrated area, yielding a probability distribution over $2\theta$. We measure the distance between predicted and reference patterns via the Wasserstein distance,
\begin{equation}
    d_{\mathrm{XRD}}(\hat{M}, M^*)
    = W_1\!\left(\hat{p}_{2\theta},\, p^*_{2\theta}\right),
    \label{eq:xrd}
\end{equation}
where $\hat{p}_{2\theta}$ and $p^*_{2\theta}$ are the area-normalised intensity distributions of the predicted and reference structures. The Wasserstein distance is appropriate for XRD patterns because it accounts for peak position shifts rather than treating each $2\theta$ bin independently, as $\ell_2$-based distances would.

\paragraph{Formation Energy.}

The formation energy of a crystal structure $M$ with composition $A = {a_1,\ldots,a_N}$ is
$$E_f(M) = E(M) - \sum_i n_i,\mu_{a_i},$$
where $E(M)$ is the total energy that the MLIP predicts, $n_i$ counts atoms of species $a_i$, and $\mu_{a_i}$ is the elemental reference energy of species $a_i$ that we take from DFT calculations (PBE) in the LeMat-Bulk dataset~\citep{siron2025lemat}. We evaluate $E(M)$ using the three-model MLIP ensemble from LeMat-GenBench~\citep{betala2025lemat} (Orb \cite{rhodes2025orb}, MACE-MP \cite{batatia2025foundation}, and UMA \cite{wood2025family}) and take the mean of the three predicted energies. We evaluate both the predicted structure $\hat{M}$ and the ground truth structure $M^*$ with this ensemble without prior relaxation. We report the absolute difference $|E_f(\hat{M}) - E_f(M^*)|$ in eV/atom.

\paragraph{Relaxed Formation Energy.} 

We relax both the predicted structure $\hat{M}$ and the ground truth structure $M^*$ independently using each MLIP in the ensemble, via BFGSLineSearch with a FrechetCellFilter (joint atomic and cell relaxation, $f_\mathrm{max} = 0.02$ eV/Å, max 50 steps), and then evaluate $E_f$ on the respective relaxed geometries. This symmetric evaluation removes the dependence on the original atomic positions and measures how close the two structures are in the energy landscape that the MLIP explores. We report $|E_f^\mathrm{relax}(\hat{M}) - E_f^\mathrm{relax}(M^*)|$ in eV/atom.

\subsubsection{Metrics not included in \benchname{}}
\label{app:excluded_metrics}

We also considered several additional metrics, including those of the CSPBench metric suite~\cite{wei2024cspbench}, and excluded three of them from \benchname{}. Graph Edit Distance, the CrystalNN fingerprint distance, and the minimal Wyckoff position distance either returned undefined values or failed to compute for a substantial fraction of the predicted structures. Since \benchname{} evaluates all methods on the intersection of valid predictions, these failures reduced the intersections to sets that were not representative of the test splits, which led us to exclude these metrics. For Graph Edit Distance and the CrystalNN fingerprint distance, we hypothesize that the failures arise because both require deciding which atoms are bonded before a distance can be computed, a decision that is unreliable for the degenerate geometries that generative models occasionally produce, such as overlapping atoms or unusually short interatomic distances. For the minimal Wyckoff position distance, we attribute the failures to its definition, which requires the predicted and reference structures to share a space group and Wyckoff configuration, a condition that unconstrained CSP predictions often do not satisfy.

We also excluded the energy above the convex hull, which is redundant in our setting. The energy above the hull of a structure $M$ with composition $A$ is $\Delta E_{\mathrm{hull}}(M) = E_f(M) - E_{\mathrm{hull}}(A)$, where $E_{\mathrm{hull}}(A)$ is the formation energy of the convex hull at composition $A$. Since a prediction and its reference share the same composition, $E_{\mathrm{hull}}(A)$ cancels in the difference,
\begin{equation}
    |\Delta E_{\mathrm{hull}}(\hat{M}) - \Delta E_{\mathrm{hull}}(M^*)| = |E_f(\hat{M}) - E_f(M^*)|,
\end{equation}
and the same holds after relaxation. This metric therefore coincides with the formation energy metrics defined above, while requiring a reference phase diagram that does not change the result.

\section{Benchmark Setup and Implementation Details}

\subsection{Statistical characterization of datasets}
\label{app:dataset_stats}

We characterize the four datasets that we use in our benchmark along two structural axes that determine the difficulty of the CSP and SICSP tasks: polymorphism, which controls the number of valid reference structures per composition, and space group distribution, which determines the symmetry diversity that methods must learn.

\subsubsection{Polymorphism distribution}
\label{app:polymorphism_stats}

We define a polymorph group as a set of structures that share both chemical formula and number of atoms in the unit cell. This definition is operationally motivated by the CSP task: a generative model receives a chemical formula as input and produces a structure with a fixed number of atoms that the formula determines, so two structures sharing the formula but differing in unit-cell atom count are not interchangeable references for a single prediction. The definition has practical consequences for Carbon-24, where every structure is crystallographically a polymorph of carbon, but our framework distinguishes 10 polymorph groups corresponding to unit cells with 6 to 24 atoms. 

Table~\ref{tab:polymorph_summary} reports polymorphism statistics independently per split, since the benchmark evaluates each method on the test split and only test-split polymorphs serve as references at evaluation time.

\begin{table}[H]
  \centering
\caption{\textbf{Polymorphism statistics across the four benchmark datasets, computed independently within each split.} 
Structures are grouped by shared chemical formula and unit-cell atom count. Poly. Groups counts groups containing more than one structure, Polymorphism Rate is the fraction of structures in such groups, and Max Polymorphs is the largest number of structures observed in a single group.}
  \label{tab:polymorph_summary}
  \small
  \setlength{\tabcolsep}{4pt}
  \begin{tabular}{llcccc}
    \toprule
    \textbf{Dataset} & \textbf{Split} & \textbf{Structures} & \textbf{Poly. Groups} & \textbf{Polymorphism Rate} & \textbf{Max Polymorphs} \\
    \midrule
    \multirow{3}{*}{Perov-5}
      & train & 11{,}356 & 3{,}330 & 58.6\% & 2 \\
      & val   & 3{,}787  & 362    & 19.1\% & 2 \\
      & test  & 3{,}785  & 376    & 19.9\% & 2 \\
    \midrule
    \multirow{3}{*}{MPTS-52}
      & train & 27{,}380 & 1{,}261 & 10.4\% & 8 \\
      & val   & 5{,}000  & 116    & 4.8\%  & 5 \\
      & test  & 8{,}096  & 190    & 5.0\%  & 6 \\
    \midrule
    \multirow{3}{*}{MP-20}
      & train & 27{,}136 & 1{,}490 & 14.0\% & 16 \\
      & val   & 9{,}047  & 264    & 6.7\%  & 7  \\
      & test  & 9{,}046  & 247    & 6.3\%  & 8  \\
    \midrule
    \multirow{3}{*}{Carbon-24}
      & train & 6{,}091 & 10 & 100.0\% & 1{,}980 \\
      & val   & 2{,}032 & 10 & 100.0\% & 671    \\
      & test  & 2{,}030 & 10 & 100.0\% & 705    \\
    \bottomrule
  \end{tabular}
\end{table}

The four datasets span a wide range of polymorphic complexity, with Carbon-24 at one extreme and MP-20, MPTS-52, and Perov-5 sharing a median of 2 polymorphs per group at rates of 6.3\%, 5.0\%, and 19.9\%. This contrast explains the asymmetric magnitude of the polymorph-handling effect: reference choice rarely matters in the first three datasets but is consequential for nearly every Carbon-24 composition. MP-20 and MPTS-52 polymorphism also exhibits structural heterogeneity relevant to the SICSP discussion in Section~\ref{results_main}: in the MP-20 test split, 35 polymorphic groups span monoclinic and orthorhombic systems, 24 span triclinic and monoclinic, and 20 span tetragonal and cubic, with the same pattern at smaller scale in MPTS-52. A SICSP method that conditions on a single space group template therefore restricts generation to a symmetry that may not correspond to the closest polymorph, reinforcing that template-based methods cannot accommodate polymorphic diversity across symmetries.

\subsubsection{Space group distribution}
\label{app:space_group_stats}

\begin{figure}[h]
  \centering
  \includegraphics[width=0.5\linewidth]{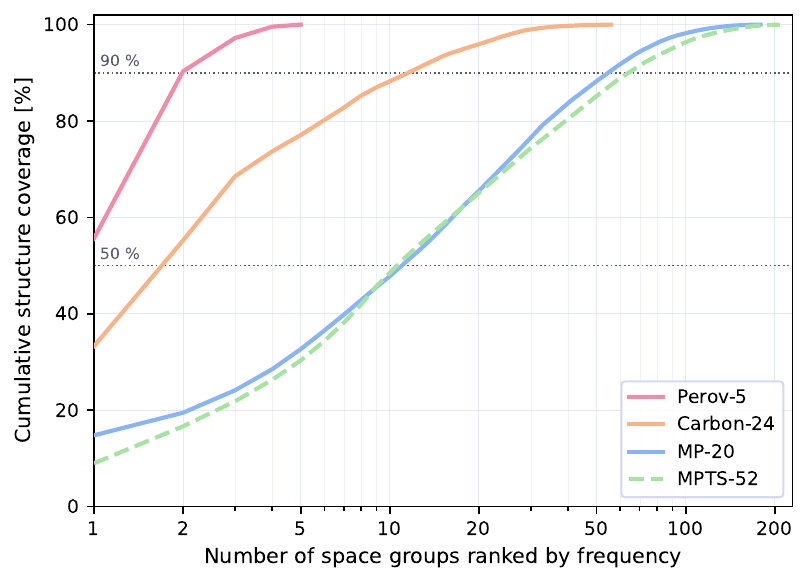}
  \caption{\textbf{Cumulative space group coverage across the four benchmark datasets.} Each curve shows the fraction of structures captured by the top-$k$ most frequent space groups, with $k$ on a logarithmic scale. Horizontal dashed lines mark the 50\% and 90\% coverage thresholds.}
  \label{fig:sg_coverage}
\end{figure}

The four datasets differ substantially in the diversity of crystallographic symmetries they contain. Table~\ref{tab:sg_summary} reports the number of distinct space groups that each dataset contains and the share of structures that the most frequent space groups capture, while Figure~\ref{fig:sg_coverage} visualizes the cumulative coverage of structures as a function of the number of space groups, after ranking them by frequency.

\begin{table}[H]
  \centering
    \caption{\textbf{Space group diversity across the four benchmark datasets.} 
    Unique SGs denotes the number of distinct space groups observed in each dataset. 
    The last three columns report the cumulative fraction of structures assigned to the top 1, top 5, and top 20 most frequent space groups within each dataset.}
  \label{tab:sg_summary}
  \small
  \begin{tabular}{lcccc}
    \toprule
    \textbf{Dataset} & \textbf{Unique SGs} & \textbf{Top 1 SG} & \textbf{Top 5 SGs} & \textbf{Top 20 SGs} \\
    \midrule
    Perov-5    & 5   & 56\% & 100\% & 100\% \\
    Carbon-24  & 56  & 33\% & 73\%  & 96\%  \\
    MP-20      & 179 & 15\% & 31\%  & 65\%  \\
    MPTS-52    & 208 & 9\%  & 26\%  & 56\%  \\
    \bottomrule
  \end{tabular}
\end{table}

Perov-5 contains only 5 distinct space groups, with a single space group (\#25, $Pmc2_1$) accounting for 56\% of the structures. This reflects the dataset's design as a collection of synthetic ABO$_3$-type perovskites that the dataset curates and constrains to specific symmetries, rather than a representative sample of materials space.

Carbon-24 spans 56 space groups but remains highly concentrated: the two most frequent space groups (\#1 and \#2, both triclinic) cover half of the dataset, and 84.2\% of structures belong to triclinic or monoclinic systems. This concentration is consistent with the dataset's origin as carbon polymorphs that ab initio random structure searching generated at 10~GPa \cite{pickard2020airss}, a regime that favors low-symmetry configurations.

MP-20 and MPTS-52 contain the most diverse symmetry distributions, with 179 and 208 unique space groups respectively, and long-tailed distributions in which the top 20 space groups cover only 65\% and 56\% of structures. The crystal system composition is also more balanced than in the other datasets, with all seven crystal systems appearing and no single system exceeding 26\% of the structures. The most frequent space group in MP-20 is \#225 ($Fm\bar{3}m$, cubic) at 14.8\%, while space group \#62 ($Pnma$, orthorhombic) dominates MPTS-52 at 9.0\%; the difference reflects the broader compositional scope of MPTS-52, which includes structures with up to 52 atoms per cell and consequently lower-symmetry arrangements.

The diversity gap between Perov-5 and the materials-project-derived datasets has direct implications for SICSP. CSPML retrieves space group templates by composition-based similarity to training structures: in Perov-5, where only 5 space groups exist and the top 5 cover the entire dataset, retrieval is essentially constrained to a small set of plausible templates. In MP-20 and MPTS-52, retrieval must select from 179 to 208 distinct space groups, with no top-20 list exceeding 65\% coverage; the probability that the retrieved template matches the correct symmetry therefore decreases substantially. This asymmetry partially explains the dataset-dependent gap between CSPML and ground truth template performance that Section~\ref{results_main} of the main paper reports.

\subsection{Species-aware alignment}
\label{app:species_alignment}

The alignment procedure that Section~\ref{sec:alignment} describes parametrizes the rotation as a unit quaternion $q \in \mathbb{R}^4$ with $\|q\|_2 = 1$ and the translation as $t \in \mathbb{R}^3$. The cost matrix assigns
\[
C_{ij} = \|R(q)x_i + t - x^*_j\|_2
\]
for same-species pairs and $+\infty$ otherwise. The Hungarian algorithm~\citep{kuhn1955hungarian} then solves the resulting assignment problem. We update $(q,t)$ with Adam (learning rate 0.1) on the RMSE under the current assignment, and recompute the assignment after every gradient step. We run 400 steps from each of 4 random initializations of $q$, with $t$ initialized at zero, and keep the transform with the lowest RMSE across all steps and initializations.

\begin{figure}[H]
    \centering
    \includegraphics[width=\linewidth]{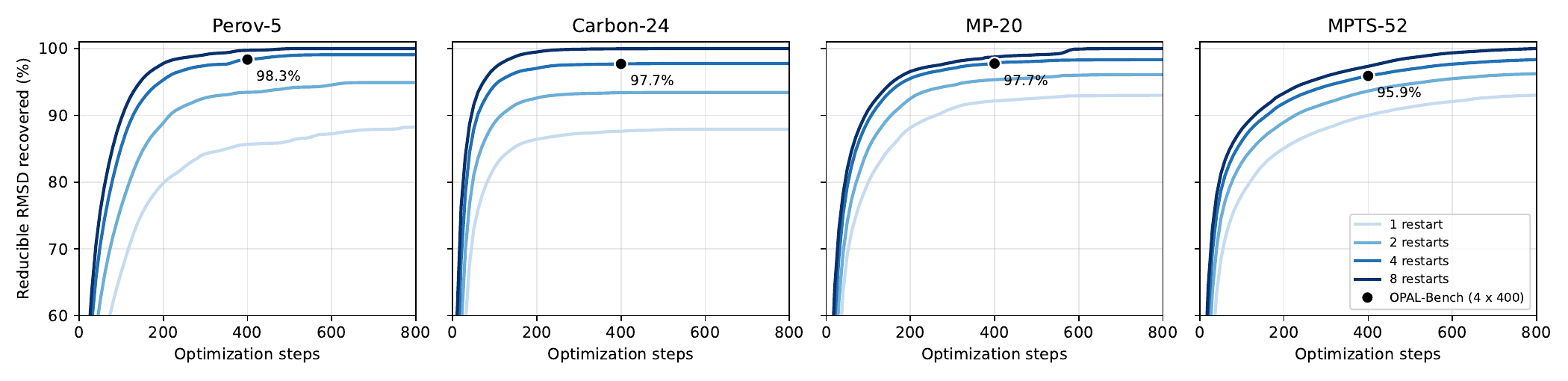}
    \caption{\textbf{Convergence of the species-aware alignment.} Fraction of the reduction from the unaligned RMSE to the lowest RMSE found with 8 initializations of 800 steps that each budget recovers, averaged over 300 pairs per dataset. Each line corresponds to a number of random initializations, and the black point marks the setting that \benchname{} uses (4 initializations of 400 steps).}
    \label{fig:alignment_convergence}
\end{figure}

We chose this budget as a trade-off between computational cost and alignment quality. To characterize its convergence, we sample 300 pairs of a prediction and one of its reference polymorphs per dataset, drawn across the seven CSP methods, and rerun the procedure with 8 random initializations of 800 steps each. For each pair, we take the lowest RMSE found with this extended budget as the reference, and the RMSE under the identity rotation and zero translation as the unaligned value. We then measure, for each smaller budget, the fraction of the gap between the unaligned value and the reference that the alignment closes. Figure~\ref{fig:alignment_convergence} shows that most of the reduction occurs within the first 200 steps and that additional initializations improve the alignment more than additional steps. Our setting of 4 initializations and 400 steps recovers 98.3\%, 97.7\%, 97.7\%, and 95.9\% of the reduction on Perov-5, Carbon-24, MP-20, and MPTS-52, respectively, with MPTS-52, whose unit cells are the largest, still improving slowly beyond 400 steps. Since a suboptimal alignment can only increase a distance, the geometric metrics that we report are upper bounds on their values under the optimal alignment, and the mean RMSE under our setting lies 2 to 4\% above the lowest RMSE found with the extended budget. As convergence slows with increasing unit cell size, which the MPTS-52 curves illustrate, datasets with larger or more complex structures may require a larger alignment budget, and we recommend repeating this analysis before applying \benchname{} to them.

\subsection{OPAL-CSP implementation details}
\label{app:opal_implementation}

\subsubsection{Architecture}
\label{app:opal_architecture}

OPAL-CSP builds on EquiformerV2~\citep{equiformer_v2} and models crystal structures as labeled point clouds evolving under a flow matching objective. We represent the chemical composition as a 3D graph $\mathcal{G} = (\mathcal{V}, \mathcal{E})$ and the lattice as a latent vector $z_{\text{lat}} \in \mathbb{R}^{d_{\text{cell}}}$.

Each vertex $i \in \mathcal{V}$ carries a block-structured feature
\begin{equation}
    h_i = \left\{ H_{i,\ell} \in \mathbb{R}^{(2\ell+1) \times C} \right\}_{\ell=0}^{\ell_{\max}},
\end{equation}
where the $(2\ell+1)$ rows index spherical-harmonic components that transform equivariantly under rotations and the $C$ columns store learnable channels that we initialize from the atomic number $a_i$. We follow the default values of $\ell_{\max}$ and $C$ from~\citet{equiformer_v2}, and set $d_{\text{cell}} = 128$. We encode the invariant polar lattice representation $\kappa_t \in \mathbb{R}^6$ into $z_{\text{lat}} \in \mathbb{R}^{d_{\text{cell}}}$ using a two-layer MLP and concatenate it to every node feature before processing.

We construct edges between each atom and up to $K$ nearest neighbours within cutoff radius $r_c$, using periodic boundary conditions. For each edge $(i,j)$, we compute the relative geometry $v_{ij} = x_j - x_i$, expand the distance $d_{ij} = \|v_{ij}\|_2$ with a Gaussian radial basis $\phi(d_{ij}) \in \mathbb{R}^B$, and project the direction $\hat{u}_{ij} = v_{ij}/d_{ij}$ through spherical harmonics $Y^{(\ell)}(\hat{u}_{ij}) \in \mathbb{R}^{2\ell+1}$ to produce SE(3)-equivariant messages.

We process the graph through $N$ equivariant Transformer layers. Each layer applies an attention block that performs irrep-aware message passing using eSCN convolutions with attention renormalization, and then applies a feed-forward block with separable $S^2$ activation and separable layer normalization. We embed the time step $t$ via sinusoidal features and use it to condition each normalization through adaptive separable layer normalization, which predicts scale, shift, and gating parameters.

\begin{figure}[H]
    \centering
    \includegraphics[width=0.9\linewidth]{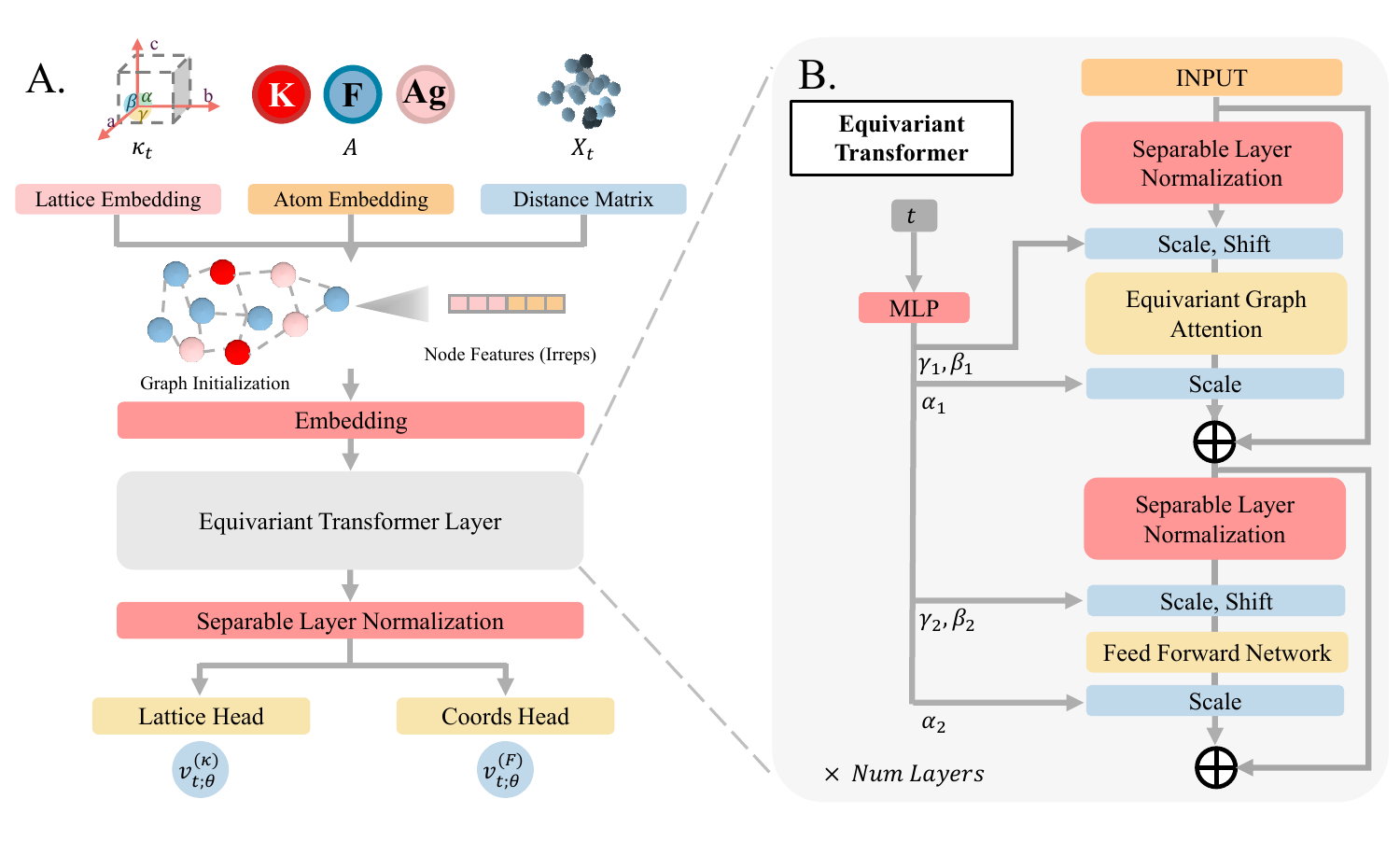}
    \caption{\textbf{OPAL-CSP architecture.} (A) Overview of the full model. The chemical composition is encoded into atomic and lattice embeddings which, together with the distance matrix, initialize a 3D graph with SE(3)-equivariant irreducible representations. The graph is processed through a stack of equivariant Transformer layers followed by separable layer normalization. Two output heads predict the flow matching velocity fields for the lattice $v^{(\kappa)}_{t;\theta}$ and fractional coordinates $v^{(F)}_{t;\theta}$. (B) Equivariant Transformer block. Each block applies SE(3)-equivariant graph attention via eSCN convolutions, followed by a feed-forward network, with residual connections and adaptive separable layer normalization conditioned on the time step.}
    \label{fig:opal_csp_architecture}
\end{figure}

Two output heads produce the velocity fields $v^{(\kappa)}_{t;\theta}$ and $v^{(F)}_{t;\theta}$. The lattice head averages node features and maps them to the 6-dimensional invariant polar space, while the coordinate head maps node features to the $3 \times N$ fractional coordinate space. Table~\ref{tab:data_specific_hparams} reports dataset-specific architectural hyperparameters, Table~\ref{tab:hparams_shared} reports shared hyperparameters, and Figure~\ref{fig:opal_csp_architecture} illustrates the architecture.

\begin{table}[h]
    \centering
    \caption{\textbf{Dataset-specific OPAL-CSP hyperparameters.}
    We report the graph-construction and architecture hyperparameters that vary across datasets: radial cutoff $r_c$ and number of layers $N$.}
    \label{tab:data_specific_hparams}
    \begin{tabular}{lcccc}
        \toprule
        & Perov-5 & Carbon-24 & MP-20 & MPTS-52 \\
        \midrule
        Cutoff $r_c$ (\AA) & 10 & 5 & 12 & 12 \\
        Layers $N$ & 3 & 6 & 6 & 6 \\
        \bottomrule
    \end{tabular}
\end{table}

\begin{table}[H]
    \centering
    \caption{\textbf{Shared OPAL-CSP hyperparameters.}}
    \label{tab:hparams_shared}
    \begin{tabular}{lc}
        \toprule
        Hyperparameter & Value \\
        \midrule
        Max degree $\ell_{\max}$ & 6 \\
        Sphere channels $C$ & 128 \\
        Max neighbours $K$ & 20 \\
        Attention heads & 8 \\
        Attention hidden channels & 64 \\
        FFN hidden channels & 128 \\
        Gaussian RBF basis $B$ & 600 \\
        Time embedding dim & 128 \\
        Lattice embedding dim $d_{\text{cell}}$ & 128 \\
        \bottomrule
    \end{tabular}
\end{table}

\subsubsection{Training and inference}
\label{app:opal_training}

We train OPAL-CSP with a flow matching regression objective that jointly supervises lattice and fractional coordinate velocity fields,
\begin{equation}
    \mathcal{L}_{\mathrm{FM}} =
    \mathbb{E}\!\left[
    \lambda_\beta \left\| v^{(\kappa)}_{t;\theta} - u^{(\kappa)}_t \right\|_2^2
    + \lambda_\alpha
    \left\| v^{(F)}_{t;\theta} - u^{(F)}_t \right\|_2^2
    \right],
\end{equation}
with $\lambda_\beta = 1$ (lattice) and $\lambda_\alpha = 10$ (fractional coordinates). The lattice follows a linear interpolation path $\kappa_t = (1-t)\kappa_0 + t\kappa_1$, and the fractional coordinates follow a wrapped path
\[
F_t = F_0 + t \cdot \mathrm{wrap}(F_1 - F_0),
\]
where $\mathrm{wrap}(x) = (x + \frac{1}{2}) \bmod 1 - \frac{1}{2}$, following CrystalFlow~\citep{luo2025crystalflow}.

Prior to training, we estimate the training-set mean $\bar{\kappa}_5$ of the fifth polar component and shift the lattice source distribution by this offset ($\kappa_0^{(5)} \leftarrow \kappa_0^{(5)} + \bar{\kappa}_5$), so that the expected velocity for each polar component is centred near zero and all six components contribute equally to the lattice loss gradient.

We train with the Adam optimizer using a learning rate of $10^{-3}$ and a reduce-on-plateau scheduler with factor $0.6$ and patience $30$, monitoring a scale-normalized validation loss $\mathcal{L}_{\text{coords}} / \mathcal{L}^0_{\text{coords}} + \mathcal{L}_{\text{lat}} / \mathcal{L}^0_{\text{lat}}$, where we estimate the zero-predictor baselines $\mathcal{L}^0$ from the training set before training so that both tasks contribute equally to the scheduler signal regardless of their natural scale difference. All datasets use a total batch size of $128$ structures across GPUs, and we train all models for $1000$ epochs. All experiments use random seed 42. We selected dataset specific hyperparameters (Table \ref{tab:data_specific_hparams}) through a limited number of exploratory runs with \benchname{} in view, rather than a systematic search or formal ablation.

At inference, we integrate the learned ODE from $t=0$ to $t=1$ using an Euler solver with $1000$ steps. At each step, we scale the predicted coordinate velocity by an annealing factor $\gamma(t) = 1 + 5\,\mathrm{ReLU}(t)$ before integration, progressively amplifying coordinate updates as $t \to 1$. We adopt this annealing schedule from CrystalFlow~\citep{luo2025crystalflow} without modification. We did not tune the sampling procedure against OPAL-Bench for OPAL-CSP or for any baseline, each of which uses the inference configuration released by its authors. Alternative sampling or integration schemes could improve the results of OPAL-CSP and of the baselines alike, and we leave their exploration to future work.

\subsection{Baseline implementation details}
\label{app:baseline_implementation}

We use the official implementations and configurations that the authors released for the baseline methods whenever available. We preserve the original training, sampling, and post-processing procedures for each method to avoid introducing implementation-dependent advantages or disadvantages. We restrict any deviations from the original implementations to dataset formatting, checkpoint loading, and evaluation-interface standardization that we need to run all methods under the same OPAL-Bench protocol.

\subsection{Computational resources}
\label{app:computational_resources}

We train all models on NVIDIA Quadro RTX 8000 GPUs (48 GB memory), using between 1 and 7 GPUs depending on the model and dataset. Metric computation runs on 2× Intel Xeon Gold 6230 CPUs (40 physical cores, 2.10 GHz). Training \modelname{} takes 15h 5m on Perov-5 and 22h 1m on Carbon-24 using 2 GPUs, and 2d 5h 5m on MP-20 and 4d 3h 22m on MPTS-52 using 4 GPUs.    

\section{Complete Results}
\label{app:complete_results}

\subsection{Validity rates and intersection sizes}
\label{app:validity_rates}

For CSP, every metric has a well-defined value for all seven methods on 100\% of test compositions on Perov-5 and Carbon-24, and on at least 99.8\% on MP-20 and MPTS-52 (at least 8{,}705 of 8{,}719 and 7{,}867 of 7{,}878 compositions, respectively), so we do not report per-method CSP validity separately. The few excluded compositions correspond to predictions for which a metric cannot be computed, mostly XRD patterns of predictions with degenerate unit cells, and we evaluate each metric on the intersection of compositions where all seven methods have a valid value. For SICSP, validity depends jointly on the method, the metric, and the dataset, since some metrics require structural or symmetry properties that not every prediction satisfies. Table~\ref{tab:sicsp_validity} reports validity rates for each method, metric, and dataset across the four SICSP method–template combinations, together with the intersection for each metric and dataset, the fraction of compositions for which all four combinations yielded a valid prediction for that metric. We compute all SICSP results in this paper on these intersections, so for any given metric we score all four methods on identical compositions. Validity is essentially complete on Perov-5 and Carbon-24. On MP-20 and MPTS-52, SGEquiDiff produces a small fraction of predictions with degenerate lattice geometries that we exclude from the affected metrics; on MPTS-52, formation-energy metrics additionally exclude a few DiffCSP++ (CSPML) predictions whose relaxations failed to converge.

\begin{table}[H]
\centering
\caption{\textbf{Validity rates and evaluation-set intersections for SICSP methods.} 
Each row reports the percentage of test formulas with valid predictions available for each metric.
Results are shown for DiffCSP++ and SGEquiDiff under both CSPML-predicted and ground truth (GT) symmetry templates. 
The \textit{Intersection} row in each dataset block reports the fraction of formulas for which all listed method--template combinations produced valid predictions for that metric, defining the common evaluation set used for comparison.}
\label{tab:sicsp_validity}
\resizebox{\linewidth}{!}{%
\begin{tabular}{llrrrrrrrrr}
\toprule
 & \textbf{Model} & \textbf{Chamf.} & \textbf{Haus.} & \textbf{Sink.} & \textbf{RMSE} & \textbf{$\rho$} & \textbf{SOAP} & \textbf{XRD} & \textbf{FE} & \textbf{R.\ FE} \\
\midrule
\multirow{5}{*}{\rotatebox[origin=c]{90}{\scalebox{0.75}{\textsc{Perov-5}}}} & DiffCSP++ (CSPML) & 100.0\% & 100.0\% & 100.0\% & 100.0\% & 100.0\% & 100.0\% & 100.0\% & 100.0\% & 100.0\% \\
 & SGEquiDiff (CSPML) & 100.0\% & 100.0\% & 100.0\% & 100.0\% & 99.9\% & 99.9\% & 99.9\% & 100.0\% & 99.9\% \\
 & DiffCSP++ (GT) & 100.0\% & 100.0\% & 100.0\% & 100.0\% & 100.0\% & 100.0\% & 100.0\% & 100.0\% & 100.0\% \\
 & SGEquiDiff (GT) & 99.9\% & 99.9\% & 99.9\% & 99.9\% & 99.9\% & 99.9\% & 99.9\% & 99.9\% & 99.9\% \\
\cmidrule(lr){2-11}
 & \textit{Intersection} & 99.8\% & 99.8\% & 99.8\% & 99.8\% & 99.8\% & 99.8\% & 99.8\% & 99.8\% & 99.8\% \\
\midrule
\multirow{5}{*}{\rotatebox[origin=c]{90}{\scalebox{0.75}{\textsc{Carbon-24}}}} & DiffCSP++ (CSPML) & 100.0\% & 100.0\% & 100.0\% & 100.0\% & 100.0\% & 100.0\% & 100.0\% & 100.0\% & 100.0\% \\
 & SGEquiDiff (CSPML) & 100.0\% & 100.0\% & 100.0\% & 100.0\% & 100.0\% & 100.0\% & 100.0\% & 100.0\% & 100.0\% \\
 & DiffCSP++ (GT) & 100.0\% & 100.0\% & 100.0\% & 100.0\% & 100.0\% & 100.0\% & 100.0\% & 100.0\% & 100.0\% \\
 & SGEquiDiff (GT) & 100.0\% & 100.0\% & 100.0\% & 100.0\% & 100.0\% & 100.0\% & 100.0\% & 100.0\% & 100.0\% \\
\cmidrule(lr){2-11}
 & \textit{Intersection} & 100.0\% & 100.0\% & 100.0\% & 100.0\% & 100.0\% & 100.0\% & 100.0\% & 100.0\% & 100.0\% \\
\midrule
\multirow{5}{*}{\rotatebox[origin=c]{90}{\scalebox{0.75}{\textsc{MP-20}}}} & DiffCSP++ (CSPML) & 100.0\% & 100.0\% & 100.0\% & 100.0\% & 100.0\% & 100.0\% & 100.0\% & 100.0\% & 100.0\% \\
 & SGEquiDiff (CSPML) & 98.1\% & 98.1\% & 98.1\% & 98.1\% & 99.6\% & 98.4\% & 98.1\% & 98.1\% & 98.1\% \\
 & DiffCSP++ (GT) & 100.0\% & 100.0\% & 100.0\% & 100.0\% & 100.0\% & 100.0\% & 100.0\% & 100.0\% & 100.0\% \\
 & SGEquiDiff (GT) & 99.1\% & 99.1\% & 99.1\% & 99.1\% & 99.1\% & 99.1\% & 99.1\% & 99.1\% & 99.1\% \\
\cmidrule(lr){2-11}
 & \textit{Intersection} & 97.4\% & 97.4\% & 97.4\% & 97.4\% & 98.7\% & 97.7\% & 97.4\% & 97.4\% & 97.4\% \\
\midrule
\multirow{5}{*}{\rotatebox[origin=c]{90}{\scalebox{0.75}{\textsc{MPTS-52}}}} & DiffCSP++ (CSPML) & 100.0\% & 100.0\% & 100.0\% & 100.0\% & 100.0\% & 100.0\% & 100.0\% & 99.6\% & 99.5\% \\
 & SGEquiDiff (CSPML) & 95.1\% & 95.1\% & 95.1\% & 95.1\% & 95.1\% & 95.0\% & 95.1\% & 95.0\% & 95.0\% \\
 & DiffCSP++ (GT) & 100.0\% & 100.0\% & 100.0\% & 100.0\% & 100.0\% & 100.0\% & 100.0\% & 100.0\% & 100.0\% \\
 & SGEquiDiff (GT) & 95.4\% & 95.4\% & 95.4\% & 95.4\% & 95.4\% & 95.4\% & 95.4\% & 95.2\% & 95.2\% \\
\cmidrule(lr){2-11}
 & \textit{Intersection} & 91.3\% & 91.3\% & 91.3\% & 91.3\% & 91.3\% & 91.1\% & 91.3\% & 90.6\% & 90.6\% \\
\bottomrule
\end{tabular}%
}
\end{table}

\subsection{95\% confidence intervals}
\label{app:confidence_intervals}
\begin{figure}[H]
    \centering
    \includegraphics[width=1.0\linewidth]{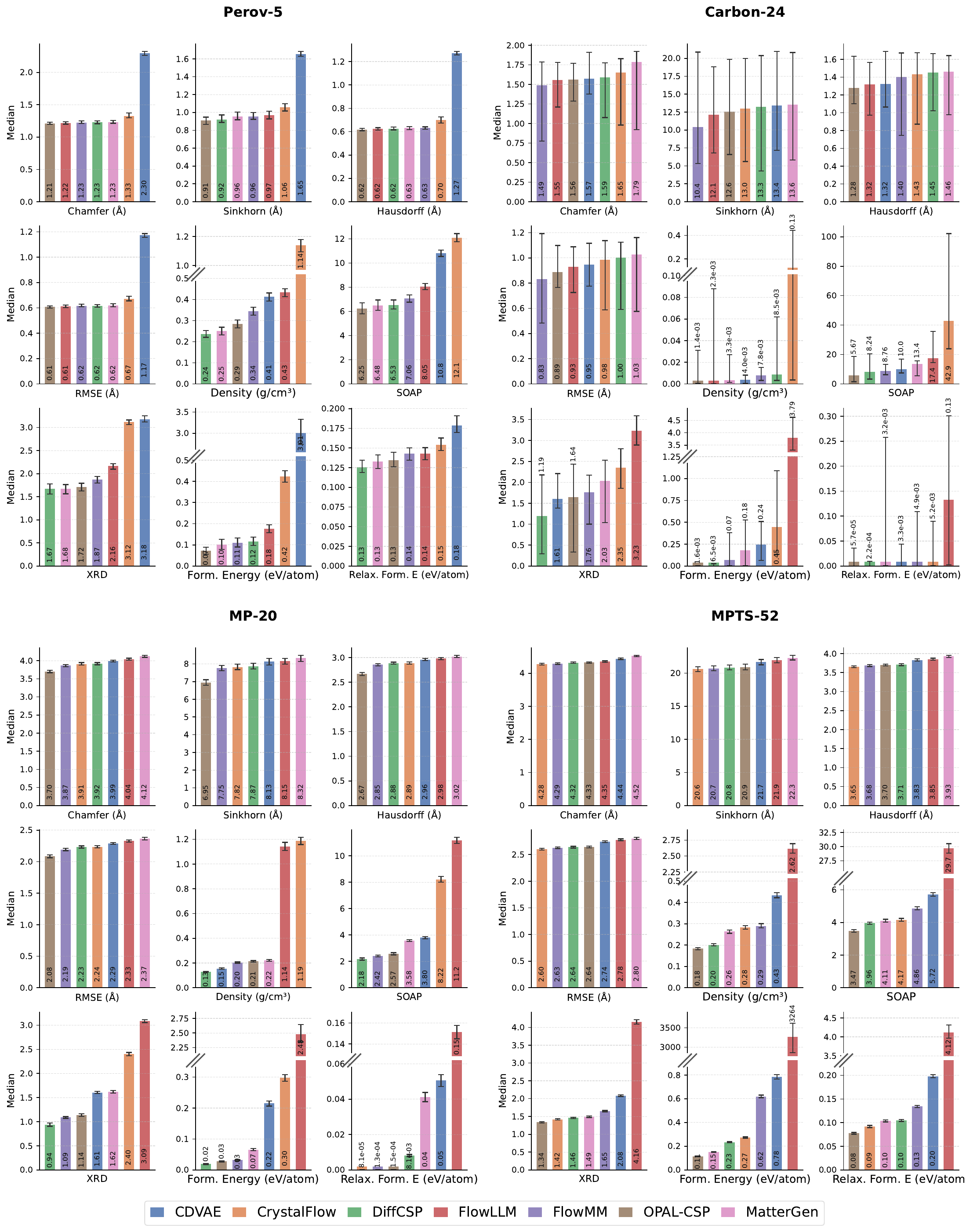}
    \caption{\textbf{Full CSP results across all four datasets and nine metrics.} Bars report the median over test compositions and error bars are 95\% bootstrap confidence intervals from $10{,}000$ resampling iterations. Lower is better for all metrics. Physical metrics reveal rankings that differ substantially from geometric metrics, confirming that geometric performance does not reliably imply physical plausibility.}
    \label{fig:csp_ci_supertable}
\end{figure}

\begin{figure}[H]
    \centering
    \includegraphics[width=1.0\linewidth]{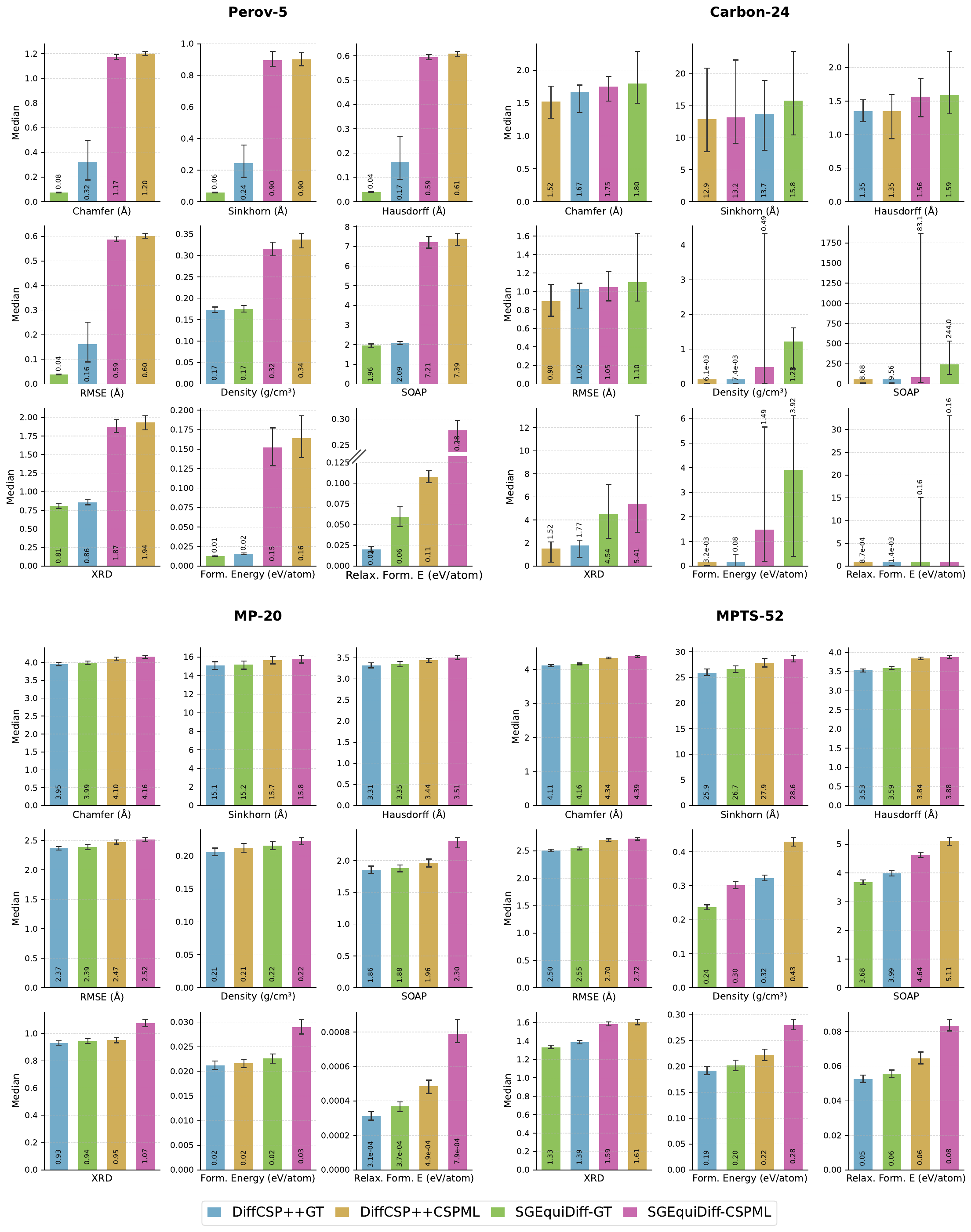}
    \caption{\textbf{Full SICSP results across all four datasets and nine metrics, for both ground truth (GT) and CSPML-predicted symmetry templates.} Bars report the median over test compositions on the shared intersection of valid predictions and error bars are 95\% bootstrap confidence intervals. GT templates yield substantially better geometric performance on Perov-5, where symmetry constraints are highly informative, while Carbon-24 is a notable exception where CSPML templates outperform GT due to the high polymorphic diversity of carbon structures.}
    \label{fig:sicsp_ci_supertable}
\end{figure}

Figures~\ref{fig:csp_ci_supertable} and~\ref{fig:sicsp_ci_supertable} extend Tables \ref{tab:csp_results} and \ref{tab:sicsp_results} with bar charts covering all metric-dataset combinations. Error bars report 95\% bootstrap confidence intervals that we compute via $10{,}000$ resampling iterations over test compositions. Non-overlapping intervals indicate statistically significant differences between methods.

On Perov-5, the structural homogeneity of perovskites yields tight confidence intervals and clear separations between methods on most metrics. Carbon-24 produces wide intervals throughout, consistent with the high structural variance of carbon polymorphs, which makes conclusive rankings difficult for this dataset. MP-20, as the largest dataset, serves as the primary reference for cross-method comparison, though rankings here vary more across metrics than on any other dataset. MPTS-52, despite its structural complexity, yields the tightest confidence intervals and the most consistent rankings within each metric family, making it the most reliable setting for definitive conclusions.

\subsection{Qualitative Results}
\label{app:qualitative_results}

\begin{figure}[H]
    \centering
    \includegraphics[width=1\linewidth]{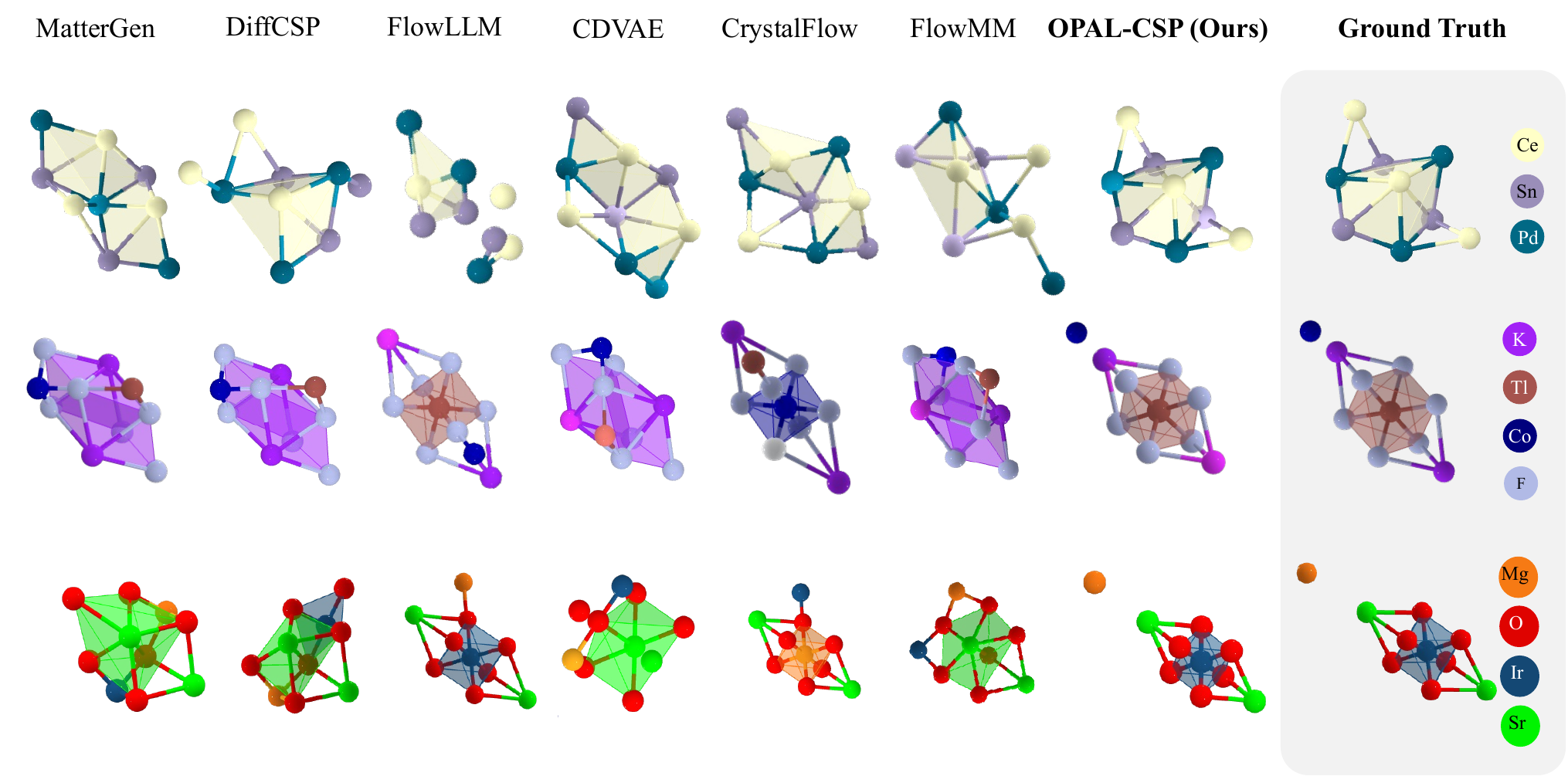}
    \caption{\textbf{Qualitative comparison on representative samples from the MP-20 dataset in the CSP task.} Predictions from the seven CSP methods are shown alongside the corresponding ground truth structures, rendered as ball-and-stick models within their unit cells. Each row corresponds to one test composition, and atom species are color-coded consistently across methods and ground truth.}
    \label{fig:qualitative_1}
\end{figure}

\begin{figure}[H]
    \centering
    \includegraphics[width=1\linewidth]{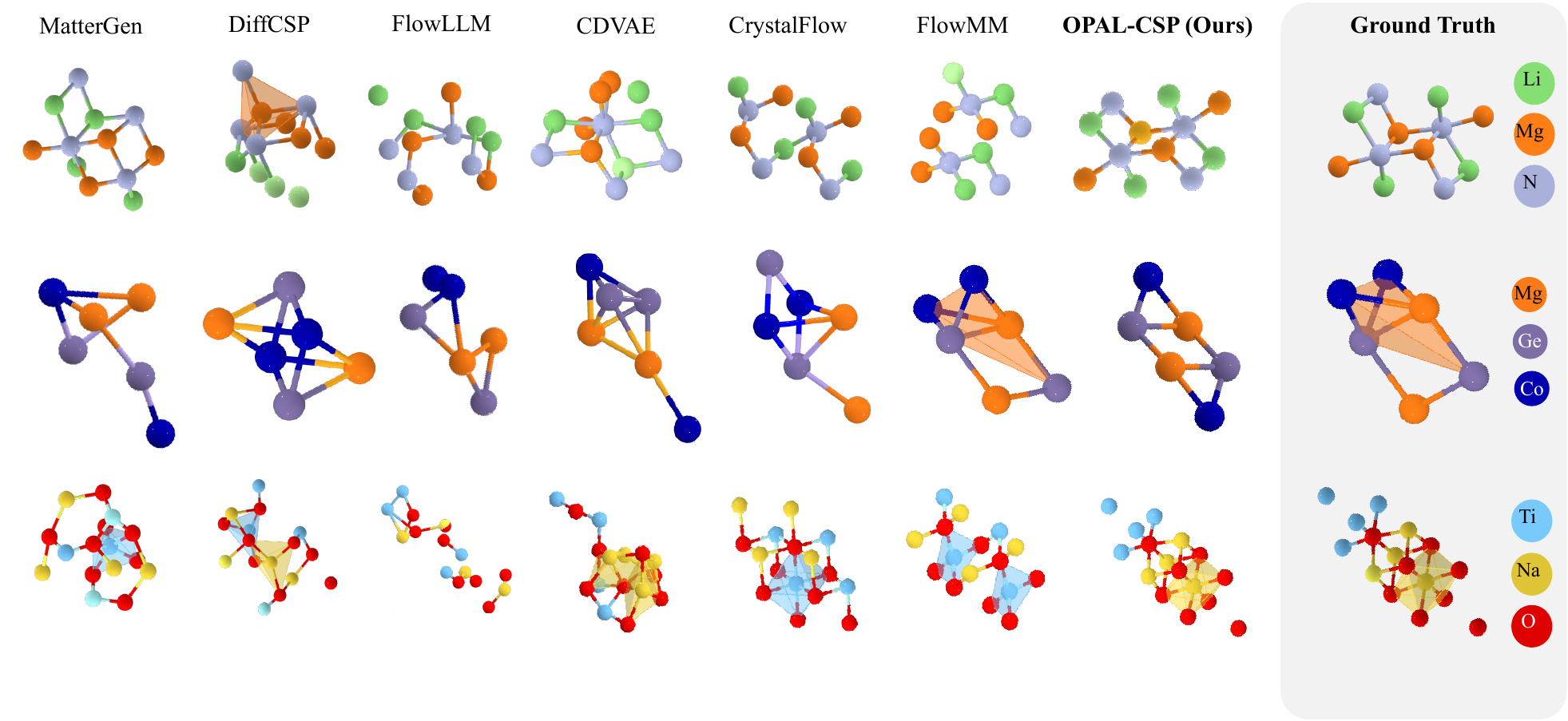}
    \caption{\textbf{Qualitative comparison on representative samples from the MPTS-52 dataset in the CSP task.} Predictions from the seven CSP methods are shown alongside the corresponding ground truth structures, rendered as ball-and-stick models within their unit cells. Each row corresponds to one test composition, and atom species are color-coded consistently across methods and ground truth.}
    \label{fig:qualitative_2}
\end{figure}

\begin{figure}[H]
    \centering
    \includegraphics[width=1\linewidth]{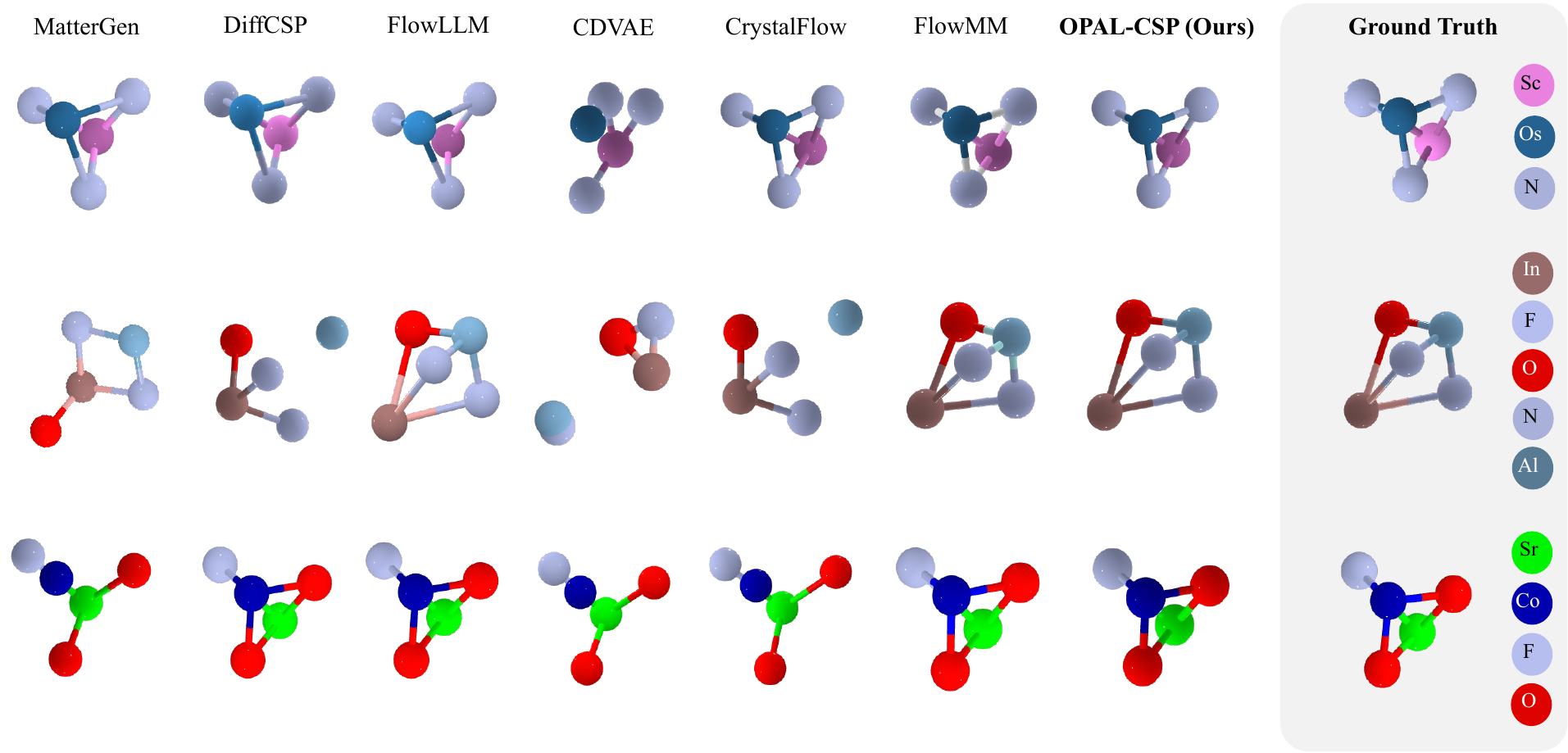}
    \caption{\textbf{Qualitative comparison on representative samples from the Perov-5 dataset in the CSP task.} Predictions from the seven CSP methods are shown alongside the corresponding ground truth structures, rendered as ball-and-stick models within their unit cells. Each row corresponds to one test composition, and atom species are color-coded consistently across methods and ground truth.}
    \label{fig:qualitative_3}
\end{figure}

\begin{figure}[H]
    \centering
    \includegraphics[width=1\linewidth]{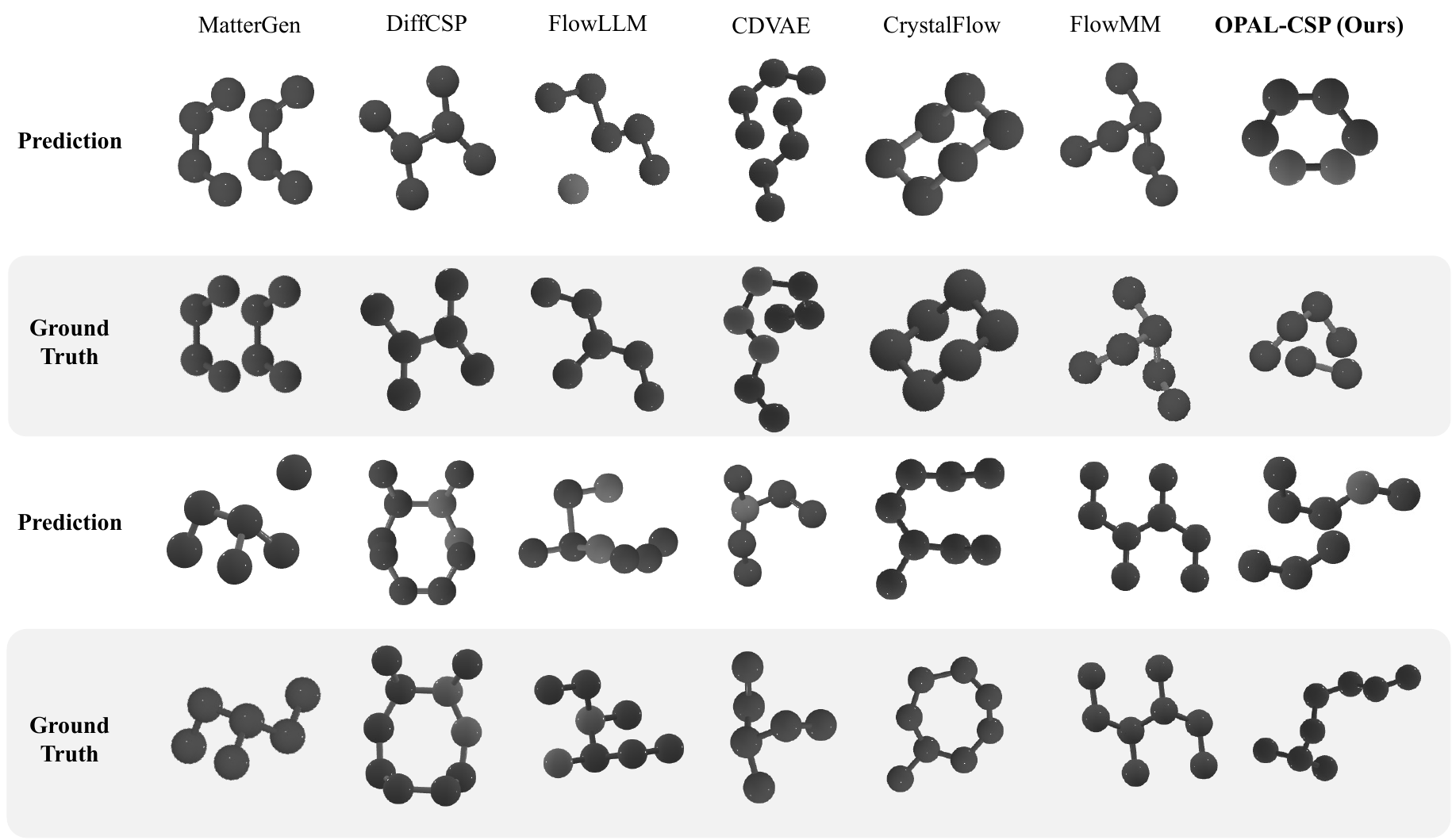}
    \caption{\textbf{Qualitative comparison on representative samples from the Carbon-24 dataset in the CSP task.} Predictions from the seven CSP methods are shown alongside their corresponding selected reference polymorphs. Given the high number of polymorphs in Carbon-24, each prediction is visualized with its closest selected reference polymorph rather than a single shared ground truth structure across all columns. Structures are rendered as ball-and-stick models within their unit cells. Each prediction--reference pair corresponds to one test composition.}
    \label{fig:qualitative_4}
\end{figure}

\begin{figure}[H]
    \centering
    \includegraphics[width=1\linewidth]{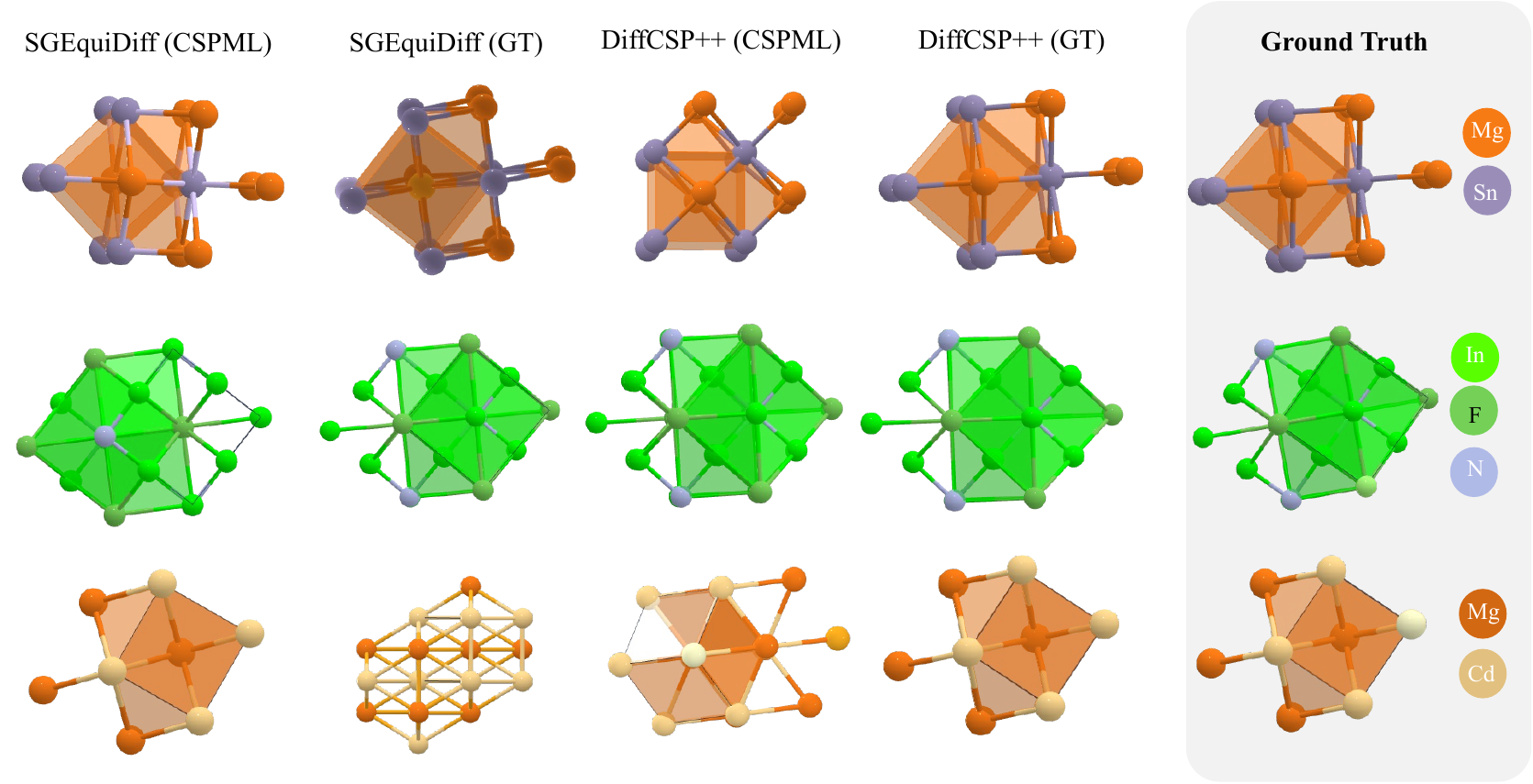}
    \caption{\textbf{Qualitative comparison on representative samples from the MP-20 dataset in the SICSP task.} Predictions from the two SICSP methods are shown under both CSPML-predicted and ground truth symmetry templates alongside the corresponding ground truth structures, rendered as ball-and-stick models within their unit cells. Each row corresponds to one test composition, and atom species are color-coded consistently across methods, template settings, and ground truth.}
    \label{fig:qualitative_5}
\end{figure}

\begin{figure}[H]
    \centering
    \includegraphics[width=1\linewidth]{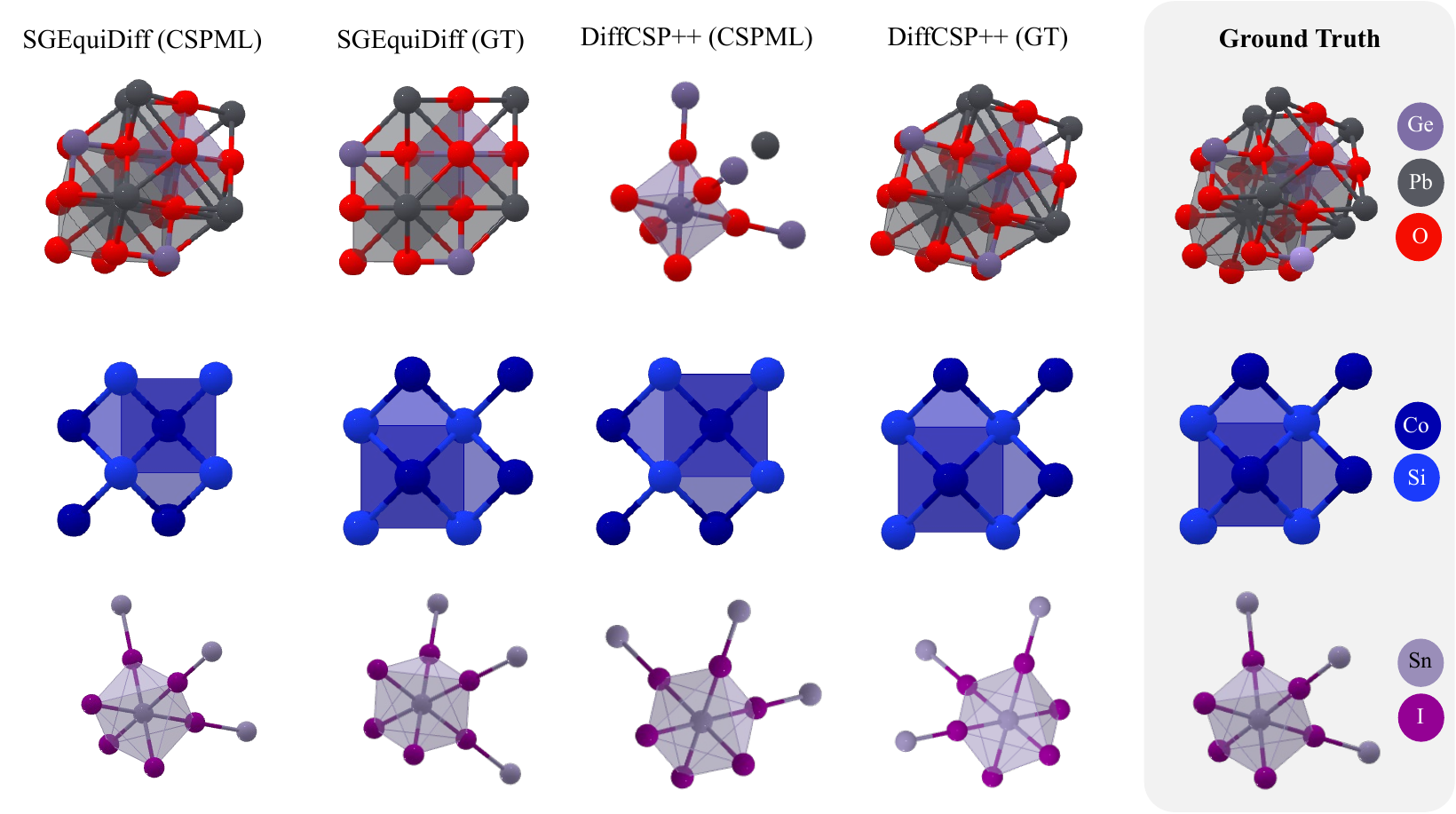}
    \caption{\textbf{Qualitative comparison on representative samples from the MPTS-52 dataset in the SICSP task.} Predictions from the two SICSP methods are shown under both CSPML-predicted and ground truth symmetry templates alongside the corresponding ground truth structures, rendered as ball-and-stick models within their unit cells. Each row corresponds to one test composition, and atom species are color-coded consistently across methods, template settings, and ground truth.}
    \label{fig:qualitative_6}
\end{figure}

\begin{figure}[H]
    \centering
    \includegraphics[width=0.95\linewidth]{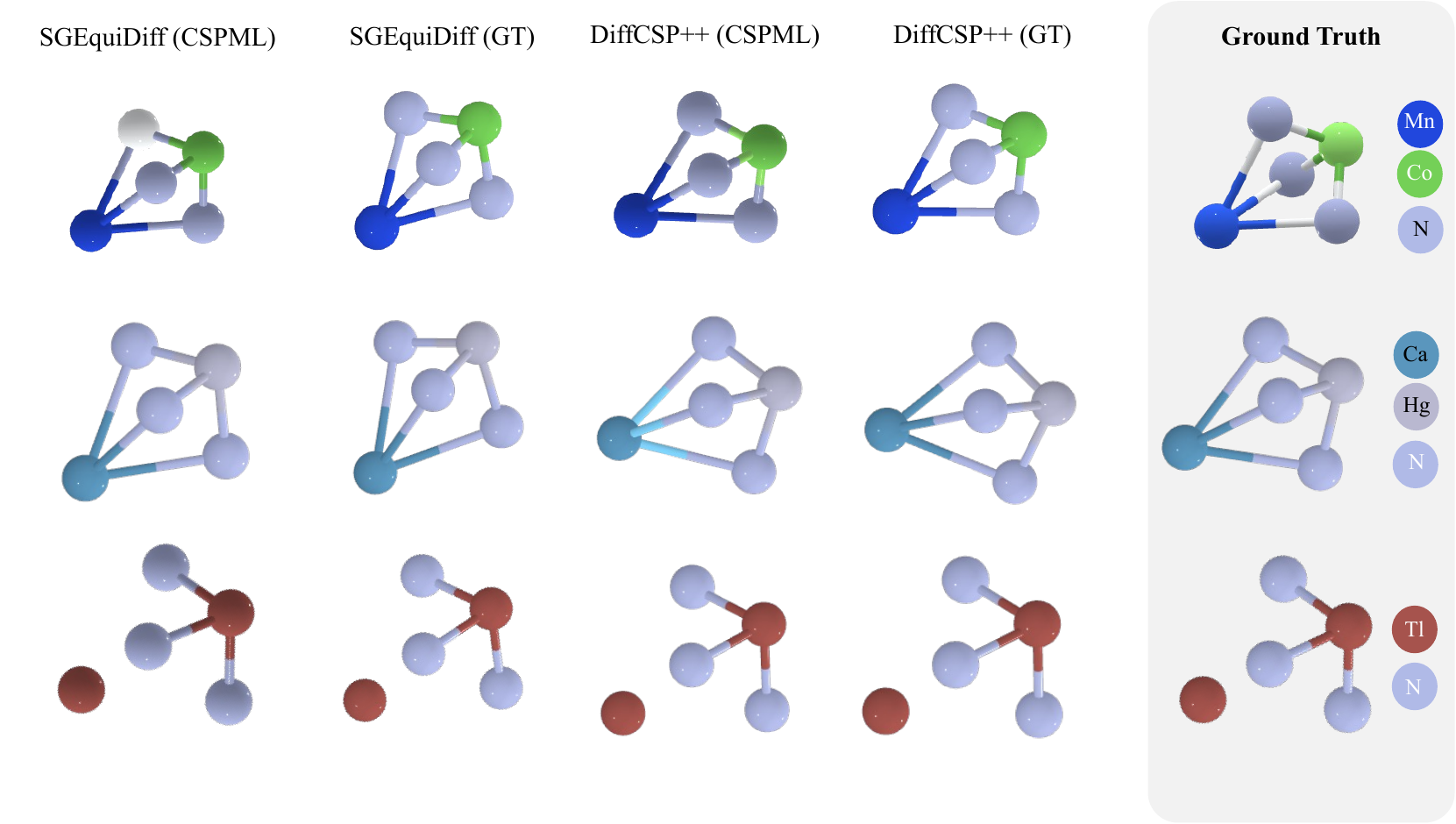}
    \caption{\textbf{Qualitative comparison on representative samples from the Perov-5 dataset in the SICSP task.} Predictions from the two SICSP methods are shown under both CSPML-predicted and ground truth symmetry templates alongside the corresponding ground truth structures, rendered as ball-and-stick models within their unit cells. Each row corresponds to one test composition, and atom species are color-coded consistently across methods, template settings, and ground truth.}
    \label{fig:qualitative_7}
\end{figure}

\begin{figure}[H]
    \centering
    \includegraphics[width=0.95\linewidth]{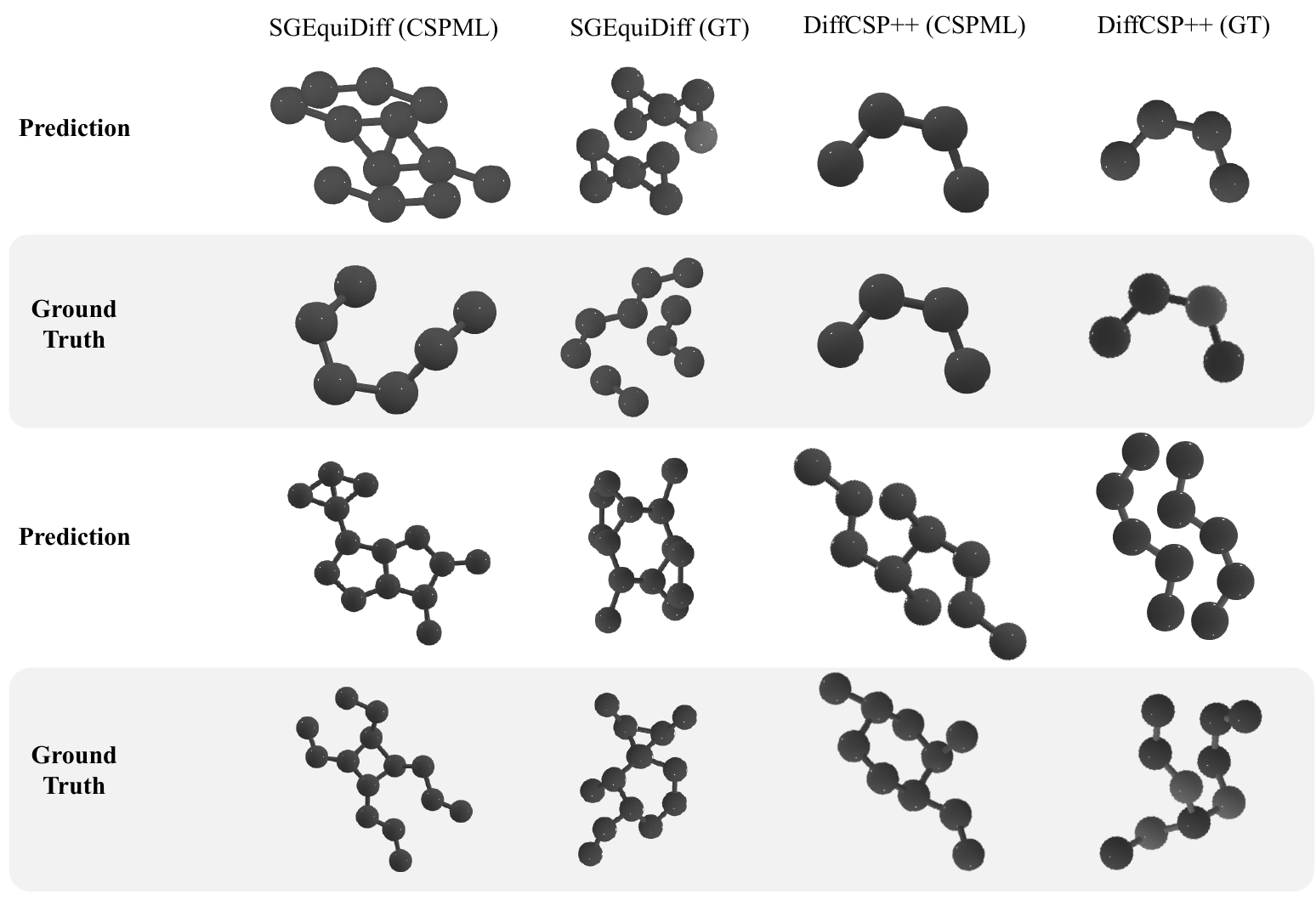}
        \caption{\textbf{Qualitative comparison on representative samples from the Carbon-24 dataset in the SICSP task.} We show predictions from SGEquiDiff and DiffCSP++ in their CSPML and GT variants, alongside their corresponding selected reference polymorphs. Given the high number of polymorphs in Carbon-24, each prediction is visualized with its closest selected reference polymorph rather than a single shared ground truth structure across all columns. Structures are rendered as ball-and-stick models within their unit cells. Each prediction--reference pair corresponds to one test composition.}
    \label{fig:qualitative_8}
\end{figure}

\begin{figure}[H]
    \centering
    \includegraphics[width=0.9\linewidth]{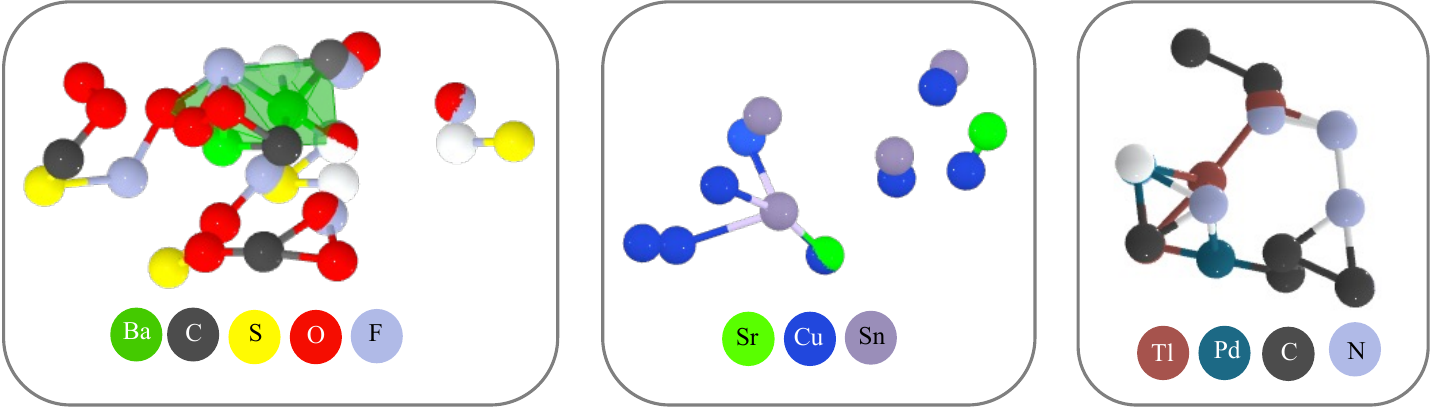}
    \caption{\textbf{Failure modes of FlowLLM predictions.} Representative structures generated by FlowLLM exhibit superposed atomic species at near-coincident positions, producing unphysical short-range overlaps. These geometric artifacts explain the anomalously large Formation Energy distances reported in Table~\ref{tab:csp_results} (e.g., 3264.4 eV/atom on MPTS-52, 3.79 eV/atom on Carbon-24), since pairwise repulsion diverges as interatomic distances approach zero.}
    \label{fig:flowLLM_outliers}
\end{figure}

Figures~\ref{fig:qualitative_1} to~\ref{fig:qualitative_4} show representative crystal structure predictions from all seven CSP methods, rendered alongside the corresponding ground truth structures as ball-and-stick models within their unit cells. Figures~\ref{fig:qualitative_5} to~\ref{fig:qualitative_8} present the analogous qualitative comparisons for the SICSP track, covering SGEquiDiff and DiffCSP++ under both CSPML-predicted and ground truth symmetry templates. Each row corresponds to one test composition, and atom species are color-coded consistently across method predictions and ground truth. Because symmetrization can alter the atom count, SICSP predictions may contain a different number of atoms than the ground truth; we show the raw outputs. Figure~\ref{fig:flowLLM_outliers} illustrates the source of FlowLLM's anomalously high formation energy distances; predicted structures contain atomic species at near-coincident positions, producing the unphysically large formation energies reflected in Table~\ref{tab:csp_results}.

\section{Extended Analysis}

\subsection{Geometric–physical correlation analysis}
\label{app:correlation_analysis}

\begin{figure}[H]
    \centering
    \includegraphics[width=0.95\linewidth]{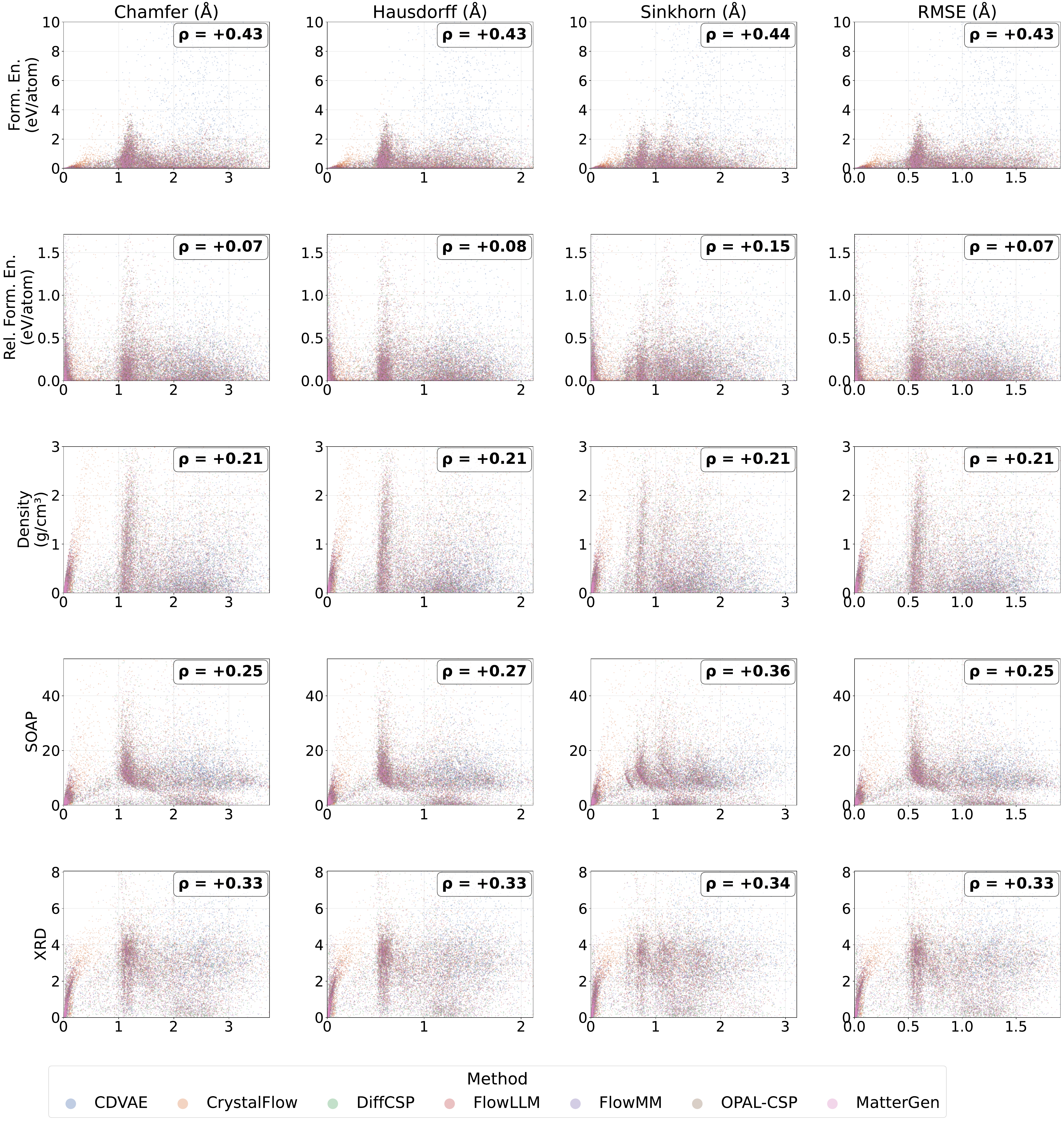}
    \caption{\textbf{Spearman correlation between geometric metrics (columns) and physical metrics (rows) across all CSP predictions in Perov-5.} Each point is a single prediction, colored by method, restricted to the 3{,}033 test structures with no polymorph in the test split (i.e., a unique reference per prediction). The low correlation coefficients $\rho$ confirm that geometric performance is weakly correlated to physical plausibility, motivating the inclusion of physical metrics in OPAL-Bench.}
    \label{fig:geo_vs_phys_perov5}
\end{figure}

\begin{figure}[H]
    \centering
    \includegraphics[width=0.95\linewidth]{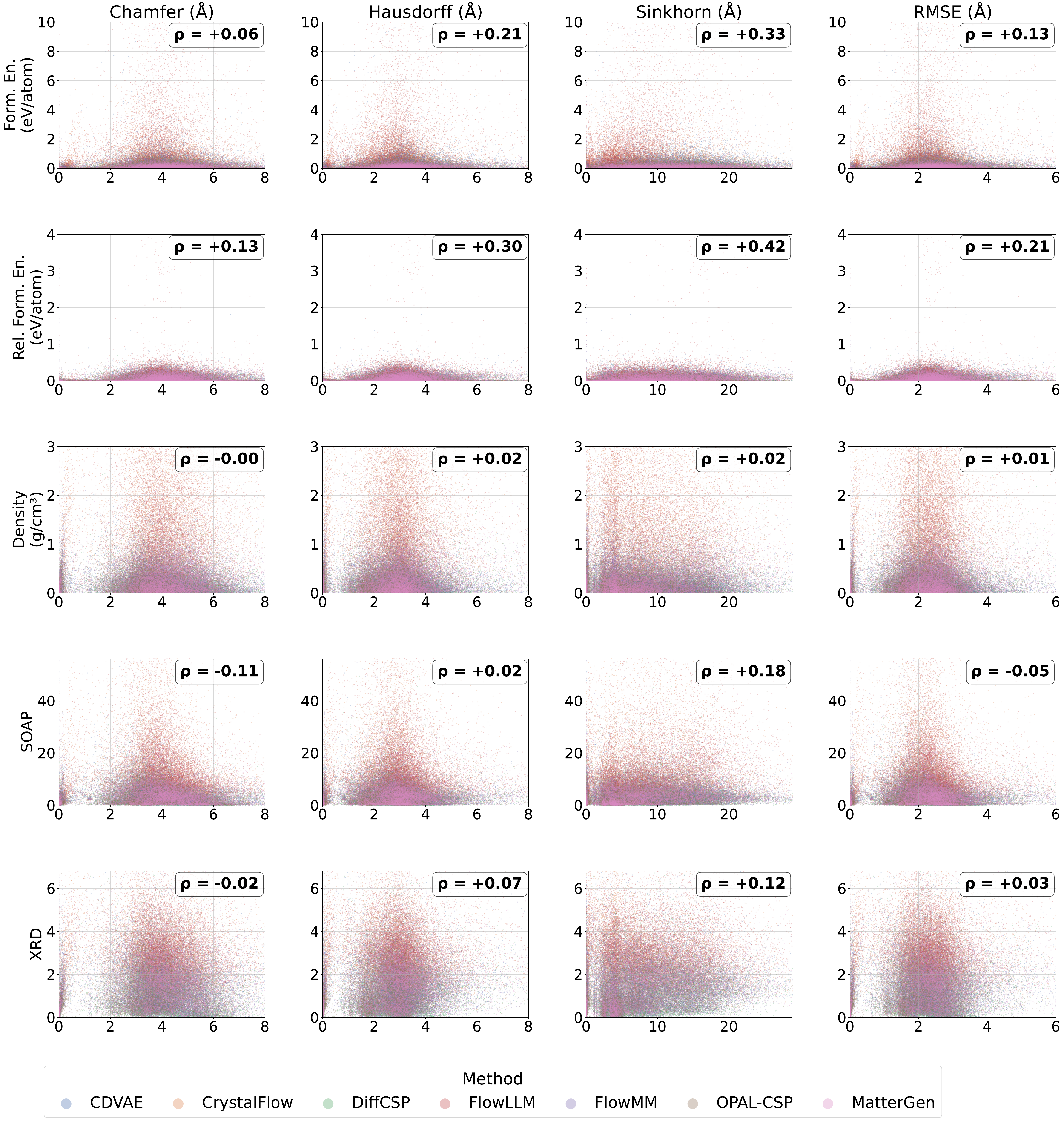}
    \caption{\textbf{Spearman correlation between geometric metrics (columns) and physical metrics (rows) across all CSP predictions in MP-20.} Each point is a single prediction, colored by method, restricted to the 8{,}472 test structures with no polymorph in the test split (i.e., a unique reference per prediction). The low correlation coefficients $\rho$ confirm that geometric performance is weakly correlated to physical plausibility, motivating the inclusion of physical metrics in OPAL-Bench.}
    \label{fig:geo_vs_phys_mp20}
\end{figure}

\begin{figure}[H]
    \centering
    \includegraphics[width=0.95\linewidth]{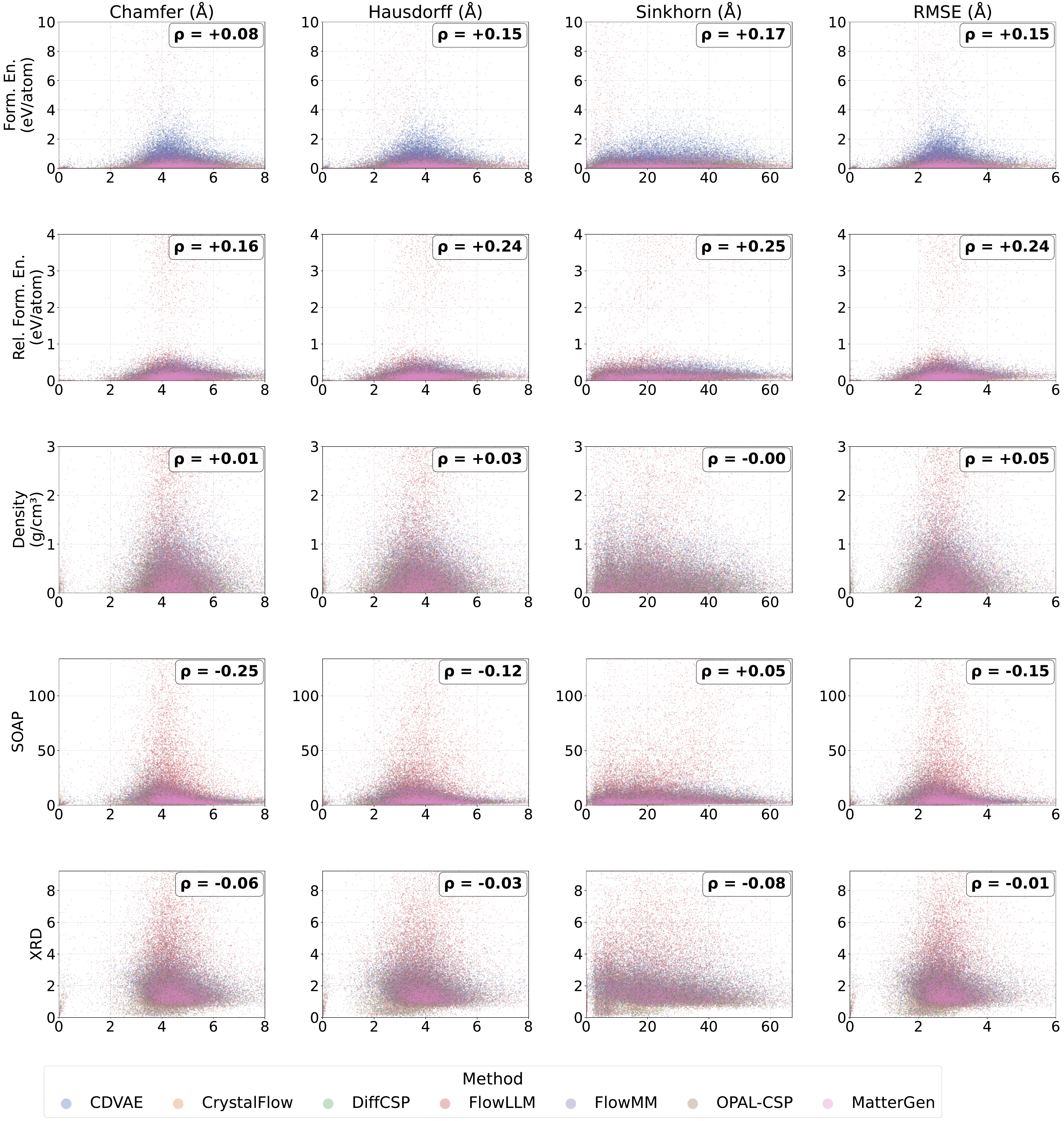}
    \caption{\textbf{Spearman correlation between geometric metrics (columns) and physical metrics (rows) across all CSP predictions in MPTS-52.} Each point is a single prediction, colored by method, restricted to the 7{,}688 test structures with no polymorph in the test split (i.e., a unique reference per prediction). The low correlation coefficients $\rho$ confirm that geometric performance is weakly correlated to physical plausibility, motivating the inclusion of physical metrics in OPAL-Bench.}
    \label{fig:geo_vs_phys}
\end{figure}

Since each metric selects its own closest polymorph, the geometric and physical metrics of a single prediction may be computed against different reference structures, which could in principle weaken their correlation. To control for this effect, Figures~\ref{fig:geo_vs_phys_perov5}, \ref{fig:geo_vs_phys_mp20}, and~\ref{fig:geo_vs_phys} plot each physical metric against each geometric metric for all CSP predictions in Perov-5, MP-20, and MPTS-52 for the subset of structures with no polymorphs in the test split (3{,}033, 8{,}472, and 7{,}688 compositions, respectively), so that all metrics share the same reference structure. The Spearman correlation coefficients $\rho$ in each panel confirm the weak relationship between geometric and physical performance.

Formation energy and relaxed formation energy show positive but mostly weak correlations with all four geometric metrics ($\rho \in [0.07, 0.44]$ on Perov-5, $[0.06, 0.42]$ on MP-20, and $[0.08, 0.25]$ on MPTS-52), indicating that lower geometric error is weakly predictive of better energetic quality. Among the geometric metrics, Sinkhorn distance shows the highest correlation with the energy metrics on all three datasets, reaching 0.44 with formation energy on Perov-5, 0.42 with relaxed formation energy on MP-20, and 0.25 with relaxed formation energy on MPTS-52, although its margin over the other geometric metrics is small on Perov-5 and MPTS-52. We hypothesize that this reflects its transport formulation, which weighs the displacement of all atoms jointly rather than relying on nearest-neighbor distances, as Chamfer does, or on the largest deviation, as Hausdorff does.

On MP-20 and MPTS-52, density and XRD show correlations near zero ($\rho \in [-0.02, 0.12]$ and $[-0.08, 0.05]$, respectively), meaning geometric accuracy provides essentially no information about unit cell volume quality or diffraction pattern similarity. Correlations are consistently highest on Perov-5, where all twenty pairs are positive ($\rho = 0.21$ for density and $0.33$ for XRD), and weaken as structural complexity grows from Perov-5 to MP-20 and MPTS-52. We hypothesize that in simple structures an accurate prediction tends to be both geometrically and physically close to the reference, whereas in more complex structures this coupling is diluted. SOAP fingerprint distance shows weak negative correlations with Chamfer and Hausdorff distances on MPTS-52 ($\rho = -0.25$ and $-0.12$, respectively), but this sign is not universal, as $\rho$(SOAP, Chamfer) is $-0.11$ on MP-20 and $+0.25$ on Perov-5. We hypothesize that this sign tracks how widely density varies within each dataset, since Chamfer distance shrinks as atoms pack more closely, whereas denser packing places more neighbors within the fixed SOAP cutoff radius and inflates the descriptor distance. These results substantiate the central finding that geometric and physical quality are weakly correlated.

Reference divergence, where the geometric and physical metrics of a prediction select different polymorphs, can only occur for compositions with more than one known polymorph. Within that subset it is frequent, affecting 30 to 44\% of predictions on Perov-5, 52 to 55\% on MP-20, and 48 to 54\% on MPTS-52, but since few compositions have more than one polymorph in these datasets (Table~\ref{tab:polymorph_summary}), it affects only 4.9\%, 1.5\%, and 1.3\% of all predictions, respectively. Including or excluding these multi-polymorph predictions barely changes the correlations, with shifts of at most 0.006 on MP-20 and 0.005 on MPTS-52 across all twenty pairs. On Perov-5, including them raises the correlations slightly (by 0.036 on average and at most 0.051), so reference divergence, where it occurs, does not inflate the apparent decorrelation. Carbon-24 is the exception, since every composition in its test split has multiple known polymorphs, which leaves no single-polymorph subset for this analysis and makes reference divergence nearly universal (91 to 100\% of predictions). Its values in Tables~\ref{tab:csp_results} and~\ref{tab:sicsp_results} should therefore be read with caution, and we analyze alternative reference-selection rules in Appendix~\ref{app:polymorph_count}.


\subsection{OPAL-Bench ablations}
\label{app:opal_bench_ablations}

Tables~\ref{tab:ablation_no_polymorphs},~\ref{tab:ablation_no_alignment}, and~\ref{tab:ablation_no_polymorphs_no_alignment} report scores under three ablated protocol variants. Since physical metrics are SE(3)-invariant, they are unaffected by alignment and reported only in Table~\ref{tab:ablation_no_polymorphs}.

Removing polymorph handling (Table~\ref{tab:ablation_no_polymorphs}) has a modest effect on Perov-5, MP-20, and MPTS-52, consistent with the low polymorph counts per composition in those datasets. The effect is most pronounced on Carbon-24, where the large number of known polymorphs per formula makes the choice of reference consequential. Methods that recover a valid but non-canonical carbon polymorph face inflated distances under the single-reference protocol even when their prediction is physically correct.

Removing SE(3) alignment (Table~\ref{tab:ablation_no_alignment}) inflates all geometric distances substantially across datasets and methods. On Perov-5, Chamfer distances increase by roughly $4\times$ relative to the full protocol and method rankings become less discriminative as rotational and permutation ambiguities dominate the signal. This indicates that alignment is necessary to recover true structural divergence.

Removing both components simultaneously (Table~\ref{tab:ablation_no_polymorphs_no_alignment}) combines these effects, producing the most inflated and least discriminative scores. Together, these results indicate that alignment and polymorph handling are complementary components of a fair evaluation, with alignment necessary on every dataset and polymorph handling becoming critical as the number of known polymorphs per composition grows, as on Carbon-24.

\begin{table}[H]
\centering
\caption{\textbf{Ablation without polymorph handling.} 
We report geometric and physical distances for the protocol variant in which each prediction is evaluated against a single reference structure rather than the closest known polymorph. 
The SE(3) alignment step is retained for geometric metrics, while physical metrics are computed directly from the predicted and reference structures. 
Values are median scores over test compositions; lower is better for all metrics. 
\textbf{Bold} marks the best value and \underline{underline} marks the second-best value for each dataset and metric.}
\label{tab:ablation_no_polymorphs}
\setlength{\tabcolsep}{4pt}

\begin{minipage}{\linewidth}
\centering
\textsc{Perov-5}\\[4pt]
\resizebox{0.75\linewidth}{!}{%
\begin{tabular}{lrrrrrrr}
\toprule
\textbf{Model} & \textbf{Chamf.}$\downarrow$ & \textbf{Haus.}$\downarrow$ & \textbf{Sink.}$\downarrow$ & \textbf{RMSE}$\downarrow$ & \textbf{$\rho$ (g/cm$^3$)}$\downarrow$ & \textbf{SOAP}$\downarrow$ & \textbf{XRD}$\downarrow$ \\
\midrule
FlowLLM     & \underline{1.252} & \underline{0.641} & 1.041          & \underline{0.629} & 0.471          & 8.610          & 2.291 \\
CDVAE       & 2.325          & 1.288          & 1.681          & 1.186          & 0.447          & 11.065         & 3.260 \\
MatterGen   & 1.273          & 0.651          & 1.029          & 0.638          & \underline{0.298} & \underline{7.406} & \textbf{1.873} \\
CrystalFlow & 1.393          & 0.741          & 1.119          & 0.701          & 1.178          & 12.440         & 3.178 \\
DiffCSP     & 1.262          & 0.650          & \underline{0.997} & 0.635       & \textbf{0.280} & 7.445          & 1.927 \\
FlowMM      & 1.265          & 0.651          & 1.024          & 0.635          & 0.381          & 7.710          & 2.037 \\
\midrule
\modelname{} & \textbf{1.245} & \textbf{0.636} & \textbf{0.987} & \textbf{0.625} & 0.322 & \textbf{7.386} & \underline{1.922} \\
\bottomrule
\end{tabular}}
\end{minipage}

\vspace{8pt}
\begin{minipage}{\linewidth}
\centering
\textsc{Carbon-24}\\[4pt]
\resizebox{0.75\linewidth}{!}{%
\begin{tabular}{lrrrrrrr}
\toprule
\textbf{Model} & \textbf{Chamf.}$\downarrow$ & \textbf{Haus.}$\downarrow$ & \textbf{Sink.}$\downarrow$ & \textbf{RMSE}$\downarrow$ & \textbf{$\rho$ (g/cm$^3$)}$\downarrow$ & \textbf{SOAP}$\downarrow$ & \textbf{XRD}$\downarrow$ \\
\midrule
FlowLLM     & \underline{2.035} & 1.843          & 17.322          & 1.241          & 0.359          & 96.851          & 5.303 \\
CDVAE       & 2.131          & 1.883          & \textbf{15.910} & 1.268          & 0.180          & 50.454          & 5.241 \\
MatterGen   & 2.142          & 2.004          & 17.498          & 1.275          & 0.276          & 79.957          & 4.362 \\
CrystalFlow & 2.107          & 1.940          & 17.565          & \underline{1.223} & 0.730       & 181.251         & 5.923 \\
DiffCSP     & \textbf{2.002} & 2.085          & 18.653          & 1.291          & \textbf{0.123} & \textbf{28.349} & \underline{3.857} \\
FlowMM      & 2.055          & \textbf{1.657} & \underline{16.044} & \textbf{1.209} & \underline{0.134} & \underline{40.201} & \textbf{3.241} \\
\midrule
\modelname   & 2.188          & \underline{1.785} & 17.106       & 1.268          & 0.604          & 156.024         & 7.113 \\
\bottomrule
\end{tabular}}
\end{minipage}

\vspace{8pt}
\begin{minipage}{\linewidth}
\centering
\textsc{MP-20}\\[4pt]
\resizebox{0.75\linewidth}{!}{%
\begin{tabular}{lrrrrrrr}
\toprule
\textbf{Model} & \textbf{Chamf.}$\downarrow$ & \textbf{Haus.}$\downarrow$ & \textbf{Sink.}$\downarrow$ & \textbf{RMSE}$\downarrow$ & \textbf{$\rho$ (g/cm$^3$)}$\downarrow$ & \textbf{SOAP}$\downarrow$ & \textbf{XRD}$\downarrow$ \\
\midrule
FlowLLM     & 4.051          & 2.990          & 8.196          & 2.338          & 1.144          & 11.169         & 3.098 \\
CDVAE       & 3.998          & 2.970          & 8.180          & 2.297          & \underline{0.156} & 3.818          & 1.613 \\
MatterGen   & 4.130          & 3.035          & 8.358          & 2.372          & 0.223          & 3.604          & 1.628 \\
CrystalFlow & 3.929          & 2.896          & 7.866          & 2.245          & 1.190          & 8.245          & 2.408 \\
DiffCSP     & 3.934          & 2.894          & 7.911          & 2.242          & \textbf{0.129} & \textbf{2.219} & \textbf{0.949} \\
FlowMM      & \underline{3.878} & \underline{2.866} & \underline{7.801} & \underline{2.200} & 0.205     & \underline{2.434} & \underline{1.093} \\
\midrule
\modelname{} & \textbf{3.720} & \textbf{2.687} & \textbf{7.004} & \textbf{2.097} & 0.215 & 2.591 & 1.144 \\
\bottomrule
\end{tabular}}
\end{minipage}

\vspace{8pt}
\begin{minipage}{\linewidth}
\centering
\textsc{MPTS-52}\\[4pt]
\resizebox{0.75\linewidth}{!}{%
\begin{tabular}{lrrrrrrr}
\toprule
\textbf{Model} & \textbf{Chamf.}$\downarrow$ & \textbf{Haus.}$\downarrow$ & \textbf{Sink.}$\downarrow$ & \textbf{RMSE}$\downarrow$ & \textbf{$\rho$ (g/cm$^3$)}$\downarrow$ & \textbf{SOAP}$\downarrow$ & \textbf{XRD}$\downarrow$ \\
\midrule
FlowLLM     & 4.357          & 3.857          & 21.964          & 2.782          & 2.619          & 29.702         & 4.160 \\
CDVAE       & 4.445          & 3.838          & 21.686          & 2.745          & 0.435          & 5.728          & 2.089 \\
MatterGen   & 4.525          & 3.937          & 22.347          & 2.808          & 0.264          & 4.122          & 1.496 \\
CrystalFlow & \textbf{4.283} & \textbf{3.662} & \textbf{20.691} & \textbf{2.607} & 0.286     & 4.173          & \underline{1.423} \\
DiffCSP     & 4.326          & 3.714          & 20.849          & 2.645          & \underline{0.202} & \underline{3.964} & 1.466 \\
FlowMM      & \underline{4.299} & \underline{3.691} & \underline{20.721} & \underline{2.635} & 0.292       & 4.881          & 1.652 \\
\midrule
\modelname   & 4.330          & 3.706          & 20.975          & 2.647          & \textbf{0.184} & \textbf{3.486} & \textbf{1.341} \\
\bottomrule
\end{tabular}}
\end{minipage}

\end{table}
\newpage
\begin{table}[H]
\centering
\caption{\textbf{Ablation without SE(3) alignment.} 
Geometric distances are reported for the protocol variant that removes the SE(3) alignment step while retaining polymorph handling. 
For each prediction, the closest reference polymorph is still selected, but distances are computed without aligning the predicted and reference structures. 
Values are median scores over test compositions; lower is better for all metrics. 
\textbf{Bold} marks the best value and \underline{underline} marks the second-best value for each dataset and metric.}
\label{tab:ablation_no_alignment}
\setlength{\tabcolsep}{4pt}
\begin{minipage}[t]{0.48\linewidth}
\centering
\smallskip
\textsc{Perov-5}\\[4pt]
\resizebox{\linewidth}{!}{%
\begin{tabular}{lrrrr}
\toprule
\textbf{Model} & \textbf{Chamf.}$\downarrow$ & \textbf{Haus.}$\downarrow$ & \textbf{Sink.}$\downarrow$ & \textbf{RMSE}$\downarrow$ \\
\midrule
FlowLLM    & \underline{5.004} & \underline{2.768} & 3.785          & 2.665          \\
CDVAE      & 5.045          & 2.811          & 3.774          & \underline{2.650} \\
MatterGen  & 5.066          & 2.814          & 3.825          & 2.697          \\
CrystalFlow& 5.037          & 2.788          & 3.798          & 2.677          \\
DiffCSP    & 5.028          & 2.781          & 3.773          & 2.676          \\
FlowMM     & 5.013          & 2.782          & \underline{3.772} & 2.673       \\
\midrule
\modelname{} & \textbf{4.944} & \textbf{2.758} & \textbf{3.752} & \textbf{2.648} \\
\bottomrule
\end{tabular}}
\end{minipage}
\hfill
\begin{minipage}[t]{0.48\linewidth}
\centering
\smallskip
\textsc{Carbon-24}\\[4pt]
\resizebox{\linewidth}{!}{%
\begin{tabular}{lrrrr}
\toprule
\textbf{Model} & \textbf{Chamf.}$\downarrow$ & \textbf{Haus.}$\downarrow$ & \textbf{Sink.}$\downarrow$ & \textbf{RMSE}$\downarrow$ \\
\midrule
FlowLLM    & \underline{1.940} & \underline{1.684} & 16.298          & 1.336          \\
CDVAE      & 2.090          & 1.846          & \underline{15.893} & 1.389       \\
MatterGen  & 2.063          & 2.026          & 16.321          & 1.412          \\
CrystalFlow& 2.010          & 1.897          & 16.537          & 1.440          \\
DiffCSP    & 2.017          & 1.904          & 17.790          & 1.340          \\
FlowMM     & 2.009          & 1.694          & 16.797          & \underline{1.273} \\
\midrule
\modelname{} & \textbf{1.886} & \textbf{1.589} & \textbf{14.693} & \textbf{1.245} \\
\bottomrule
\end{tabular}}
\end{minipage}

\vspace{10pt}

\begin{minipage}[t]{0.48\linewidth}
\centering
\smallskip
\textsc{MP-20}\\[4pt]
\resizebox{\linewidth}{!}{%
\begin{tabular}{lrrrr}
\toprule
\textbf{Model} & \textbf{Chamf.}$\downarrow$ & \textbf{Haus.}$\downarrow$ & \textbf{Sink.}$\downarrow$ & \textbf{RMSE}$\downarrow$ \\
\midrule
FlowLLM    & 6.196          & 4.708          & 13.590          & 3.920          \\
CDVAE      & 5.829          & 4.295          & 12.384          & 3.594          \\
MatterGen  & 6.338          & 4.766          & 14.009          & 3.982          \\
CrystalFlow& 5.819          & 4.233          & 12.311          & 3.562          \\
DiffCSP    & 5.827          & 4.256          & 12.346          & 3.564          \\
FlowMM     & \underline{5.730} & \underline{4.157} & \underline{12.059} & \underline{3.486} \\
\midrule
\modelname{} & \textbf{5.313} & \textbf{4.006} & \textbf{11.125} & \textbf{3.356} \\
\bottomrule
\end{tabular}}
\end{minipage}
\hfill
\begin{minipage}[t]{0.48\linewidth}
\centering
\smallskip
\textsc{MPTS-52}\\[4pt]
\resizebox{\linewidth}{!}{%
\begin{tabular}{lrrrr}
\toprule
\textbf{Model} & \textbf{Chamf.}$\downarrow$ & \textbf{Haus.}$\downarrow$ & \textbf{Sink.}$\downarrow$ & \textbf{RMSE}$\downarrow$ \\
\midrule
FlowLLM    & 6.630          & 6.290          & 33.163          & 4.992          \\
CDVAE      & 5.901          & 5.419          & 30.002          & 4.233          \\
MatterGen  & 6.265          & 5.905          & 33.122          & 4.612          \\
CrystalFlow & \textbf{5.700} & \textbf{5.155} & \textbf{28.652} & \textbf{4.023} \\
DiffCSP    & \underline{5.759} & 5.277       & \underline{29.126} & 4.125       \\
FlowMM     & 5.774          & \underline{5.245} & 29.173       & \underline{4.073} \\
\midrule
\modelname{}  & 5.857          & 5.381          & 29.776          & 4.205          \\
\bottomrule
\end{tabular}}
\end{minipage}
\end{table}

\begin{table}[H]
\centering
\caption{\textbf{Ablation without polymorph handling or SE(3) alignment.} 
Geometric distances are reported for the protocol variant that removes both polymorph-aware reference selection and SE(3) alignment. 
Each prediction is evaluated against a single unaligned reference structure. 
Values are median scores over test compositions; lower is better for all metrics. 
\textbf{Bold} marks the best value and \underline{underline} marks the second-best value for each dataset and metric.}
\label{tab:ablation_no_polymorphs_no_alignment}
\setlength{\tabcolsep}{4pt}
\begin{minipage}[t]{0.48\linewidth}
\centering
\smallskip
\textsc{Perov-5}\\[4pt]
\resizebox{\linewidth}{!}{%
\begin{tabular}{lrrrr}
\toprule
\textbf{Model} & \textbf{Chamf.}$\downarrow$ & \textbf{Haus.}$\downarrow$ & \textbf{Sink.}$\downarrow$ & \textbf{RMSE}$\downarrow$ \\
\midrule
FlowLLM     & \underline{5.040} & \underline{2.786} & 3.802          & 2.683          \\
CDVAE       & 5.080          & 2.831          & 3.796          & \textbf{2.667} \\
MatterGen   & 5.111          & 2.837          & 3.853          & 2.716          \\
CrystalFlow & 5.070          & 2.812          & 3.819          & 2.696          \\
DiffCSP     & 5.065          & 2.799          & 3.793          & 2.698          \\
FlowMM      & 5.043          & 2.803          & \textbf{3.782} & 2.692          \\
\midrule
\modelname{} & \textbf{4.983} & \textbf{2.782} & \underline{3.784} & \underline{2.676} \\
\bottomrule
\end{tabular}}
\end{minipage}
\hfill
\begin{minipage}[t]{0.48\linewidth}
\centering
\smallskip
\textsc{Carbon-24}\\[4pt]
\resizebox{\linewidth}{!}{%
\begin{tabular}{lrrrr}
\toprule
\textbf{Model} & \textbf{Chamf.}$\downarrow$ & \textbf{Haus.}$\downarrow$ & \textbf{Sink.}$\downarrow$ & \textbf{RMSE}$\downarrow$ \\
\midrule
FlowLLM     & 2.895          & 2.950          & 26.017          & \textbf{2.222} \\
CDVAE       & \textbf{2.825} & \underline{2.916} & 27.738        & 2.596          \\
MatterGen   & \underline{2.860} & 2.952       & \underline{24.379} & \underline{2.243} \\
CrystalFlow & 2.889          & 3.189          & \textbf{23.069} & 2.399         \\
DiffCSP     & 3.474          & 3.665          & 29.502          & 2.819          \\
FlowMM      & 3.011          & \textbf{2.907} & 25.513          & 2.293          \\
\midrule
\modelname   & 3.135          & 3.254          & 25.008          & 2.644          \\
\bottomrule
\end{tabular}}
\end{minipage}

\vspace{10pt}

\begin{minipage}[t]{0.48\linewidth}
\centering
\smallskip
\textsc{MP-20}\\[4pt]
\resizebox{\linewidth}{!}{%
\begin{tabular}{lrrrr}
\toprule
\textbf{Model} & \textbf{Chamf.}$\downarrow$ & \textbf{Haus.}$\downarrow$ & \textbf{Sink.}$\downarrow$ & \textbf{RMSE}$\downarrow$ \\
\midrule
FlowLLM     & 6.217          & 4.732          & 13.658          & 3.936          \\
CDVAE       & 5.852          & 4.315          & 12.438          & 3.608          \\
MatterGen   & 6.365          & 4.788          & 14.090          & 4.001          \\
CrystalFlow & 5.838          & 4.250          & 12.374          & 3.576          \\
DiffCSP     & 5.849          & 4.278          & 12.438          & 3.585          \\
FlowMM      & \underline{5.751} & \underline{4.184} & \underline{12.143} & \underline{3.503} \\
\midrule
\modelname{} & \textbf{5.342} & \textbf{4.038} & \textbf{11.241} & \textbf{3.375} \\
\bottomrule
\end{tabular}}
\end{minipage}
\hfill
\begin{minipage}[t]{0.48\linewidth}
\centering
\smallskip
\textsc{MPTS-52}\\[4pt]
\resizebox{\linewidth}{!}{%
\begin{tabular}{lrrrr}
\toprule
\textbf{Model} & \textbf{Chamf.}$\downarrow$ & \textbf{Haus.}$\downarrow$ & \textbf{Sink.}$\downarrow$ & \textbf{RMSE}$\downarrow$ \\
\midrule
FlowLLM     & 6.642          & 6.303          & 33.306          & 5.005          \\
CDVAE       & 5.913          & 5.438          & 30.084          & 4.248          \\
MatterGen   & 6.282          & 5.926          & 33.200          & 4.635          \\
CrystalFlow & \textbf{5.717} & \textbf{5.182} & \textbf{28.737} & \textbf{4.037} \\
DiffCSP     & \underline{5.776} & 5.303        & \underline{29.258} & 4.141        \\
FlowMM      & 5.785          & \underline{5.255} & 29.287         & \underline{4.084} \\
\midrule
\modelname   & 5.874          & 5.400          & 29.911          & 4.221          \\
\bottomrule
\end{tabular}}
\end{minipage}
\end{table}

\subsection{Sensitivity of the physical metrics to descriptor settings}
\label{app:physical_sensitivity}

The SOAP and XRD metrics rely on fixed descriptor settings (Appendix~\ref{physical_appendix}). These are not matching tolerances, but they are design choices, so we test whether our conclusions depend on them. We use the MPTS-52 subset of compositions with a single known polymorph (7{,}688 compositions, 53{,}816 predictions across the seven CSP methods), so that all metrics share the same reference structure. We sweep SOAP over 47 configurations and XRD over 38, varying the parameters over the ranges in Table~\ref{tab:physical_sensitivity}, and recompute each metric on all predictions under every configuration.

\begin{table}[H]
  \centering
  \caption{\textbf{Sensitivity of per-structure rankings to descriptor settings on the single-polymorph subset of MPTS-52.} Each row sweeps one parameter. The last column reports the worst-case Spearman correlation between the per-structure values under the swept setting and under the default setting.}
  \label{tab:physical_sensitivity}
  \small
  \begin{tabular}{llccc}
    \toprule
    \textbf{Metric} & \textbf{Parameter} & \textbf{Range swept} & \textbf{Default} & \textbf{Worst-case $\rho$} \\
    \midrule
    SOAP & $r_\mathrm{cut}$ (\AA)            & 3.5--8.0 & 4.0 & 0.96 \\
    SOAP & $n_\mathrm{max}$                  & 4--13    & 12  & 0.95 \\
    SOAP & $l_\mathrm{max}$                  & 3--12    & 9   & 1.00 \\
    SOAP & $\sigma$ (\AA)                    & 0.2--1.0 & 0.5 & 0.89 \\
    XRD  & $2\theta_\mathrm{min}$ ($^\circ$) & 3--10    & 5   & 0.94 \\
    XRD  & $2\theta_\mathrm{max}$ ($^\circ$) & 60--120  & 90  & 0.96 \\
    \bottomrule
  \end{tabular}
\end{table}

For each configuration, we compute the Spearman correlation between its per-structure values and those of the default setting, and Table~\ref{tab:physical_sensitivity} reports the worst case over each sweep. Per-structure rankings stay close to the default for every parameter, with worst cases of 0.89 or higher for SOAP and 0.94 or higher for XRD. The largest change comes from the SOAP Gaussian width $\sigma$ (0.89).

The weak relationship between geometric and physical performance persists across these settings. Across all 85 configurations, the mean absolute Spearman correlation between each physical metric and each geometric metric is 0.14 for SOAP (range 0.08 to 0.28) and 0.05 for XRD (range 0.00 to 0.12). The most extreme value is 0.28, between Chamfer and SOAP.

Method rankings are also largely preserved. The ordering of the seven methods under SOAP is unchanged in 41 of the 47 configurations, and under XRD in 34 of the 38. The XRD deviations are a single recurring transposition of DiffCSP and MatterGen, whose median scores differ by 0.03 with overlapping confidence intervals, so all 38 XRD configurations yield one of only two orderings. Of the 21 method pairs, 19 show no reversal in any configuration.

Absolute values do shift, with the median relative change in the metric reaching roughly 3{,}860\% for $\sigma$ and 334\% for $r_\mathrm{cut}$. Since all methods are scored under identical settings, our conclusions require rank stability rather than magnitude stability, and absolute SOAP and XRD values should only be compared within one configuration.

\subsection{Effect of the number of known polymorphs}
\label{app:polymorph_count}

\begin{figure}[H]
    \centering
    \includegraphics[width=0.7\linewidth]{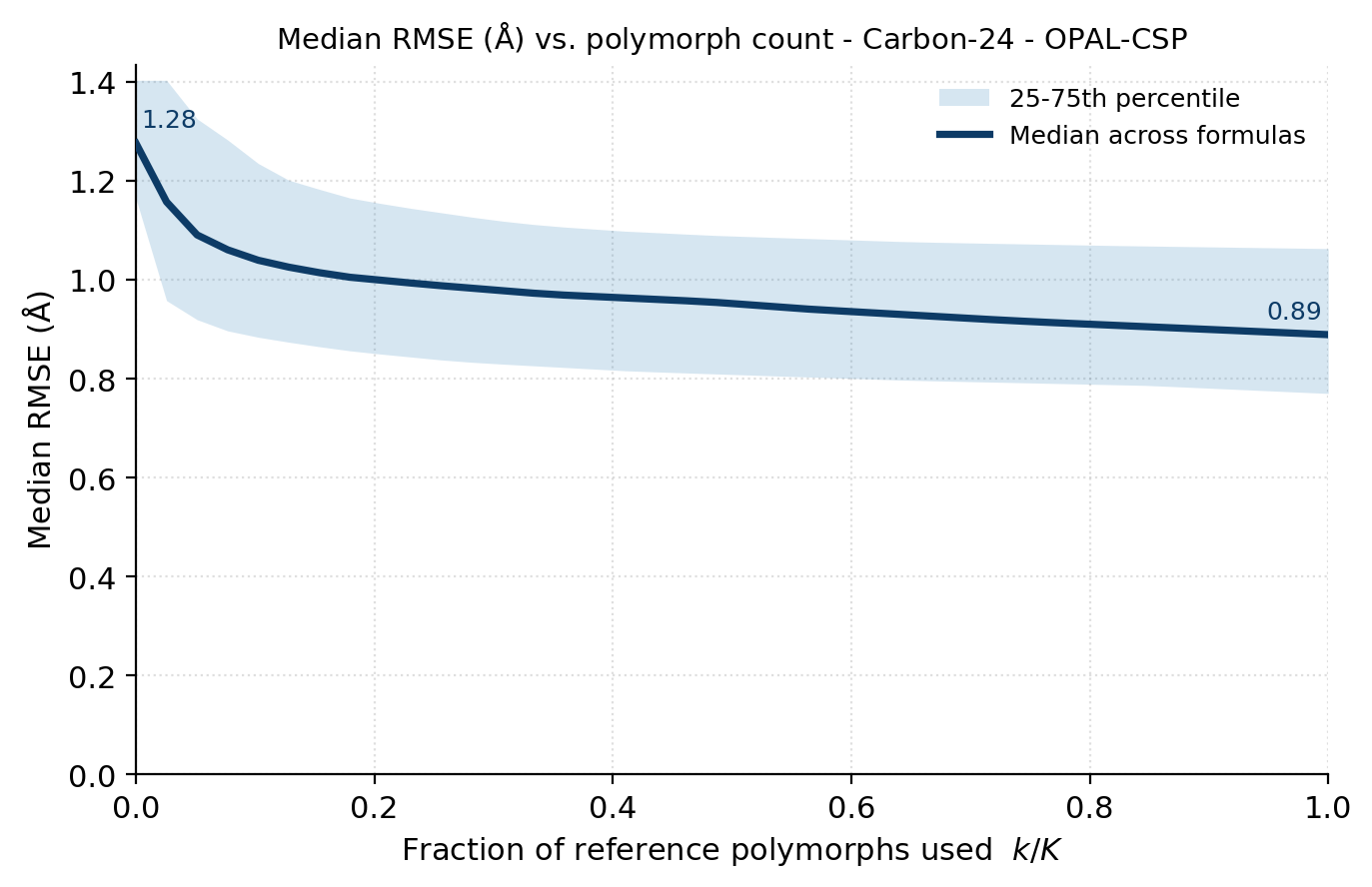}
    \caption{\textbf{Median minimum RMSE of \modelname{} on Carbon-24 as a function of the fraction of known polymorphs used as references.} For each composition, we recompute the minimum RMSE against random subsets of $k$ of its $K$ known polymorphs. The line shows the median across compositions and the shaded band the 25th to 75th percentiles.}
    \label{fig:polymorph_count}
\end{figure}

Each metric reports the minimum distance from a prediction to the known polymorphs of its composition, so a larger reference set raises the chance that some known structure lies close to the prediction, and the reported distance decreases as the number of polymorphs grows. We quantify this effect for the RMSE of \modelname{} on Carbon-24, the dataset with the most polymorphs per composition (Table~\ref{tab:polymorph_summary}). We hold each prediction fixed, recompute its minimum distance against random subsets of size $k$ drawn from its $K$ known polymorphs, and take the median across compositions at each $k$. The median minimum RMSE falls from 1.28~\AA{} at $k=1$ to 0.89~\AA{} at $k=K$ (Figure~\ref{fig:polymorph_count}), so the reference count alone drives a substantial decrease.

To assess how this effect changes the comparison between methods, we recompute all metrics under two alternative rules, which use as the single reference for every metric the polymorph closest to the prediction in formation energy or in RMSE, and compare the resulting rankings against those of the original protocol (Tables~\ref{tab:energy_reference_rule} and~\ref{tab:rmse_reference_rule}). On Perov-5, MP-20, and MPTS-52 the rankings are largely preserved, with occasional swaps between methods with close scores that do not affect our conclusions. Carbon-24 behaves differently, since every metric changes its ranking under at least one alternative rule. For XRD, MatterGen rises from fifth under the original protocol to first under the energy rule, and FlowMM rises from fourth to first under the RMSE rule, while DiffCSP falls from first to second and third, respectively. This is consistent with the overlapping bootstrap confidence intervals on Carbon-24 (Appendix~\ref{app:confidence_intervals}), which show no statistically significant advantage between methods.

\begin{table}[!tbp]
\centering
\caption{\textbf{CSP results under the formation-energy reference rule.} Each prediction is evaluated against a single reference, the known polymorph closest to it in formation energy, which is used for every metric. Formation energy distances therefore coincide with those of the original protocol (Table~\ref{tab:csp_results}). Values are median scores over test compositions; lower is better for all metrics. \textbf{Bold} marks the best method per dataset and metric; \underline{underline} marks the second best.}
\label{tab:energy_reference_rule}
{\renewcommand{\arraystretch}{1.2}\setlength{\tabcolsep}{4pt}%
\resizebox{\linewidth}{!}{%
\begin{tabular}{llrrrrrrrrr}
\toprule
\multicolumn{2}{c}{} & \multicolumn{4}{c}{\textbf{Geometric Distances (\AA)}} & \multicolumn{5}{c}{\textbf{Physical Distances}} \\
\cmidrule(lr){3-6}\cmidrule(lr){7-11}
 & \textbf{Model} & \textbf{Chamf.} $\downarrow$ & \textbf{Haus.} $\downarrow$ & \textbf{Sink.} $\downarrow$ & \textbf{RMSE} $\downarrow$ & \textbf{$\rho$ (g/cm$^3$)} $\downarrow$ & \textbf{SOAP} $\downarrow$ & \textbf{XRD} $\downarrow$ & \textbf{FE (eV/at.)} $\downarrow$ & \textbf{R.\ FE (eV/at.)} $\downarrow$ \\
\midrule
\multirow{7}{*}{\rotatebox[origin=c]{90}{\textsc{Perov-5}}} & FlowLLM & \underline{1.234} & \underline{0.633} & 1.010 & \underline{0.621} & 0.448 & 8.07 & 2.187 & 0.175 & $1.532 \cdot 10^{-1}$ \\
 & CDVAE & 2.327 & 1.291 & 1.676 & 1.188 & 0.439 & 10.95 & 3.221 & 3.008 & $1.971 \cdot 10^{-1}$ \\
 & CrystalFlow & 1.380 & 0.733 & 1.107 & 0.696 & 1.161 & 12.28 & 3.147 & 0.422 & $1.670 \cdot 10^{-1}$ \\
 & MatterGen & 1.255 & 0.643 & 1.004 & 0.629 & \underline{0.263} & \underline{6.56} & \textbf{1.688} & \underline{0.101} & \underline{$1.426 \cdot 10^{-1}$} \\
 & DiffCSP & 1.244 & 0.638 & \underline{0.963} & 0.625 & \textbf{0.242} & 6.63 & \underline{1.697} & 0.116 & {\boldmath $1.347 \cdot 10^{-1}$} \\
 & FlowMM & 1.249 & 0.641 & 1.000 & 0.626 & 0.359 & 7.11 & 1.901 & 0.109 & $1.518 \cdot 10^{-1}$ \\
\cmidrule(lr){2-11}
 & \modelname{} & \textbf{1.223} & \textbf{0.625} & \textbf{0.939} & \textbf{0.613} & 0.290 & \textbf{6.29} & 1.735 & \textbf{0.072} & $1.455 \cdot 10^{-1}$ \\
\midrule
\multirow{7}{*}{\rotatebox[origin=c]{90}{\textsc{Carbon-24}}} & FlowLLM & \textbf{1.931} & \textbf{1.623} & 16.283 & \textbf{1.094} & 0.338 & 92.30 & 4.800 & 3.791 & $1.331 \cdot 10^{-1}$ \\
 & CDVAE & 2.102 & 1.830 & 17.507 & 1.242 & \textbf{0.171} & \textbf{52.55} & 4.671 & 0.243 & $9.816 \cdot 10^{-2}$ \\
 & CrystalFlow & 2.116 & 1.876 & \textbf{15.274} & 1.241 & 0.515 & 140.31 & 7.578 & 0.448 & $2.085 \cdot 10^{-1}$ \\
 & MatterGen & \underline{2.017} & 1.843 & \underline{15.819} & \underline{1.177} & 0.242 & 64.00 & \textbf{3.970} & 0.182 & $1.581 \cdot 10^{-1}$ \\
 & DiffCSP & 2.039 & 2.200 & 17.564 & 1.271 & 0.206 & \underline{55.77} & \underline{4.203} & \underline{0.006} & \underline{$8.335 \cdot 10^{-2}$} \\
 & FlowMM & 2.028 & 1.854 & 16.491 & 1.211 & 0.266 & 68.42 & 4.335 & 0.069 & $1.647 \cdot 10^{-1}$ \\
\cmidrule(lr){2-11}
 & \modelname{} & 2.026 & \underline{1.789} & 16.445 & 1.288 & \underline{0.190} & 60.56 & 4.751 & \textbf{0.002} & {\boldmath $3.141 \cdot 10^{-3}$} \\
\midrule
\multirow{7}{*}{\rotatebox[origin=c]{90}{\textsc{MP-20}}} & FlowLLM & 4.052 & 2.990 & 8.196 & 2.337 & 1.146 & 11.17 & 3.099 & 2.481 & $1.515 \cdot 10^{-1}$ \\
 & CDVAE & 4.000 & 2.973 & 8.186 & 2.297 & \underline{0.156} & 3.82 & 1.612 & 0.216 & $5.073 \cdot 10^{-2}$ \\
 & CrystalFlow & 3.931 & 2.900 & 7.883 & 2.247 & 1.189 & 8.25 & 2.410 & 0.298 & {\boldmath $8.760 \cdot 10^{-5}$} \\
 & MatterGen & 4.132 & 3.033 & 8.339 & 2.373 & 0.223 & 3.59 & 1.626 & 0.066 & $4.120 \cdot 10^{-2}$ \\
 & DiffCSP & 3.933 & 2.894 & 7.910 & 2.241 & \textbf{0.129} & \textbf{2.20} & \textbf{0.949} & \textbf{0.019} & $8.197 \cdot 10^{-3}$ \\
 & FlowMM & \underline{3.875} & \underline{2.865} & \underline{7.793} & \underline{2.199} & 0.205 & \underline{2.43} & \underline{1.092} & 0.032 & \underline{$1.551 \cdot 10^{-4}$} \\
\cmidrule(lr){2-11}
 & \modelname{} & \textbf{3.715} & \textbf{2.686} & \textbf{7.004} & \textbf{2.096} & 0.215 & 2.58 & 1.142 & \underline{0.028} & $1.816 \cdot 10^{-4}$ \\
\midrule
\multirow{7}{*}{\rotatebox[origin=c]{90}{\textsc{MPTS-52}}} & FlowLLM & 4.358 & 3.858 & 21.979 & 2.783 & 2.619 & 29.73 & 4.162 & 3264.376 & $4.115 \cdot 10^{0}$ \\
 & CDVAE & 4.445 & 3.839 & 21.702 & 2.745 & 0.435 & 5.73 & 2.090 & 0.785 & $1.979 \cdot 10^{-1}$ \\
 & CrystalFlow & \textbf{4.282} & \textbf{3.661} & \textbf{20.670} & \textbf{2.605} & 0.285 & 4.17 & \underline{1.422} & 0.274 & \underline{$9.189 \cdot 10^{-2}$} \\
 & MatterGen & 4.525 & 3.938 & 22.342 & 2.809 & 0.265 & 4.12 & 1.495 & \underline{0.150} & $1.036 \cdot 10^{-1}$ \\
 & DiffCSP & 4.329 & 3.715 & 20.860 & 2.646 & \underline{0.202} & \underline{3.96} & 1.466 & 0.234 & $1.049 \cdot 10^{-1}$ \\
 & FlowMM & \underline{4.297} & \underline{3.691} & \underline{20.723} & \underline{2.634} & 0.292 & 4.87 & 1.651 & 0.619 & $1.343 \cdot 10^{-1}$ \\
\cmidrule(lr){2-11}
 & \modelname{} & 4.330 & 3.706 & 20.971 & 2.646 & \textbf{0.184} & \textbf{3.48} & \textbf{1.340} & \textbf{0.113} & {\boldmath $7.749 \cdot 10^{-2}$} \\
\bottomrule
\end{tabular}%
}}
\end{table}

\begin{table}[!tbp]
\centering
\caption{\textbf{CSP results under the RMSE reference rule.} Each prediction is evaluated against a single reference, the known polymorph closest to it in RMSE, which is used for every metric. RMSE values therefore coincide with those of the original protocol (Table~\ref{tab:csp_results}). Values are median scores over test compositions; lower is better for all metrics. \textbf{Bold} marks the best method per dataset and metric; \underline{underline} marks the second best.}
\label{tab:rmse_reference_rule}
{\renewcommand{\arraystretch}{1.2}\setlength{\tabcolsep}{4pt}%
\resizebox{\linewidth}{!}{%
\begin{tabular}{llrrrrrrrrr}
\toprule
\multicolumn{2}{c}{} & \multicolumn{4}{c}{\textbf{Geometric Distances (\AA)}} & \multicolumn{5}{c}{\textbf{Physical Distances}} \\
\cmidrule(lr){3-6}\cmidrule(lr){7-11}
 & \textbf{Model} & \textbf{Chamf.} $\downarrow$ & \textbf{Haus.} $\downarrow$ & \textbf{Sink.} $\downarrow$ & \textbf{RMSE} $\downarrow$ & \textbf{$\rho$ (g/cm$^3$)} $\downarrow$ & \textbf{SOAP} $\downarrow$ & \textbf{XRD} $\downarrow$ & \textbf{FE (eV/at.)} $\downarrow$ & \textbf{R.\ FE (eV/at.)} $\downarrow$ \\
\midrule
\multirow{7}{*}{\rotatebox[origin=c]{90}{\textsc{Perov-5}}} & FlowLLM & \underline{1.218} & \underline{0.623} & 0.971 & \underline{0.610} & 0.459 & 8.30 & 2.225 & 0.197 & $1.527 \cdot 10^{-1}$ \\
 & CDVAE & 2.295 & 1.272 & 1.655 & 1.172 & 0.438 & 10.95 & 3.242 & 3.038 & $1.931 \cdot 10^{-1}$ \\
 & CrystalFlow & 1.328 & 0.698 & 1.058 & 0.670 & 1.178 & 12.42 & 3.174 & 0.450 & $1.659 \cdot 10^{-1}$ \\
 & MatterGen & 1.231 & 0.631 & 0.955 & 0.619 & \underline{0.276} & 7.01 & \textbf{1.773} & \underline{0.128} & \underline{$1.442 \cdot 10^{-1}$} \\
 & DiffCSP & 1.229 & 0.625 & \underline{0.922} & 0.616 & \textbf{0.257} & \underline{6.98} & 1.802 & 0.141 & {\boldmath $1.368 \cdot 10^{-1}$} \\
 & FlowMM & 1.226 & 0.631 & 0.956 & 0.616 & 0.371 & 7.44 & 1.970 & 0.144 & $1.523 \cdot 10^{-1}$ \\
\cmidrule(lr){2-11}
 & \modelname{} & \textbf{1.211} & \textbf{0.617} & \textbf{0.909} & \textbf{0.607} & 0.303 & \textbf{6.72} & \underline{1.795} & \textbf{0.090} & $1.462 \cdot 10^{-1}$ \\
\midrule
\multirow{7}{*}{\rotatebox[origin=c]{90}{\textsc{Carbon-24}}} & FlowLLM & 1.699 & \textbf{1.373} & 13.768 & 0.931 & 0.370 & 101.35 & 5.688 & 3.898 & $3.232 \cdot 10^{-1}$ \\
 & CDVAE & \underline{1.615} & 1.479 & 13.863 & 0.947 & 0.375 & 93.42 & \underline{4.321} & 0.388 & \underline{$1.151 \cdot 10^{-1}$} \\
 & CrystalFlow & 1.664 & 1.445 & 12.992 & 0.984 & 0.569 & 147.64 & 5.310 & 0.507 & $2.014 \cdot 10^{-1}$ \\
 & MatterGen & 1.876 & 1.463 & 13.762 & 1.029 & 0.316 & 86.45 & 5.309 & 0.402 & $2.686 \cdot 10^{-1}$ \\
 & DiffCSP & 1.731 & 1.647 & 13.764 & 1.004 & \textbf{0.199} & \textbf{54.77} & 4.508 & \textbf{0.111} & {\boldmath $7.047 \cdot 10^{-2}$} \\
 & FlowMM & \textbf{1.499} & \underline{1.401} & \textbf{10.401} & \textbf{0.833} & \underline{0.224} & \underline{55.27} & \textbf{4.134} & 0.225 & $1.999 \cdot 10^{-1}$ \\
\cmidrule(lr){2-11}
 & \modelname{} & \underline{1.615} & 1.481 & \underline{12.563} & \underline{0.889} & 0.324 & 103.80 & 5.769 & \underline{0.194} & $2.186 \cdot 10^{-1}$ \\
\midrule
\multirow{7}{*}{\rotatebox[origin=c]{90}{\textsc{MP-20}}} & FlowLLM & 4.039 & 2.981 & 8.157 & 2.329 & 1.146 & 11.16 & 3.099 & 2.481 & $1.516 \cdot 10^{-1}$ \\
 & CDVAE & 3.988 & 2.960 & 8.129 & 2.289 & \underline{0.156} & 3.82 & 1.612 & 0.216 & $5.090 \cdot 10^{-2}$ \\
 & CrystalFlow & 3.911 & 2.889 & 7.821 & 2.236 & 1.190 & 8.25 & 2.408 & 0.298 & {\boldmath $8.665 \cdot 10^{-5}$} \\
 & MatterGen & 4.120 & 3.023 & 8.321 & 2.366 & 0.223 & 3.60 & 1.628 & 0.066 & $4.135 \cdot 10^{-2}$ \\
 & DiffCSP & 3.920 & 2.886 & 7.869 & 2.231 & \textbf{0.129} & \textbf{2.21} & \textbf{0.949} & \textbf{0.019} & $8.327 \cdot 10^{-3}$ \\
 & FlowMM & \underline{3.868} & \underline{2.854} & \underline{7.760} & \underline{2.194} & 0.205 & \underline{2.43} & \underline{1.094} & 0.032 & \underline{$1.515 \cdot 10^{-4}$} \\
\cmidrule(lr){2-11}
 & \modelname{} & \textbf{3.700} & \textbf{2.668} & \textbf{6.955} & \textbf{2.084} & 0.215 & 2.59 & 1.143 & \underline{0.028} & $1.850 \cdot 10^{-4}$ \\
\midrule
\multirow{7}{*}{\rotatebox[origin=c]{90}{\textsc{MPTS-52}}} & FlowLLM & 4.354 & 3.853 & 21.949 & 2.779 & 2.619 & 29.72 & 4.162 & 3297.953 & $4.056 \cdot 10^{0}$ \\
 & CDVAE & 4.439 & 3.833 & 21.671 & 2.741 & 0.436 & 5.72 & 2.089 & 0.786 & $1.980 \cdot 10^{-1}$ \\
 & CrystalFlow & \textbf{4.276} & \textbf{3.655} & \textbf{20.656} & \textbf{2.600} & 0.285 & 4.17 & \underline{1.422} & 0.274 & \underline{$9.206 \cdot 10^{-2}$} \\
 & MatterGen & 4.523 & 3.928 & 22.311 & 2.803 & 0.265 & 4.12 & 1.496 & \underline{0.151} & $1.038 \cdot 10^{-1}$ \\
 & DiffCSP & 4.323 & 3.710 & 20.825 & 2.641 & \underline{0.202} & \underline{3.96} & 1.466 & 0.234 & $1.049 \cdot 10^{-1}$ \\
 & FlowMM & \underline{4.294} & \underline{3.684} & \underline{20.699} & \underline{2.630} & 0.292 & 4.87 & 1.651 & 0.620 & $1.343 \cdot 10^{-1}$ \\
\cmidrule(lr){2-11}
 & \modelname{} & 4.325 & 3.700 & 20.943 & 2.642 & \textbf{0.184} & \textbf{3.48} & \textbf{1.341} & \textbf{0.113} & {\boldmath $7.764 \cdot 10^{-2}$} \\
\bottomrule
\end{tabular}%
}}
\end{table}

These results do not change our conclusions on Perov-5, MP-20, and MPTS-52, where few compositions have more than one known polymorph, but they show that Carbon-24 scores depend on the size of its reference sets and on the rule used to select among them. Carbon-24 values in Tables~\ref{tab:csp_results} and~\ref{tab:sicsp_results} should therefore be read with caution.

\subsection{Sensitivity of the energy metrics to the choice of MLIP}
\label{app:mlip_sensitivity}

We recompute formation energy and relaxed formation energy using each of the three potentials individually for all seven CSP methods on Perov-5 and Carbon-24. Since every method is scored under the same ensemble, a shift shared by all methods cancels in the ranking. For unrelaxed formation energy on Perov-5, the variability between potentials is 2.5 to 6.3\%, and the ranking under every potential is identical to the ensemble ranking. On Carbon-24 the magnitudes vary more, but only MACE-MP reorders the methods, transposing the top two, whose scores differ by about 1\%.

Relaxed formation energy varies more across potentials, with relative differences of 68.9 to 91.3\% on Perov-5. This variation comes from the definition of the metric rather than from unstable rankings. Since the predicted and reference structures are relaxed independently, both move toward low-energy configurations and the resulting difference is small for every method, between 0.126 and 0.178 eV/atom on Perov-5 (Table~\ref{tab:csp_results}), so even modest absolute changes translate into large relative differences. These changes also affect all methods in a similar way. For every method, Orb gives the lowest values and UMA the highest, and moving from Orb to MACE-MP or UMA raises the scores of all methods by a similar amount, 0.086 and 0.119 eV/atom on average. As a result, UMA reproduces the ensemble ranking exactly, and Orb and MACE-MP only swap methods whose scores differ by at most 0.010 eV/atom. On Carbon-24, six of the seven methods reach values between $10^{-4}$ and $10^{-2}$ eV/atom. At this scale the differences between methods are mostly numerical noise, so the ranking changes across potentials, consistent with the overlapping confidence intervals in Appendix~\ref{app:confidence_intervals}.

We also compare the ensemble with DFT on LiNbO$_3$, a representative Perov-5 composition. We compute the energy of the ground truth and of each prediction with Quantum ESPRESSO\footnote{\url{https://www.quantum-espresso.org}} under PBE, using cutoff and $k$-point settings converged for this structure, and obtain the formation energy distance as in Appendix~\ref{physical_appendix}. DFT and the ensemble order the seven methods identically, with a mean absolute error of 0.075 eV/atom between them. This error is smaller than the differences between most methods but larger than the gap between the closest ones, so the ensemble follows DFT at a coarse level, although this comparison does not validate the fine-grained ordering.

For relaxed formation energy, we relax the ground truth and the seven predictions with DFT. Two relaxations of the ground truth from different perturbed starting points reach the same minimum, which coincides with the one reached by the prediction of MatterGen (relaxed distance below 0.001 eV/atom). CrystalFlow and DiffCSP relax to higher energies (0.024 and 0.049 eV/atom), while the predictions of \modelname{}, CDVAE, FlowLLM, and FlowMM reach a common minimum 0.036 to 0.041 eV/atom below the ground truth. The ensemble agrees with DFT in placing MatterGen first and CrystalFlow among the closest methods, but it places \modelname{} second, whereas DFT relaxes it to the lower minimum, and the Spearman correlation between the two rankings is 0.21. This suggests that relaxed formation energy can be more sensitive than its unrelaxed counterpart to the relaxation protocol, since MLIP and DFT relaxations may end in different minima. Since the metric is an absolute difference, it also assigns a nonzero distance to predictions that relax to a lower energy than the reference. This is a single example, however, and assessing the actual extent of these effects would require repeating the analysis on more structures.

A related consideration is that each potential relaxes the structures independently, so the ensemble value for relaxed formation energy averages energies evaluated at up to three different relaxed geometries. We use the mean over potentials, although the median could be a more robust alternative when one potential deviates from the other two.

\subsection{Agreement between metric rankings}
\label{app:ranking_agreement}

Removing matching tolerances eliminates one source of ranking instability, but a protocol with several metrics can still rank methods differently depending on the metric. To assess how structured this variation is, we compute for each dataset the Spearman correlation between the rankings of the seven CSP methods under every pair of metrics, and average it within three groups, namely the six pairs of geometric metrics, the ten pairs of physical metrics, and the twenty pairs that combine one geometric and one physical metric. As a reference for the previous convention, we also compute the correlation between the rankings under Match Rate and RMSE (Table~\ref{tab:supp_mr_rmse}).

\begin{table}[H]
  \centering
  \caption{\textbf{Agreement between the method rankings of different metrics.} Mean Spearman correlation between the rankings of the seven CSP methods under pairs of metrics, averaged within each group of pairs. The last column reports the correlation between the rankings under the conventional Match Rate and RMSE.}
  \label{tab:ranking_agreement}
  \small
  \begin{tabular}{lcccc}
    \toprule
    \textbf{Dataset} & \textbf{Geometric} & \textbf{Physical} & \textbf{Geometric vs.\ physical} & \textbf{MR vs.\ RMSE} \\
    \midrule
    Perov-5   & 0.846 & 0.832 & 0.608    & $-0.180$ \\
    Carbon-24 & 0.786 & 0.543 & 0.059    & $-0.955$ \\
    MP-20     & 0.976 & 0.604 & 0.521    & 0.649 \\
    MPTS-52   & 0.964 & 0.914 & 0.243    & 0.714 \\
    \bottomrule
  \end{tabular}
\end{table}

Rankings agree strongly within the geometric family (0.786 to 0.976) and within the physical family (0.543 to 0.914). In contrast, Match Rate and RMSE range from reasonable agreement on MP-20 and MPTS-52 (0.649 and 0.714) to low or negative agreement on Perov-5 and Carbon-24 ($-0.180$ and $-0.955$), even though both aim to capture geometric similarity. Agreement drops mainly when one geometric and one physical metric are compared (0.059 to 0.608), which reflects the separation between geometric and physical performance that the benchmark is designed to expose. The variation that remains is therefore organized by metric family rather than arbitrary. Carbon-24 shows the lowest agreement within each family, consistent with its structural diversity and the closely spaced scores of its methods (Appendix~\ref{app:confidence_intervals}).

\subsection{Evaluation beyond known polymorphs}
\label{app:beyond_known}

\benchname{} evaluates each prediction against the polymorphs available in the reference datasets, so it measures how well a method recovers known structures. A prediction that corresponds to a valid polymorph absent from these datasets therefore receives a large distance, even though it may be physically correct, and the benchmark cannot distinguish it from an incorrect prediction. As crystal databases incorporate newly reported polymorphs, future versions of \benchname{} can extend the reference sets accordingly, which reduces this effect over time.

\benchname{} also generates a single prediction per composition, which does not capture the diversity of the structures that a method can produce. With several predictions per composition, future versions could measure diversity as the number of distinct structures after relaxation and deduplication, and coverage as the fraction of known polymorphs that these predictions recover, stratified by the number of polymorphs per composition. Finally, for predictions that do not match any known polymorph, the MLIP ensemble that we use for the energy metrics can still compute their formation energy without a known reference structure, which provides a first estimate of their physical plausibility.


\newpage
\section*{NeurIPS Paper Checklist}

\begin{enumerate}

\item {\bf Claims}
    \item[] Question: Do the main claims made in the abstract and introduction accurately reflect the paper's contributions and scope?
    \item[] Answer: \answerYes{}  
    \item[] Justification: The abstract and introduction state three contributions, all substantiated in the paper: (1) OPAL-Bench, a threshold-free polymorph-aware benchmark with SE(3) alignment unifying geometric and physical metrics (Section 3); (2) a systematic re-evaluation of 7 CSP and 2 SICSP methods showing geometric performance does not reliably imply physical plausibility (Section 5.2, Tables 1–2, Appendix D.1); (3) OPAL-CSP, a flow-matching reference baseline (Section 4, Appendix B.3). Limitations and scope are discussed in Section 5.4.
    

\item {\bf Limitations}
    \item[] Question: Does the paper discuss the limitations of the work performed by the authors?
    \item[] Answer: \answerYes{} 
    \item[] Justification: The paper includes a dedicated limitations section discussing four key aspects of \benchname{}: (1) the polymorph-aware protocol penalizes predictions matching undocumented polymorphs; (2) systematic biases in the fixed relaxation and property-evaluation pipeline propagate into physical metric scores; (3) the single prediction per-composition design does not capture generative diversity; (4) evaluating pretrained models requires checking for data leakage between their training data and the test splits.


\item {\bf Theory assumptions and proofs}
    \item[] Question: For each theoretical result, does the paper provide the full set of assumptions and a complete (and correct) proof?
    \item[] Answer: \answerNA{} 
    \item[] Justification: The paper does not include theoretical results; it focuses on the design and empirical evaluation of a benchmark for crystal structure prediction.


    \item {\bf Experimental result reproducibility}
    \item[] Question: Does the paper fully disclose all the information needed to reproduce the main experimental results of the paper to the extent that it affects the main claims and/or conclusions of the paper (regardless of whether the code and data are provided or not)?
    \item[] Answer: \answerYes{}  
    \item[] Justification: The paper provides all information necessary to reproduce the main experimental results. The benchmark methodology is fully described, including the formal problem formulation, the complete set of geometric and physical metrics, the polymorph-aware evaluation protocol, and the species-aware SE(3) alignment procedure with its mathematical formulation (Sections 3.1–3.5). Dataset splits are reported in Section 5.1, and model training details (optimizer, learning rate, scheduler, batch size, number of epochs and GPUs) and inference settings for \modelname{} in Appendix B.3.2, with architecture details in Appendix B.3.1 and baseline details in Appendix B.4. The re-evaluation of the seven CSP and two SICSP methods follows the original training procedures described in their respective papers, and any protocol-specific preprocessing steps, such as supercell construction for SICSP geometric evaluation, are explicitly described. Together, these disclosures are sufficient to reproduce the benchmark results and the central finding that geometric performance does not reliably imply physical plausibility.

\item {\bf Open access to data and code}
    \item[] Question: Does the paper provide open access to the data and code, with sufficient instructions to faithfully reproduce the main experimental results, as described in supplemental material?
    \item[] Answer: \answerYes{}  
    \item[] Justification: We release the full OPAL-Bench evaluation framework, including the polymorph-aware protocol, species-aware SE(3) alignment, and all geometric and physical metric implementations, at \url{https://github.com/BCV-Uniandes/OPAL}. Training and data preprocessing scripts for \modelname{} are available at the same link.

\item {\bf Experimental setting/details}
    \item[] Question: Does the paper specify all the training and test details (e.g., data splits, hyperparameters, how they were chosen, type of optimizer) necessary to understand the results?
    \item[] Answer: \answerYes{} 
    \item[] Justification: Training and architectural details of the baseline model, including hyperparameters and optimizer settings, are fully specified in the appendix. Data splits follow the official partitions of the established benchmark datasets used, as described in the main paper, and all metric definitions and computation details are presented in both the main paper and the appendix.

\item {\bf Experiment statistical significance}
    \item[] Question: Does the paper report error bars suitably and correctly defined or other appropriate information about the statistical significance of the experiments?
    \item[] Answer: \answerYes{} 
    \item[] Justification: The paper reports median metric scores across all test compositions for each method, where the median is computed over the distribution of per-prediction metric values. Bootstrap confidence intervals are derived by resampling this distribution and computing the median over each resample, yielding a robust estimate of uncertainty around the reported scores. These confidence intervals are reported alongside all main benchmark results, providing appropriate statistical grounding for the comparisons made across methods and datasets.


\item {\bf Experiments compute resources}
    \item[] Question: For each experiment, does the paper provide sufficient information on the computer resources (type of compute workers, memory, time of execution) needed to reproduce the experiments?
    \item[] Answer: \answerYes{} 
    \item[] Justification: The paper reports compute resources in Section B.5: all training was performed on NVIDIA Quadro RTX 8000 GPUs (48 GB memory), using between 1 and 7 GPUs per run depending on the model and dataset, and metric computation ran on 2× Intel Xeon Gold 6230 CPUs (40 physical cores, 2.10 GHz). Section B.3.2 specifies the OPAL-CSP training schedule (1000 epochs, total batch size 128 across GPUs, Adam optimizer with reduce-on-plateau scheduling), and Appendix B.4 documents that baseline methods were re-trained following their original procedures. Appendix B.5 reports the training time of \modelname{} on each dataset; we do not report the compute used during preliminary development.
    
\item {\bf Code of ethics}
    \item[] Question: Does the research conducted in the paper conform, in every respect, with the NeurIPS Code of Ethics \url{https://neurips.cc/public/EthicsGuidelines}?
    \item[] Answer: \answerYes{} 
    \item[] Justification: The research conforms with the NeurIPS Code of Ethics in all applicable respects. The work is purely computational and involves no human subjects, personally identifiable data, or sensitive information. The datasets used are publicly available and are used in accordance with their respective licenses. The research poses no foreseeable risks related to safety, security, discrimination, or surveillance, as it focuses on the development of an evaluation benchmark for crystal structure prediction. Computational resources are transparently reported, and all research artifacts are released publicly with sufficient documentation to enable reproducibility and external scrutiny.


\item {\bf Broader impacts}
    \item[] Question: Does the paper discuss both potential positive societal impacts and negative societal impacts of the work performed?
    \item[] Answer: \answerYes{} 
    \item[] Justification: The paper discusses societal impacts in Section 5.4. On the positive side, \benchname{} supports more rigorous evaluation of crystal structure prediction methods, which may benefit materials discovery in areas such as energy storage, catalysis, and microelectronics. On the negative side, the paper acknowledges that more capable generative models could potentially accelerate the design of undesirable materials. We assess this risk as limited because computationally predicted structures still require experimental synthesis and characterization before physical realization.
    
\item {\bf Safeguards}
    \item[] Question: Does the paper describe safeguards that have been put in place for responsible release of data or models that have a high risk for misuse (e.g., pre-trained language models, image generators, or scraped datasets)?
    \item[] Answer: \answerNA{} 
    \item[] Justification: The models and datasets released in this paper pose no significant risk of misuse. \modelname{} is a domain-specific crystal structure prediction model whose outputs require rigorous experimental validation before physical realization, and the benchmark datasets are established publicly available resources that contain no sensitive or personally identifiable information.

\item {\bf Licenses for existing assets}
    \item[] Question: Are the creators or original owners of assets (e.g., code, data, models), used in the paper, properly credited and are the license and terms of use explicitly mentioned and properly respected?
    \item[] Answer: \answerYes{} 
    \item[] Justification: All datasets and methods used in this paper are properly credited and their licenses are respected. The four benchmark datasets are publicly available and appropriately cited: Carbon-24 is released under a CC BY 4.0 license via Materials Cloud, and MP-20 and MPTS-52 are sourced from the Materials Project, which distributes its data under a CC BY 4.0 license; Perov-5 is a publicly available dataset cited from its original publication. The code of the six previously published CSP methods and the two SICSP methods re-evaluated under OPAL-Bench is publicly available through open-source repositories accompanying their respective published papers, and we use them in accordance with their stated licenses. We also use the MLIP ensemble of LeMat-GenBench, the LeMat-Bulk dataset, CSPML, Dans\_Diffraction, and Quantum ESPRESSO, all cited or linked in the paper and used under their respective licenses.


\item {\bf New assets}
    \item[] Question: Are new assets introduced in the paper well documented and is the documentation provided alongside the assets?
    \item[] Answer: \answerYes{} 
    \item[] Justification: The paper introduces two new assets: \benchname{}, a benchmark for crystal structure prediction evaluation, and \modelname{}, a reference baseline model. We release OPAL-Bench at \url{https://github.com/BCV-Uniandes/OPAL}. The repository includes the full implementation of the benchmark protocol, covering the polymorph-aware evaluation procedure, the species-aware SE(3) alignment, and the computation of all geometric and physical metrics. The implementation, training scripts, and inference code for \modelname{} are available at the same link. Documentation is provided in the repository itself, with additional details on the benchmark methodology in the main paper (Sections 3.1–3.5) and further architecture, training, and implementation details in the appendix, ensuring that both assets are sufficiently documented to be understood, reproduced, and extended by the research community.


\item {\bf Crowdsourcing and research with human subjects}
    \item[] Question: For crowdsourcing experiments and research with human subjects, does the paper include the full text of instructions given to participants and screenshots, if applicable, as well as details about compensation (if any)? 
    \item[] Answer: \answerNA{} 
    \item[] Justification: The paper does not involve crowdsourcing or research with human subjects, as it focuses entirely on computational methods for crystal structure prediction evaluation.


\item {\bf Institutional review board (IRB) approvals or equivalent for research with human subjects}
    \item[] Question: Does the paper describe potential risks incurred by study participants, whether such risks were disclosed to the subjects, and whether Institutional Review Board (IRB) approvals (or an equivalent approval/review based on the requirements of your country or institution) were obtained?
    \item[] Answer: \answerNA{} 
    \item[] Justification: The paper does not involve human subjects. All experiments are purely computational and operate exclusively on crystal structure datasets.

\item {\bf Declaration of LLM usage}
    \item[] Question: Does the paper describe the usage of LLMs if it is an important, original, or non-standard component of the core methods in this research? Note that if the LLM is used only for writing, editing, or formatting purposes and does \emph{not} impact the core methodology, scientific rigor, or originality of the research, declaration is not required.
    \item[] Answer: \answerNA{} 
    \item[] Justification: LLMs are not used as any important, original, or non-standard component of the core methodology. The paper focuses on crystal structure prediction evaluation and does not incorporate LLMs in any aspect of its methods or experiments.


\end{enumerate}

\end{document}